\documentclass[10pt,twocolumn,letterpaper]{article}

\usepackage{mysty}
\usepackage{times}
\usepackage{epsfig}
\usepackage{graphicx}
\usepackage{amsmath}
\usepackage{amssymb}

\usepackage{color}
\usepackage{calc}
\usepackage{algorithm,algpseudocode}
\usepackage{booktabs}
\usepackage{multirow}
\usepackage{makecell}

\usepackage{array}
\usepackage{subfig}
\usepackage[table]{xcolor}

\usepackage{placeins}

\newcommand{\patchify}[1]{\operatorname{patchify}{\left(#1\right)}}
\newcommand{\shuffle}[1]{\operatorname{shuffle}{\left(#1\right)}}
\newcommand{\unshuffle}[1]{\operatorname{unshuffle}{\left(#1\right)}}
\newcommand{\argsort}[1]{\operatorname{argsort}{\left(#1\right)}}
\newcommand{\sequence}[3]{\left\{#1\right\}_{#2}^{#3}}
\newcommand{\seqin}[3]{\{#1\}_{#2}^{#3}}
\newcommand{\sindex}[2]{\operatorname{index}{\left(#1,#2\right)}}

\newcommand{\norm}[1]{\left \Vert #1 \right \Vert}

\newcommand{\expb}[1]{\exp\left( #1 \right)}

\newcommand{\similar}[2]{\operatorname{sim}{\left(#1,#2\right)}}

\usepackage[pagebackref=true,breaklinks=true,letterpaper=true,colorlinks,bookmarks=false]{hyperref}

\iccvfinalcopy 

\ificcvfinal\fi

\begin{document}

\title{Inter-Instance Similarity Modeling for Contrastive Learning}

\author{Chengchao Shen\textsuperscript{1}, Dawei Liu\textsuperscript{1}, Hao Tang\textsuperscript{1}, 
Zhe Qu\textsuperscript{1}, Jianxin Wang\textsuperscript{1}\\
\textsuperscript{1}Central South University\\
{\tt\small \{scc.cs,tanghao,zhe\_qu\}@csu.edu.cn,liudw0702@163.com,jxwang@mail.csu.edu.cn}
}

\maketitle
\ificcvfinal\thispagestyle{empty}\fi

\begin{abstract}
   The existing contrastive learning methods widely adopt one-hot instance discrimination as pretext task for self-supervised learning, which inevitably neglects rich inter-instance similarities among natural images, then leading to potential representation degeneration. 
   
   In this paper, we propose a novel image mix method, PatchMix, for contrastive learning in Vision Transformer (ViT), to model inter-instance similarities among images. 
   Following the nature of ViT, we randomly mix multiple images from mini-batch in patch level to construct mixed image patch sequences for ViT. 
   Compared to the existing sample mix methods, our PatchMix can flexibly and efficiently mix more than two images and simulate more complicated similarity relations among natural images.
   In this manner, our contrastive framework can significantly reduce the gap between contrastive objective and ground truth in reality.
   Experimental results demonstrate that our proposed method significantly outperforms the previous state-of-the-art on both ImageNet-1K and CIFAR datasets, e.g., 3.0\% linear accuracy improvement on ImageNet-1K and 8.7\% kNN accuracy improvement on CIFAR100. 
   Moreover, our method achieves the leading transfer performance on downstream tasks, object detection and instance segmentation on COCO dataset. 
   The code is available at \url{https://github.com/visresearch/patchmix}.

\end{abstract}

\section{Introduction}

\begin{figure}[t]
   \centering
   \includegraphics[width=\linewidth]{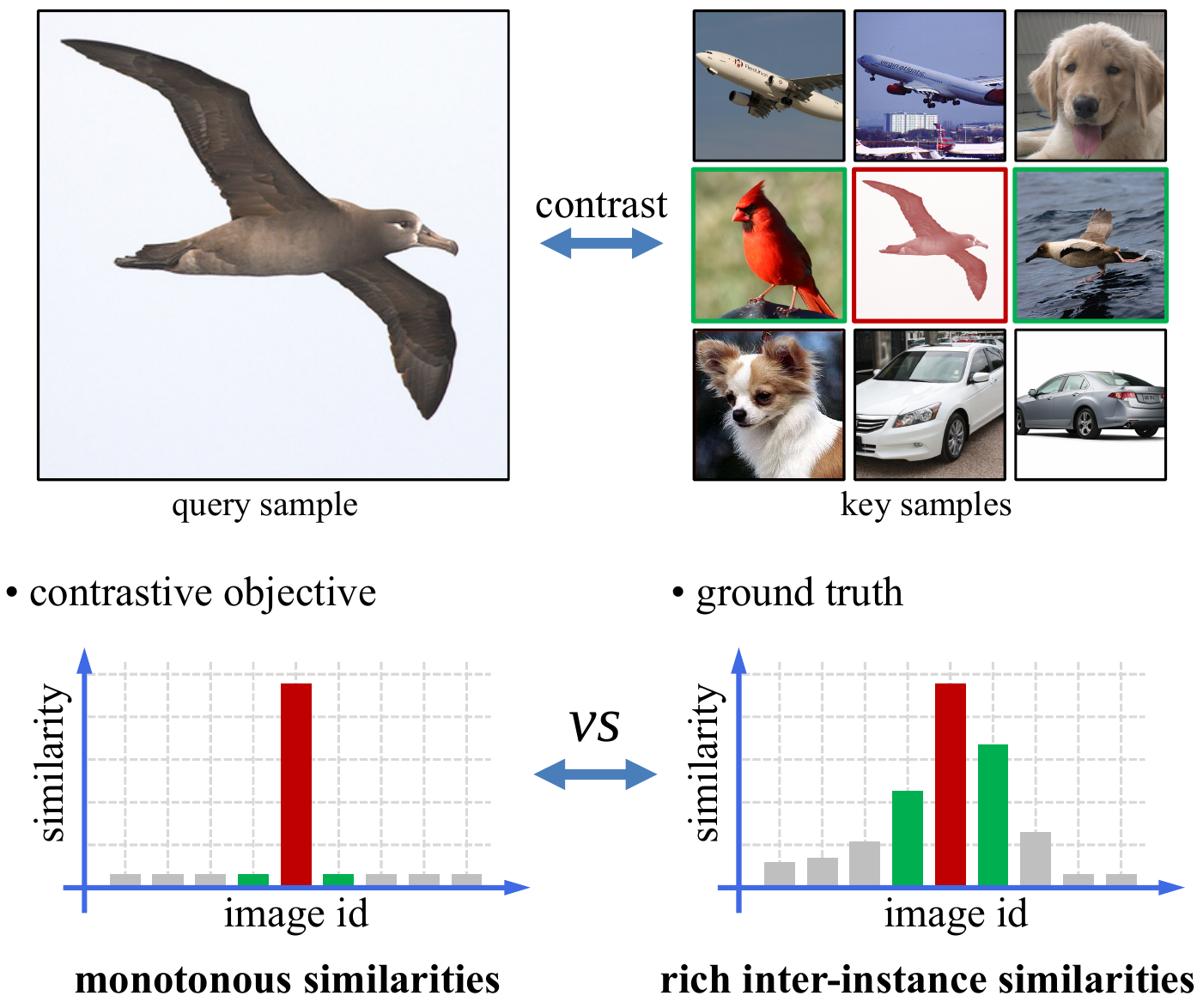}
   \caption{Monotonous similarity in contrastive objective \emph{vs} rich inter-instance similarities among natural images. 
   Simply contrasting different views from the same image may miss potential positive samples and thus degrade representation performance.
   }
   \label{fig:problem}
\end{figure}

Massive contrastive learning studies~\cite{wu2018unsupervised,chen2020simple,he2020momentum,chen2020improved} have achieved impressive performance on unsupervised visual representation learning. 
Although various designs are proposed to improve the representation performance, these methods can be summarized into an instance discrimination task.
In this pretext task, the unlabeled images are augmented into several views with different appearances. 
Under the contrastive objective, the model is trained to capture appearance-invariant representations and understand the semantics consistency between positive pairs in the different views. 
Specifically, this objective aims to maximize the similarity between views augmented from the same images, meanwhile minimize the one between samples from different images. 
However, as shown in Figure~\ref{fig:problem}, there are rich similarities among different image samples in reality, which is ignored by the existing contrastive frameworks. 
In other words, this issue overlooks massive potential positive samples, thus introducing inaccurate targets into contrastive learning.
The discrepancy between one-hot targets and real label distributions hurts the performance of unsupervised representation. 

Supervised classification tasks suffer a similar issue, where one-hot classification targets omit potential inter-class similarity. 
Image mix strategies, such as Mixup~\cite{zhang2017mixup} and CutMix~\cite{yun2019cutmix}, are proposed to soften one-hot labels by simply mixing image pairs. 
These strategies are also applied to contrastive learning~\cite{lee2021mix,verma2021towards,kalantidis2020hard} to alleviate inaccurate target issue. 
However, for contrastive learning task, inter-instance similarity is significantly richer than inter-class similarity in supervised classification task, not only including inter-class similarity as classification task but also intra-class similarity (image instances from the same class are more likely to be similar to each other). 
In other words, the real targets for contrastive learning are softer than the ones for supervised classification. 
The plain Mixup and CutMix strategy only supports similarity modeling between two images, but fails to construct similarities among more images. 

Vision Transformer (ViT)~\cite{dosovitskiy2021vit} takes image as the sequence of image patches, which provides a novel perspective for image modeling. 
Furthermore, the success of masked image modeling~\cite{bao2022beit,xie2022simmim,he2022masked} demonstrates that small portion of patches from image, e.g., 25\%, can well represent the semantics of the original image, which allows mixing more images into a hybrid one. 

Following the nature of ViT, we propose a novel image mix method, \emph{PatchMix}, which randomly mixes several patch sequences of images, to simulate training samples with rich inter-instance similarities in an unsupervised manner. 
For example, we mix four images, including dog, bird, airplane and car, into a hybrid patch sequence of image, whose patches contain dog head, bird wing, tail plane and wheel. 
This hybrid sample can effectively encourage the trained model to explore rich component similarities among multiple images and improve the generalization of representations. 
Moreover, in the hybrid sequence, the inconsecutive patches from the same image regularize the trained model to construct long distance dependency among patches. 

To model inter-instance similarities, we conduct contrastive learning from three aspects. 
First, \emph{mix-to-origin contrast} is conducted to construct similarities between the mixed images and the original ones, where each mixed image corresponds to multiple original images. 
Second, \emph{mix-to-mix contrast} provides more positive pairs than mix-to-origin contrast to further model more complicated relations between the mixed images. 
Third, \emph{origin-to-origin contrast} is used to eliminate potential representation gap produced by the domain gap between mixed images and original ones.

In summary, the main contribution of our paper is a novel image mix method dedicated to ViT, PatchMix, which can effectively mixes multiple images to simulate rich inter-instance similarities among natural images. 
Experimental results demonstrate that our method effectively captures rich inter-instance similarities and significantly improves the quality of unsupervised representations, achieving state-of-the-art performance on image classification, object detection and instance segmentation.

\section{Related Work}
In this section, we discuss related work from two aspects, contrastive learning and image mix strategies and analyze the main differences between our method and related work.

\subsection{Contrastive Learning}
To learn meaningful representations from unlabeled image data, contrastive learning methods~\cite{wu2018unsupervised,chen2020simple,he2020momentum} conduct instance discrimination by maximizing representation similarity between positive pairs and minimizing the one between negative pairs. 
Several key components are proposed to improve the performance of contrastive representations. 

First, more and stronger data augmentation techniques, such as random crop and color jitter, are applied to positive pairs~\cite{chen2020simple,grill2020bootstrap}. 
Hence, the trained model is encouraged to learn appearance-invariant representations from samples under different distortions. 
Second, additional modules, such as projection~\cite{chen2020simple,he2020momentum} and prediction head~\cite{grill2020bootstrap}, are used to improve the transferability of contrastive representations by decoupling the representations from contrastive pretext task. 
To construct a more stable dynamic look-up dictionary, momentum encoder is introduced by MoCo series~\cite{he2020momentum,chen2020improved,chen2021empirical}. 
Third, clustering-based methods~\cite{caron2020unsupervised} and prediction-based methods~\cite{grill2020bootstrap,chen2021exploring} are proposed to explore contrastive learning without negative pairs.

In spite of encouraging performance achieved, these works neglect the potential similarities among image instances, where massive potential positive samples are ignored during training and the trained model is hurt by misleading targets.
In this work, we efficiently construct learning targets with multi-instance similarities by mixing multiple images. 
In this manner, the trained network is encouraged to model rich inter-instance similarities and then improves the quality of unsupervised representations.

\subsection{Image Mix}

Image mix strategies are widely adopted in image classification to reduce the overfitting risk by softening the classification labels. 
Mixup~\cite{zhang2017mixup} forms a mixed sample by applying a weighted sum between two randomly sampled images, where the weights for sum are used as the classification labels. 
Following the idea of Mixup, CutMix~\cite{yun2019cutmix} pastes the patch from image to another one, to regularize the training of model. 
To alleviate uninformative patch mix introduced by CutMix, saliency-guided mix strategies~\cite{uddin2021saliencymix,dabouei2021supermix,kim2021co} are proposed to mix informative parts from two images according to saliency maps. 
To reduce computation cost of saliency-guided mix strategies, TokenMixup~\cite{choi2022tokenmixup} mixes informative parts from two images by off-the-shelf attention scores of Transformer architecture. 
Apart from mix on raw images, TokenMix~\cite{liu2022tokenmix} and TokenMixup~\cite{choi2022tokenmixup} also mix the tokens from two images under Transformer architecture. 
Manifold Mixup~\cite{verma2019manifold} and Alignmixup~\cite{venkataramanan2022alignmixup} also explore image mix on intermediate representations. 

Image mix strategies are also be researched in contrastive learning~\cite{lee2021mix,verma2021towards,kalantidis2020hard}. 
SDMP~\cite{ren2022simple} conducts a series of image mix augmentation strategies: Mixup, CutMix and ResizeMix~\cite{qin2020resizemix} to obtain a novel mixed image from two given images, which form a positive triplet to perform contrastive learning.
RegionCL~\cite{xu2022regioncl} swaps the region of two images to obtain additional positive pairs for contrastive learning.

Although these methods effectively alleviate the risk of overfitting, they only consider similarity between two image instances and fail to simulate richer inter-instance similarities in reality. 
The proposed PatchMix can be easily extended to a more general case, where any number of images can be mixed to model more sophisticated similarity relations among images in reality.

\section{The Proposed Method}
In this section, we present our patch mix strategy to construct hybrid image instances for the simulation of rich inter-instance similarities in reality. 
Then, we conduct mix-to-origin, mix-to-mix and origin-to-origin contrast using hybrid samples constructed by the proposed patch mix to model inter-instance similarities. 

\begin{table}[t]
   \centering
   \resizebox{\linewidth}{!}
   {
      \begin{tabular}{c|m{6cm}<{\centering}}
         \hline
         \textbf{symbol} & \textbf{description} \\
         \hline
         $x_i$      & \makecell[l]{the $i$-th image sample in image batch $x$} \\
         \hline
         $x_i^{\rm shuffle}$ & \makecell[l]{the image $x_i$ whose patch sequence is \\ randomly shuffled.} \\
         \hline
         $x_i^{\rm smix}$ & \makecell[l]{the image applied patch mix in a shuffled \\ patch order} \\
         \hline
         $p_{ij}$   & \makecell[l]{the $j$-th image patch from sample $x_i$} \\
         \hline
         $G_{im}$   & \makecell[l]{the $m$-th patch group of $x_i$} \\
         \hline
         $k(j)$     & \makecell[l]{the index item after shuffling $\{ j \}_{j=0}^{T-1}$ } \\
         \hline
         $u(i,m)$   & \makecell[l]{the index transform for patch mix} \\
         \hline
         $r(j)$     & \makecell[l]{the index after recovering the order of patch \\ sequence} \\
         \hline
         $v(i, j)$  & \makecell[l]{the index transform from group index to \\ patch index} \\
         \hline
         $w(i, j)$  & \makecell[l]{the index transform from shuffled index to \\ unshuffled index} \\
         \hline
         $l$        & \makecell[l]{1-d index for the original patch sequence of \\ mini-batch} \\
         \hline
         $q$        & \makecell[l]{1-d index for patch mix in mini-batch} \\
         \hline
      \end{tabular}
   }
   \caption{List of symbols for patch mix strategy.}
   \label{tab:symbol}
\end{table}

\subsection{Patch Mix}

\begin{figure}[t]
   \centering
   \includegraphics[width=\linewidth]{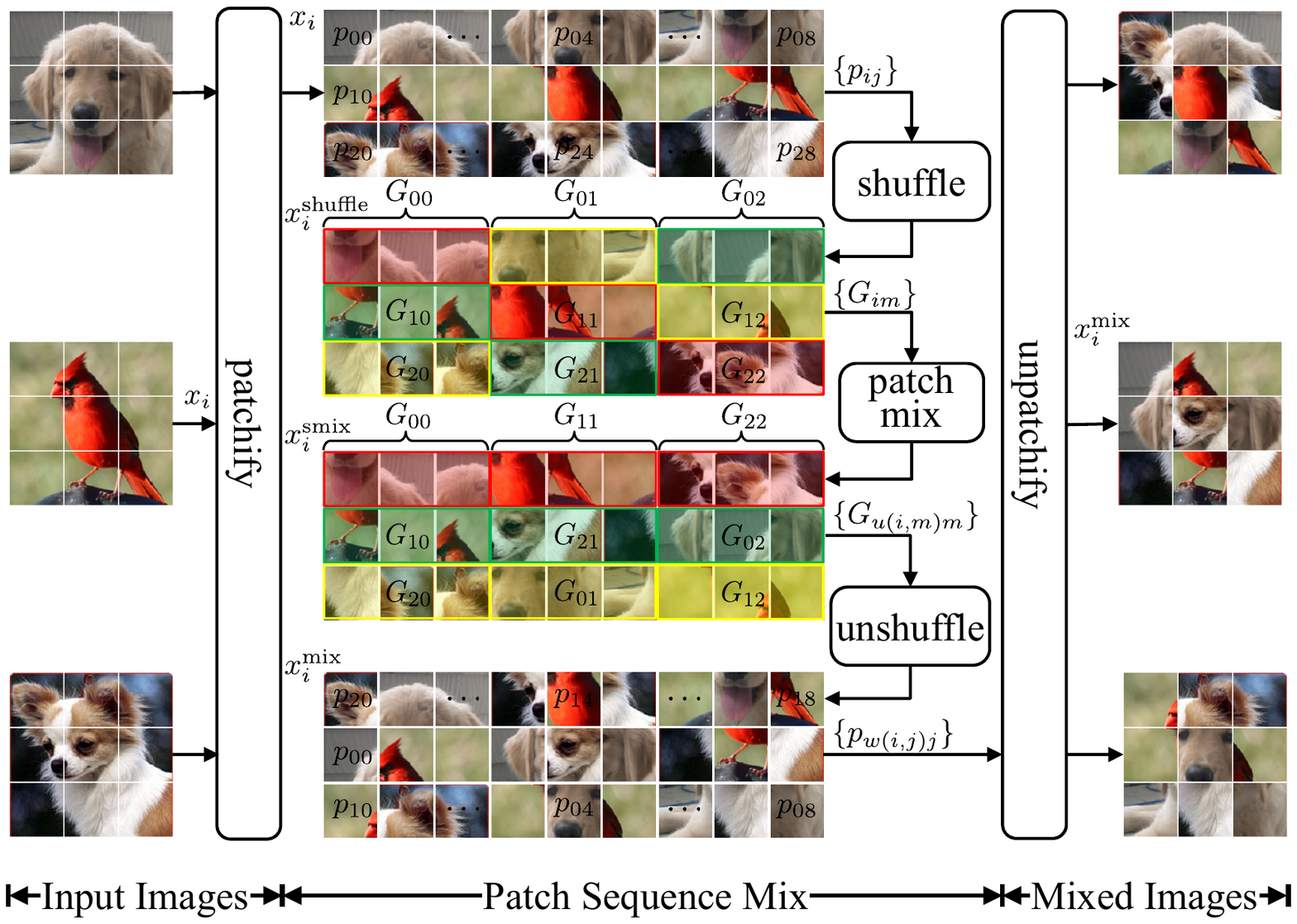}
   \caption{The overview of patch mix strategy. 
   First, the patch sequence $\sequence{p_{ij}}{j=0}{T-1}$ is shuffled into $x_i^{\rm shuffle}$ to conduct unbiased patch sampling.
   Second, the shuffled patch sequences are mixed in a group-wise fashion to reform novel mixed image $x_i^{\rm smix}$
   (e.g., patch group $\{G_{00}, G_{11}, G_{22}\}$ is combined into mixed image $x_0^{\rm smix}$.).
   Third, the patches in the shuffled mixed image $x_i^{\rm smix}$ are recovered into original positions to obtain the final mixed image $x_i^{\rm mix}$.
   }
   \label{fig:patchmix}
\end{figure}

To simulate rich inter-instance similarities in reality, we construct image instances and soft training targets by mixing multiple images in patch level. 
In this way, the mixed patch sequence can possess features from several instances and provide a more practical target for contrastive learning. 
The overview of PatchMix is shown in Figure~\ref{fig:patchmix}. 
For better illustration, the symbols used in this section are listed in Table~\ref{tab:symbol}. 
In spite of somewhat complicated description, our method provides a general-purpose but efficient solution to implement patch mix with any number of images. 

Let $x_i$ denote the $i$-th image instance in the image batch $x = \sequence{x_i}{i=0}{N-1}$,
\footnote{In this paper, ``$\{\cdot\}$'' denotes ordered sequence, instead of unordered set.}
where $N$ denotes batch size of $x$.
Without loss of generality, we patchify $x_i$ into an image patch sequence $\sequence{p_{ij}}{j=0}{T-1}$, which can be written as 
\begin{equation}
   \sequence{p_{ij}}{j=0}{T-1} = \patchify{x_i},
\end{equation}
where $p_{ij}$ denotes the $j$-th image patch from sequence and $T$ is the length of sequence.

To alleviate the bias of image patch mix, we conduct uniform sampling strategy on the patches of each image. 
So each patch of the original image has equal opportunity to be selected in the mixed one. 
For easier implementation, we replace uniform sampling with random shuffle operation. 
In this way, we obtain the indices of sampled patches by $\sequence{k(j)}{j=0}{T-1} = \shuffle{\sequence{j}{j=0}{T-1}}$.
Then the indices are used to obtain the uniformly sampled patches by
\begin{equation}\label{eq:shuffle}
   x_i^{\rm shuffle} = \shuffle{\sequence{p_{ij}}{j=0}{T-1}} = \sequence{p_{ik(j)}}{j=0}{T-1}.
\end{equation}

For convenience of patch mix, we divide the shuffled sequence $\sequence{p_{ik(j)}}{j=0}{T-1}$ into $M$ groups,  where $M$ denotes the number of images for patch mix. 
Each group is denoted as $G_{im} = \sequence{p_{ik(j)}}{j=m \cdot S}{(m + 1) \cdot S - 1}$, where $S = \left \lfloor T/M \right \rfloor$ denotes the number of patches in group $G_{im}$.
For brevity, the shuffled image $x_i^{\rm shuffle}$ can be also presented as 
\begin{equation}\label{eq:group}
   x_i^{\rm shuffle} = \sequence{p_{ik(j)}}{j=0}{T-1} = \sequence{G_{im}}{m=0}{M-1}.
\end{equation}

To mix patches from different images, we randomly replace the patches of image with the ones of other images in the same position.
Specifically, we rearrange the patches from different images to form mixed image by
\begin{equation}\label{eq:smix}
   x_i^{\rm smix} = \{G_{u(i,m)m}\}_{m=0}^{M-1}, 
\end{equation}
where $u(i,m) = (i+m) \bmod N$ conducts patch mix operation in image batch. 
\footnote{More explanations can be found in the supplementary material.\label{note}} 
For example, as shown in Figure~\ref{fig:patchmix}, $\{G_{00}, G_{11}, G_{22}\}$, $\{G_{10}, G_{21}, G_{02}\}$ and $\{G_{20}, G_{01}, G_{12}\}$ respectively reform the corresponding mixed image.

To recover the order of patch sequence, we introduce additional unshuffle operation. 
To this end, we compute the indices for the rearrangement of patch sequence by 
\begin{equation}
   \sequence{r(j)}{j=0}{T-1} = \argsort{\sequence{k(j)}{j=0}{T-1}},
\end{equation}
which obtains the element indices by sorting $\sequence{k(j)}{j=0}{T-1}$ in ascending order.
Hence, the unshuffle operation can be implemented by $\unshuffle{\cdot} = \sindex{\cdot}{\sequence{r(j)}{j=0}{T-1}}$, where $\sindex{\cdot}{\cdot}$ rearranges elements of the given sequence according to indices $\sequence{r(j)}{j=0}{T-1}$.

For convenience, we expand the patch group $G_{u(i,m)m}$ in the shuffled mixed image $x_i^{\rm smix}$ and simplify the Eq.~\ref{eq:smix} as
\begin{align}
   x_i^{\rm smix} 
   &= \sequence{\sequence{p_{u(i,m)k(j)}}{j=m \cdot S}{(m+1) \cdot S - 1}}{m=0}{M-1} \nonumber \\
   &= \sequence{p_{v(i,j)k(j)}}{j=0}{T-1},
\end{align}
where $v(i,j)$ denotes the index mapping function for simplification.
Then, we recover the order of patch indices by $\sequence{j}{j=0}{T-1} = \unshuffle{\sequence{k(j)}{j=0}{T-1}}$ and $w(i,j) = \unshuffle{v(i,j)}$.
Finally, we obtain the mixed image by
\begin{equation}\label{eq:unshuffle}
   x_i^{\rm mix} = \unshuffle{x_i^{\rm smix}} = \sequence{p_{w(i,j)j}}{j=0}{T-1}.
\end{equation}

\begin{figure*}[ht]
   \centering
   \includegraphics[width=\linewidth]{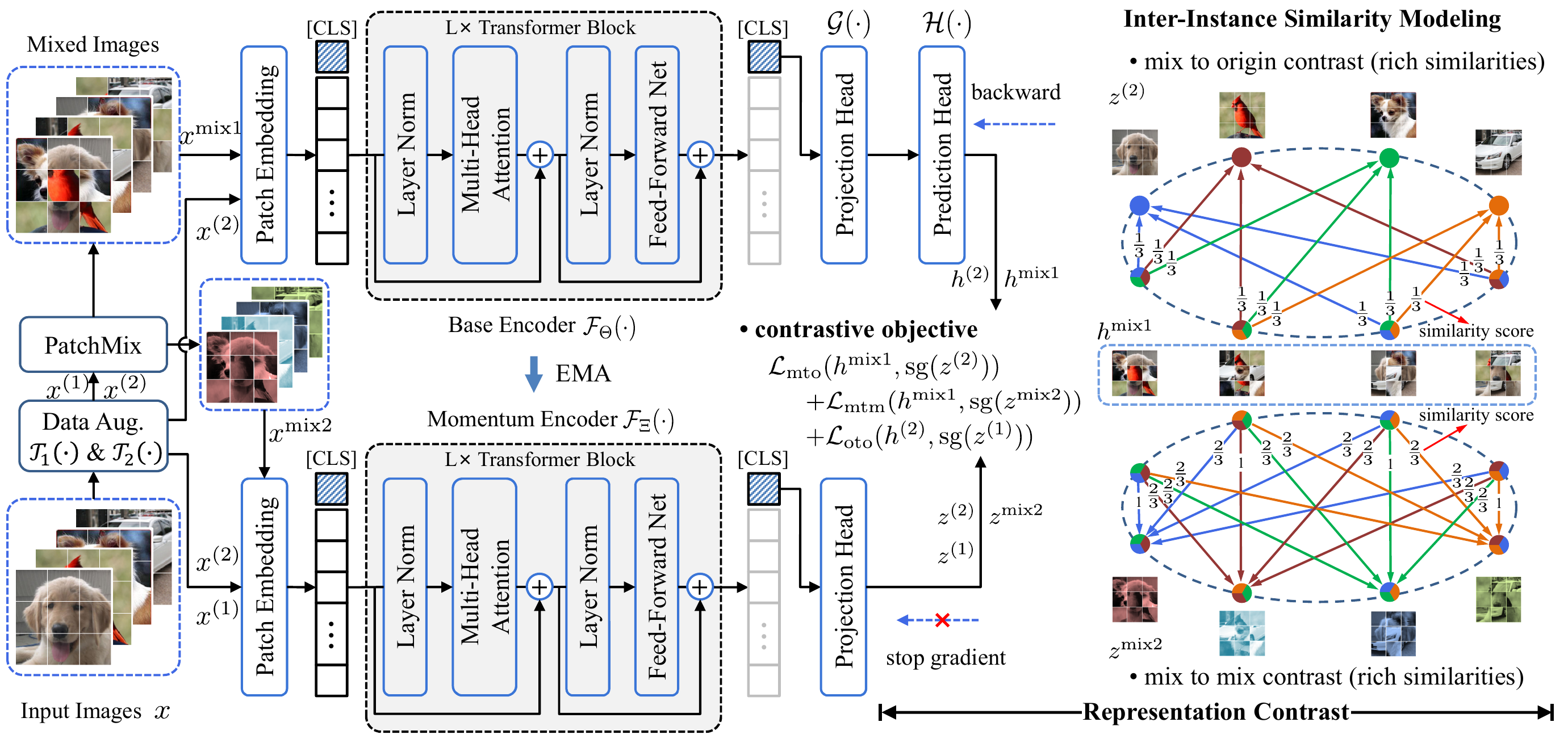}
   \caption{The framework of inter-instance similarity modeling. 
   The patch mix strategy is applied to image batch to construct positive pairs with rich instance similarities among images in the batch. 
   Then ViT model is trained by mix to mix contrastive objective and mix to original one, which guides the model to capture rich inter-instance similarities. 
   To reduce the potential representation gap between mixed images and original images, additional contrastive objective between original images is also applied.
   }
   \label{fig:model}
\end{figure*}

To implement the efficient of tensor operation, we flatten the indices of patch groups in image batch $x^{\rm shuffle}$ as 
\begin{equation}\label{eq:group_index}
   l = \sequence{\sequence{i \cdot M + m}{m=0}{M-1}}{i=0}{N-1}.
\end{equation}
The indices for patch mix operation are rewritten as 
\begin{equation}\label{eq:mix_index}
   q = (l + (l \bmod M) \cdot M) \bmod L , \textsuperscript{\ref{note}}   
\end{equation}
where $L = N \cdot M$ denotes the number of patches in image batch.
The patch mix operation for the shuffled image can be presented as $x^{\rm smix} = \sindex{x^{\rm shuffle}}{q}$ and the unshuffle operation as $x^{\rm mix} = \unshuffle{x^{\rm smix}}$.
The labels for mixed image batch $x^{\rm mix}$ with respect to original image batch $x$ can be obtained by 
\begin{equation}\label{eq:mix_target}
   y^{\rm mto} = \sequence{\sequence{u(i,m)}{m=0}{M-1}}{i=0}{N-1}.
\end{equation}
The labels for contrast between mixed image batch and mixed image batch can be written as 
\begin{equation}\label{eq:mix_mix}
   y^{\rm mtm} = \sequence{\sequence{j}{j=(i-M+1+N) \bmod N}{(i+M-1) \bmod N}}{i=0}{N-1}. \textsuperscript{\ref{note}}
\end{equation}
In other words, each image in mixed image batch has $(2M - 1)$ positive pairs in another augmented view.
Moreover, due to similarity difference between mixed images, the weights (similarity scores) for each item in label $y^{\rm mtm}$ can be written as
\begin{equation}\label{eq:mix_mix_weight}
   \omega^{\rm mtm} = \sequence{\sequence{1 - \left| M-j-1 \right| \big/ M }{j=0}{2M-2}}{i=0}{N-1}. \textsuperscript{\ref{note}}
\end{equation}
The algorithm of patch mix can be summarized as Algorithm~\ref{alg:patchmix}

\renewcommand{\algorithmicrequire}{\textbf{Input:}} 
\renewcommand{\algorithmicensure}{\textbf{Output:}}
\begin{algorithm}[!t]
   \caption{Patch Mix Strategy}
   \label{alg:patchmix}
   \begin{algorithmic}[1]
         \Require{Input image batch $x$; 
         batch size $N$; 
         the image number for mix $M$.}
         \Ensure{The mixed image batch $x^{\rm mix}$; 
         the mixed target batch $y^{mix}$.}
         \Function{patchmix}{Batch $x$} \algorithmiccomment{PatchMix operation}
            \State Obtain shuffled image batch by $x^{\rm shuffle}$ by Eq.~\ref{eq:shuffle};
            \State Divide into $M$ groups, {\small $x^{\rm shuffle} = \seqin{\seqin{G_{im}}{m=0}{M-1}}{i=0}{N-1}$};
            \State Flatten patch groups {\small $\seqin{\seqin{G_{im}}{m=0}{M-1}}{i=0}{N-1}$} as $\seqin{G_{l}}{l=0}{L-1}$;
            \State Obtain the indices for patch mix $q$ by Eq.~\ref{eq:mix_index};
            \State Conduct patch mix by $x^{\rm smix} = \sindex{x^{\rm shuffle}}{q}$;
            \State Recover the patch order $x^{\rm mix} = \unshuffle{x^{\rm smix}}$;
            \State Compute mixed targets $y^{\rm mix}$ by Eq.~\ref{eq:mix_target};
            \State \Return{$x^{\rm mix}$, $y^{\rm mix}$}
         \EndFunction
   \end{algorithmic}
\end{algorithm}

\subsection{Inter-Instance Similarity Modeling}

In this section, we apply the obtained mixed images and soft targets to guide the trained model to capture potential rich similarities among massive image instances. 

As shown in Figure~\ref{fig:model}, the input image batch $x$ are processed by two random data augmentations $\mathcal{T}_1(\cdot)$ and $\mathcal{T}_2(\cdot)$ to obtain transformed ones $x^{(1)} = \mathcal{T}_1(x)$ and $x^{(2)} = \mathcal{T}_2(x)$.
To construct similarity-rich positive pairs, patch mix strategy is respectively conducted on image batch $x^{(1)}$ and $x^{(2)}$ to obtain mixed image batch $x^{{\rm mix}1}$ and $x^{{\rm mix}2}$.

More inter-instance similarities introduced by patch mix is two fold.
(1) Different from the plain contrastive framework, mixed image $x_i^{\rm mix1}$ contains $M$ positive samples in image batch $x^{(2)}$, namely $\{x_{u(i,m)}^{(2)}\}_{m=0}^{M-1}$, each of which is partially similar to mixed image $x_i^{\rm mix1}$. 
The same case is also applicable to mixed image batch $x^{\rm mix2}$. 
We call this contrastive framework \emph{mix-to-origin contrast}.
(2) Mixed image $x_i^{\rm mix1}$ has $(2M-1)$ positive samples in mixed image batch $x^{\rm mix2}$, namely $\{x_j^{\rm mix2}\}_{j=(i-M+1+N) \bmod N}^{(i+M-1) \bmod N}$, which provides more inter-instance similarities than mix-to-origin contrast. 
We call this contrastive framework \emph{mix-to-mix contrast}.

To stabilize the training of vision Transformer model $\mathcal{F}_\Theta(\cdot)$, we introduce additional momentum encoder  $\mathcal{F}_\Xi(\cdot)$, whose parameters $\Xi$ are updated by the parameters of base encoder $\mathcal{F}_\Theta(\cdot)$ in an EMA (exponential moving average) manner.
To improve the transferability of contrastive representations, projection head $\mathcal{G}(\cdot)$ and prediction head $\mathcal{H}(\cdot)$~\cite{chen2021empirical,grill2020bootstrap,chen2021exploring} are also applied. 

Combining module $\mathcal{G}(\cdot)$ and $\mathcal{H}(\cdot)$, the image batch $x^{\rm mix1}$ is fed into model $\mathcal{F}_\Theta(\cdot)$ to obtain predicted representation $h^{\rm mix1} = \mathcal{H}(\mathcal{G}(\mathcal{F}_\Theta(x^{\rm mix1})))$.
Then, both $x^{\rm mix2}$ and $x^{(2)}$ are fed into momentum encoder $\mathcal{F}_\Xi(\cdot)$ to obtain projected representation $z^{\rm mix2} = \mathcal{G}(\mathcal{F}_\Xi(x^{\rm mix2}))$ and $z^{(2)} = \mathcal{G}(\mathcal{F}_\Xi(x^{(2)}))$.
To optimize the model, we introduce mix-to-origin contrastive objective as
\begin{small}
\begin{align}
   &\mathcal{L}_{\rm mto}(h^{\rm mix1}, z^{(2)}) \nonumber \\
   &= \!\! - \frac{1}{N \! \cdot \! M} \!\!
   \sum_{i=0}^{N-1}{ 
      \sum_{j=0}^{M-1}{\!\!
         \log{\!
            \frac{\expb{ \similar{h_{i}^{\rm mix1}}{z_{ y_{ij}^{\rm mto}}^{(2)}} / \tau }
            }
            {
               \sum_{t=0}^{N-1}{\!\!
                  \expb{\! \similar{\! h_{i}^{\rm mix1}}{z_{t}^{(2)}\!} / \tau \!}
               }
            }
         }
      }
   },
\end{align}
\end{small}
where $\similar{h}{z} = \frac{h^{\rm T}\cdot z}{\norm{h} \cdot \norm{z}}$ denotes cosine similarity between $h$ and $z$; $\tau$ denotes temperature coefficient.

Furthermore, mix-to-mix contrastive objective is used to model richer inter-instance similarity by
\begin{small}
\begin{align}
   &\mathcal{L}_{\rm mtm}(h^{\rm mix1}, z^{\rm mix2}) \nonumber \\
   &= \!\! - \frac{1}{N} \!\!
   \sum_{i=0}^{N-1}{
      \sum_{j=0}^{2M\!-\!2}{\!\!\!
         \omega^{\rm mtm}_{ij} \! \cdot \!
         \log{\!
            \frac{
               \expb{\! \similar{h_{i}^{\rm mix1}}{z_{ y_{ij}^{\rm mtm}}^{\rm mix2}} \! / \tau }
            }
            {
               \sum_{t=0}^{N-1}{\!\!
                  \expb{\! \similar{\! h_{i}^{\rm mix1}}{z_{t}^{\rm mix2}\!} / \tau \!}
               }
            }
         }
      }
   }.
\end{align}
\end{small}

Since there is potential domain gap between mixed images and original images, we further introduce plain contrastive objective between original images, which can be presented as
\begin{small}
\begin{align}
   \!\!\!\! \mathcal{L}_{\rm oto}(h^{(2)},\! z^{(1)}) \!\!
   = \!\! - \frac{1}{N} \!\!
   \sum_{i=0}^{N-1}{ \!\!
      \log{ \!\!
         \frac{\expb{ \similar{h_{i}^{(2)}}{z_{i}^{(1)}} / \tau }
         }
         {
            \sum_{t=0}^{N-1}{ \!\!
               \expb{ \similar{\!\! h_{i}^{(2)}\!}{z_{t}^{(1)}\!\!}\! / \! \tau \! }
            }
         }
      }
   },
\end{align}
\end{small}
where $h^{(2)} = \mathcal{H}(\mathcal{G}(\mathcal{F}_\Theta(x^{(2)})))$ and $z^{(1)} = \mathcal{G}(\mathcal{F}_\Xi(x^{(1)}))$.
To further alleviate representation degeneration, stop gradient operation $\operatorname{sg}(\cdot)$ is also applied to $z$ and the total contrastive objective can be presented as
\begin{small}
   \begin{align}
      \mathcal{L}_{\rm total} 
      & = \mathcal{L}_{\rm mto}(h^{\rm mix1},  \operatorname{sg}(z^{(2)})) 
       + \mathcal{L}_{\rm mtm}(h^{\rm mix1},  \operatorname{sg}(z^{\rm mix2})) \nonumber \\
      & + \mathcal{L}_{\rm oto}(h^{(2)}, \operatorname{sg}(z^{(1)})).
   \end{align}
\end{small}
Overall, our approach can be summarized as Algorithm~\ref{alg}.

\renewcommand{\algorithmicrequire}{\textbf{Input:}} 
\renewcommand{\algorithmicensure}{\textbf{Output:}}
\begin{algorithm}[!t]
   \caption{Inter-Instance Similarity Modeling}
   \label{alg}
   \begin{algorithmic}[1]
      \Require{Training dataset $\mathcal{X}$; 
      ViT model $\mathcal{F}_{*}$;
      projection head $\mathcal{G}$; 
      prediction head $\mathcal{H}$;
      momentum coefficient $\mu$.
      }
      \Ensure{The parameters $\Theta$ of ViT model $\mathcal{F}_{*}$}.
      \State Initialize the parameters of $\mathcal{F}_{*}$, $\mathcal{G}$ and $\mathcal{H}$;
      \State Initialize the momentum parameters $\Xi \leftarrow \Theta$;
      \For{batch $x$ in dataset $\mathcal{X}$ }
         \State Augment image batch $x^{(1)}, x^{(2)} = \mathcal{T}_1(x), \mathcal{T}_2(x)$;

         \State Obtain $x^{\rm mix1}, y^{\rm mto1} = \Call{patchmix}{x^{(1)}}$;
         \State Obtain $x^{\rm mix2}, y^{\rm mto2} = \Call{patchmix}{x^{(2)}}$;
         \State Obtain label $y^{\rm mtm}$ and weight $\omega^{\rm mtm}$ by Eq.~\ref{eq:mix_mix}\&\ref{eq:mix_mix_weight};

         \State $h^{\rm mix1}, h^{(2)} = \mathcal{H}(\!\mathcal{G}(\!\mathcal{F}_\Theta(\!x^{\rm mix1}\!)\!)\!), \mathcal{H}(\!\mathcal{G}(\!\mathcal{F}_\Theta(\!x^{(2)}\!)\!)\!)$;
         \State {\small $z^{(1)}, \!z^{(2)}, \!z^{\rm mix2} \!\! = \!\! \mathcal{G}(\!\mathcal{F}_\Xi(\!x^{(1)}\!)\!), \mathcal{G}(\!\mathcal{F}_\Xi(\!x^{(2)}\!)\!), \!\mathcal{G}(\!\mathcal{F}_\Xi(\!x^{\rm mix2}\!)\!)$};
         \State Compute loss $\mathcal{L}_{\rm total}$ and update the parameters $\Theta$;
         \State Update momentum params $\Xi \leftarrow \mu \cdot \Xi + (1 - \mu) \cdot \Theta$;
      \EndFor

   \end{algorithmic}
\end{algorithm}

\section{Experiments}\label{sec:exp}

In this section, we first introduce the experimental settings, including datasets, network architecture, optimization and evaluation. 
Then, we report the main experimental results of our proposed method on ImageNet-1K, CIFAR10 and CIFAR100 to evaluate its effectiveness. 
Finally, ablation study is conducted in detail to validate the effectiveness of each component in our approach.

\begin{table*}[ht]
   \centering
   {
      \begin{tabular}{lccccccc}
         \toprule[1pt]
         \textbf{Method} & \textbf{Backbone} & \textbf{\#Views} & \textbf{Batch size} & \textbf{\#Epochs} & \textbf{Finetune (\%)} & \textbf{Linear (\%)} & \textbf{kNN (\%)}\\ 
         \hline
         SimCLR~\cite{chen2020simple}     & ViT-S/16 & 4(224)       & 4096 & 300  & NA   & 69.0 & NA \\
         SwAV~\cite{caron2020unsupervised}& ViT-S/16 & 4(224)       & 4096 & 300  & NA   & 67.1 & NA \\
         BYOL~\cite{grill2020bootstrap}   & ViT-S/16 & 4(224)       & 4096 & 300  & NA   & 71.0 & NA \\
         MoCo v3~\cite{chen2021empirical} & ViT-S/16 & 4(224)       & 4096 & 300  & 81.4 & 72.5 & 67.8 \\
         DINO~\cite{caron2021emerging}    & ViT-S/16 & 4(224)       & 1024 & 300  & NA   & 72.5 & 67.9 \\ 
         DINO$^*$~\cite{caron2021emerging}& ViT-S/16 & 4(224)+8(96) & 1024 & 300  & NA   & 76.1 & 72.8 \\ 
         iBOT~\cite{zhou2022ibot}         & ViT-S/16 & 4(224)       & 1024 & 800 & NA & 76.2 & 72.4\\ 
         SDMP~\cite{ren2022simple}        & ViT-S/16 & 4(224)       & 1024 & 300 & 79.1 & 73.8 & NA\\
         \hline
         \rowcolor{cyan!20}
         \textbf{PatchMix (ours)}         & ViT-S/16 & 4(224)       & 1024 & 300 & \textbf{82.8} & \textbf{77.4} & \textbf{73.3} \\
         \rowcolor{cyan!20}
         \textbf{PatchMix (ours)}         & ViT-S/16 & 4(224)       & 1024 & 800 & \textbf{83.4} & \textbf{77.9} & \textbf{74.3} \\

         \bottomrule[1pt]
         SimCLR~\cite{chen2020simple}     & ViT-B/16 & 4(224)       & 4096 & 300 & NA   & 73.9 & NA \\
         SwAV~\cite{caron2020unsupervised}& ViT-B/16 & 4(224)       & 4096 & 300 & NA   & 71.6 & NA \\
         BYOL~\cite{grill2020bootstrap}   & ViT-B/16 & 4(224)       & 4096 & 300 & NA   & 73.9 & NA \\
         MoCo v3~\cite{chen2021empirical} & ViT-B/16 & 4(224)       & 4096 & 300 & 83.2 & 76.5 & 70.7 \\
         DINO~\cite{caron2021emerging}    & ViT-B/16 & 4(224)       & 1024 & 400 & NA   & 72.8 & 68.9\\ 
         DINO$^*$~\cite{caron2021emerging}& ViT-B/16 & 4(224)+8(96) & 1024 & 400  & 82.3 & 78.2 & 76.1\\ 
         iBOT~\cite{zhou2022ibot}         & ViT-B/16 & 4(224)  & 1024 & 400 & NA   & 76.0 & 71.2\\ 
         SDMP~\cite{ren2022simple}        & ViT-B/16 & 4(224)  & 1024 & 300 & NA   & 77.2 & NA \\
         \hline
         \rowcolor{cyan!20}
         \textbf{PatchMix (ours)}         & ViT-B/16 & 4(224) & 1024 & 300 & \textbf{84.1} & \textbf{80.2} & \textbf{76.2}\\
         
         \bottomrule[1pt]

      \end{tabular}
   }
   \caption{Performance comparison on ImageNet-1K dataset under finetune, linear and kNN evaluation protocols. 
   ``NA'' denotes that the result is not available in original paper.
   ``$^*$'' denotes the model pretrained with multi-crop strategy~\cite{caron2020unsupervised}. 
   ``4(224)+8(96)'' denotes 4 images with size $224 \times 224$ and 8 images with size $96 \times 96$. 
   ``\# Epochs'' denotes the number of pretraining epochs. 
   Both finetune accuracy and linear accuracy are evaluated by finetuing the pretrained model for 100 epochs.
   }
   \label{table:imagenet}
\end{table*}

\subsection{Experimental Settings}

\subsubsection{Datasets}
To validate the effectiveness of our proposed method, we evaluate the performance on small-scale datasets: CIFAR10 (60,000 colour images with 10 classes including 50,000 images as training set and 10,000 images as validation set) and CIFAR100~\cite{krizhevsky2009learning} (60,000 colour images with 100 classes including 50,000 images as training set and 10,000 images as validation set) and large-scale dataset, ImageNet-1K~\cite{ILSVRC15} (including 1.2 million images as training set and 50,000 images as validation set, from 1,000 categories). 
For CIFAR and ImageNet-1K, the size of image is set to $32 \times 32 $ and $224 \times 224$, respectively. 
Additionally, detailed data augmentation techniques for contrastive learning are depicted in the supplementary material. 

To validate the transferability of unsupervised representation, we also finetune the ImageNet-1K pretrained model on COCO~\cite{lin2014microsoft}, which contains 118,000 images as training set and 5,000 images as validation set. 
The images are padded as $1024 \times 1024$ size during training and testing.

\subsubsection{Networks and Optimization}
We adopt basic ViT~\cite{dosovitskiy2021vit} as backbone.
The patch sizes for patchify are respectively $2 \times 2$ and $16 \times 16$ for CIFAR and ImageNet-1K. 
Specifically, we evaluate our method on ViT-T/2\footnote{The details are clarified in the supplementary material}, ViT-S/2 and ViT-B/2 for CIFAR, then ViT-S/16 and ViT-B/16 for ImageNet-1K.
Following MoCo v3~\cite{chen2021empirical}, additional projection and prediction module are applied.

For ImageNet-1K, we pretrain the ViT-S/16 model with AdamW optimizer with learning rate $2 \times 10^{-3}$ for 300 epochs and 800 epochs. 
And we pretrain ViT-B/16 model with learning rate $3 \times 10^{-3}$ for 300 epochs.
For the pretraining of all models, the batch size is 1024. 
We conduct linear warmup on learning rate for 10 epochs and then follow a cosine learning rate decay schedule for the rest 260 epochs. 
The weight decay follows a cosine schedule from 0.04 to 0.4.
And momentum coefficient $\mu$ follows a cosine schedule from 0.996 to 1.
By default, the number of images for PatchMix is set to 3. 
The temperature $\tau$ is set to $0.2$.

For CIFAR datasets, the model is pretrained for 800 epochs, where 100 epochs for warmup.
The learning rates for ViT-S/2 and ViT-B/2 model are $1 \times 10^{-3}$ and $1.5 \times 10^{-3}$.
The batch size is 512. 
Other hyperparameters are consistent with ImageNet-1K.
For ViT-T/2, the initial learning rate is set to $4 \times 10^{-3}$ and the weight decay follows a cosine schedule from 0.02 to 0.2. 
Other hyperparameters are consistent with ImageNet-1K.

\subsubsection{Evaluation}
To evaluate the representation performance, we adopt three common evaluation protocols: finetune evaluation, linear evaluation and k-NN (k-nearest neighbor) evaluation. 
For finetune evaluation, we initialize the model with the pretrained weights and then adapt them to downstream tasks by finetuning.
This protocol can be flexibly applicable to various downstream tasks by transfer learning, such as objection detection, semantic segmentation and so on. 
For linear evaluation, we fix the pretrained model and feed the predicted representation to a linear classifier for image recognition.
This protocol evaluates the representation quality without changing the original representations, which avoids disturbance due to additional training. 
However, this protocol is not suitable for the evaluation of non-linear representations~\cite{bao2022beit,devlin2019bert,he2022masked}. 
For k-NN evaluation, it doesn't require any learnable parameters and provides more stable evaluation results. 
However, it is also limited to simple representation evaluation.
Hence, we report all three protocols in our experiments for a more comprehensive evaluation.
For the fairness of comparative experiments, all pretrained model are finetuned for 100 epochs under both finetune and linear evaluation protocols.

\subsection{Experimental Results}

\subsubsection{Image Classification on ImageNet-1K}
In this experiment, we pretrain the model on the training set of ImageNet-1K and then evaluate the pretrained model on the validation set of ImageNet-1K under the finetune, linear and kNN evaluation protocols. 

As shown in Table~\ref{table:imagenet}, our proposed PatchMix achieves 82.8\%, 77.4\% and 73.3\% accuracy with ViT-S/16 pretrained for 300 epochs under finetune, linear and kNN evaluation protocol, respectively. 
It significantly outperforms DINO counterpart by a large margin, even DINO with multi-crop strategy.
When pretrained for 800 epochs, our method with ViT-S/16 respectively achieves 83.4\%, 77.9\% and 74.3\% accuracy under finetune, linear and kNN evaluation protocol, substantially superior to iBOT counterpart. 
For ViT-B/16 pretrained for 300 epochs, our method respectively reaches 84.1\%, 80.2\% and 76.2\% under the finetune, linear and kNN evaluation protocol, which surpasses other self-supervised methods, including DINO pretrained for 400 epochs with multi-crop strategy.
Especially under linear evaluation protocol, the model pretrained by our method for 300 epochs outperforms the previous state-of-the-art: SDMP, by 3.0\% accuracy using ViT-B/16 backbone. 
The above experimental results strongly support the effectiveness of our proposed PatchMix on unsupervised representation learning. 
We believe that inter-instance similarity modeling introduced by our proposed PatchMix can effectively improve the representation generalization among different images.

\begin{table*}[ht]
   \centering
   \resizebox{\linewidth}{!}
   {
      \begin{tabular}{l|c|c|c|c|ccc|ccc}
         \hline
           \multirow{2}{*}{\textbf{Method}} & \multirow{2}{*}{\textbf{Backbone}} 
         & \multirow{2}{*}{\textbf{\#Epochs}} & \multirow{2}{*}{\textbf{Batch}}
         & \multirow{2}{*}{\textbf{\#FLOPs}} 
         & \multicolumn{3}{c|}{\textbf{CIFAR10}} & \multicolumn{3}{c}{\textbf{CIFAR100}}\\
         \cline{6-11} 
         &  &  &  &  & \textbf{Tune} & \textbf{Linear} & \textbf{kNN} & \textbf{Tune} & \textbf{Linear} & \textbf{kNN} \\ 
         \hline
         MoCo v3~\cite{chen2021empirical} & ViT-T/2 & 800 & 512 & 44.4G & 95.5 & 89.8 & 88.1 & 78.6 & 67.1 & 58.9 \\
         DINO~\cite{caron2021emerging}    & ViT-T/2 & 800 & 512 & 80.2G & 93.5 & 88.4 & 87.3 & 75.4 & 61.8 & 57.4 \\
         iBOT~\cite{zhou2022ibot}         & ViT-T/2 & 800 & 512 & 142.8G & 96.3 & 93.0 & 92.3 & 82.0 & 63.6 & 58.2 \\
         SDMP~\cite{ren2022simple}        & ViT-T/2 & 800 & 512 & 50.0G & 96.4 & 93.2 & 92.2 & 82.3 & 72.4 & 65.7 \\
         \hline
         \rowcolor{cyan!20}
         \textbf{PatchMix (ours)}         & ViT-T/2 & 800 & 512 & 50.0G & \textbf{97.5} & \textbf{94.4} & \textbf{92.9} & \textbf{84.9} & \textbf{74.7} & \textbf{68.8} \\
         \rowcolor{cyan!20}
         \textbf{PatchMix (ours)}         & ViT-T/2 & 1600 & 512 & 50.0G & \textbf{97.8} & \textbf{94.8} & \textbf{93.5} & \textbf{84.9} & \textbf{76.0} & \textbf{70.1} \\
         \hline
         MoCo v3~\cite{chen2021empirical} & ViT-S/2 & 800 & 512 & 175.6G & 95.8 & 90.2 & 89.1 & 78.8 & 66.4 & 62.3 \\
         DINO~\cite{caron2021emerging}    & ViT-S/2 & 800 & 512 & 311.4G & 96.3 & 92.6 & 91.8 & 75.9 & 63.8 & 60.6 \\
         iBOT~\cite{zhou2022ibot}         & ViT-S/2 & 800 & 512 & 571.4G & 97.0 & 94.8 & 93.1 & 82.8 & 67.8 & 64.2 \\
         SDMP~\cite{ren2022simple}        & ViT-S/2 & 800 & 512 & 197.6G & 96.9 & 94.2 & 92.1 & 85.0 & 77.1 & 66.7 \\
         \hline
         \rowcolor{cyan!20}
         \textbf{PatchMix (ours)}         & ViT-S/2 & 800 & 512 & 197.6G & \textbf{98.1} & \textbf{96.0} & \textbf{94.6} & \textbf{86.0} & \textbf{78.7} & \textbf{75.4} \\
         \rowcolor{cyan!20}
         \textbf{PatchMix (ours)}         & ViT-S/2 & 1600 & 512 & 197.6G & \textbf{98.2} & \textbf{96.4} & \textbf{95.1} & \textbf{86.1} & \textbf{78.9} & \textbf{74.6} \\
         \hline
         MoCo v3~\cite{chen2021empirical} & ViT-B/2 & 800 & 512 & 700.0G  & 96.1 & 90.9 & 89.5 & 79.5 & 67.3 & 63.4 \\
         DINO~\cite{caron2021emerging}    & ViT-B/2 & 800 & 512 & 1237.2G & 96.8 & 92.8 & 92.1 & 76.4 & 65.6 & 62.7 \\
         iBOT~\cite{zhou2022ibot}         & ViT-B/2 & 800 & 512 & 2014.1G & 97.3 & 94.9 & 93.2 & 83.3 & 68.7 & 65.5 \\
         SDMP~\cite{ren2022simple}        & ViT-B/2 & 800 & 512 & 787.3G  & 97.2 & 94.3 & 92.4 & 85.1 & 77.3 & 68.3 \\
         \hline
         \rowcolor{cyan!20}
         \textbf{PatchMix (ours)}         & ViT-B/2 & 800 & 512 & 787.3G & \textbf{98.3} & \textbf{96.6} & \textbf{95.8} & \textbf{86.0} & \textbf{79.7} & \textbf{75.7} \\
         \hline
      \end{tabular}
   }
   \caption{Performance (\%) comparison on CIFAR datasets using finetune, linear and kNN evaluation protocols, respectively.
   ``Tune'' denotes classification accuracy using finetune protocol. 
   ``\# FLOPs'' denotes the number of floating-point operations per iteration during training, where batch size is 2. 
   ``\# Epochs'' denotes the number of pretraining epochs. 
   Both finetune accuracy and linear accuracy are evaluated by finetuing the pretrained model for 100 epochs.
   }
   \label{table:cifar}
\end{table*}

\subsubsection{Image Classification on CIFAR10 and CIFAR100}
In this experiment, we respectively pretrain the model on the training set of CIFAR10 and CIFAR100, and then respectively evaluate the corresponding pretrained model on the validation set of CIFAR10 and CIFAR100, under the finetune, linear and kNN evaluation protocols.

As shown in Table~\ref{table:cifar}, under all three evaluation protocols, our approach consistently outperforms the previous contrastive methods by a significant margin on both CIFAR10 and CIFAR100. 
Specially under kNN evaluation protocol, our PatchMix achieves significantly large improvement over the existing contrastive methods, e.g. 8.7\% and 5.1\% kNN accuracy improvement on CIFAR100 using ViT-S/2 and ViT-B/2, respectively. 
The reason is analyzed as follows. 
Since kNN protocol doesn't introduce any additional learnable parameters, the evaluation results reflect, to some degree, the intrinsic quality of representations. 
The superiority of our PatchMix, better inter-instance similarity modeling, is fully presented. 
Moreover, in spite of data-hungry ViT architecture, our powerful capacity for inter-instance similarity modeling significantly reduces the risk of overfitting and achieves excellent performance on small datasets, CIFAR10 and CIFAR100.

\subsection{Transfer Learning on Downstream Tasks}

To further evaluate the transferability of our proposed PatchMix, we conduct transfer learning experiments on downstream tasks: object detection and instance segmentation on COCO~\cite{lin2014microsoft} by Mask RCNN~\cite{he2017mask} with FPN~\cite{lin2017feature} as MAE~\cite{he2022masked}. 
The results are reported in Table~\ref{table:coco}.
With ViT-B/16 backbone pretrained on ImageNet-1K for 300 epochs, our PatchMix achieves 51.9 on AP\textsuperscript{box} and 46.1 on AP\textsuperscript{mask}, surpassing the previous state-of-the-art iBOT by 0.7 on AP\textsuperscript{box} and 1.9 on AP\textsuperscript{mask}. 
Meanwhile, compared to the previous leading self-supervised learning methods, MAE and SIM, our method requires significantly fewer pretraining epochs but achieves significantly transfer learning performance on object detection and instance segmentation task. 
The experimental results demonstrate that our PatchMix can effectively model image semantics and structure. 
We believe that our method can effectively balance the global and local features under the guidance of patch-mixed images. 

\begin{table}[!htp]
   \centering
   {
      \begin{tabular}{lccc}
         \toprule[1pt]
         \textbf{Method}                  & \textbf{\#Pre-Epochs} & \textbf{AP\textsuperscript{\rm box}} & \textbf{AP\textsuperscript{\rm mask}}\\ 
         \hline
         MoCo v3~\cite{chen2021empirical} & 300                   & 47.9       & 42.7 \\
         \hline
         MAE~\cite{he2022masked}          & 1600                  & 50.3       & 44.9 \\
         \hline
         CAE~\cite{chen2022context}       & 300                   & 48         & 42.3 \\
         \hline
         PeCo~\cite{dong2021peco}         & 300                   & 43.9       & 39.8 \\
         \hline
         SIM~\cite{tao2022siamese}        & 1600                  & 49.1       & 43.8 \\
         \hline
         \rowcolor{cyan!20}
         \textbf{PatchMix (ours)}         & 300                   & \textbf{51.9}    & \textbf{46.1} \\
         
         \bottomrule[1pt]

      \end{tabular}
   }
   \caption{Transfer learning performance on object detection and instance segmentation task using Mask RCNN with ViT-B/16 backbone (pretrained on ImageNet-1K for 300 epochs) on COCO dataset. 
   All pretrained models are finetuned on COCO dataset for 100 epochs.
   }
   \label{table:coco}
\end{table}

\subsection{Ablation Study}
To evaluate the effectiveness of each module in our method, we implement ablation study on CIFAR10 and CIFAR100 datasets using ViT-S/2 backbone. 
In this section, we mainly analyze the effect of our proposed patch mix, the image number for patch mix, the loss function items and the number of pretraining epochs.

\subsubsection{The Effect of Patch Mix}

To validate the effectiveness of our proposed patch mix strategy, we compare it with popular image mix methods applied on contrastive learning.
As shown in Table~\ref{table:mix}, image mix strategies, including CutMix+Mixup~\cite{zhang2017mixup}, RegionSwap~\cite{xu2022regioncl} and ResizeMix~\cite{ren2022simple}, indeed improve the representation quality of contrastive learning. 
Notably, our proposed PatchMix achieves $96.0\%$ linear accuracy and $94.6\%$ kNN accuracy on CIFAR10, $78.7\%$ linear accuracy and $75.4$ kNN accuracy on CIFAR100, consistently outperforms other image mix strategies by a significantly large margin, and effectively boosts the performance of contrastive representations. 
We owe the improvement to more complicated inter-instance similarity modeling capacity introduced by our PatchMix strategy.

\begin{table}[ht]
   \centering
   \resizebox{\linewidth}{!}
   {
      \begin{tabular}{l|cc|cc}
         \hline
           \multirow{2}{*}{\textbf{Method}}
         & \multicolumn{2}{c|}{\textbf{CIFAR10}} 
         & \multicolumn{2}{c}{\textbf{CIFAR100}}\\
         \cline{2-5} 
         & \textbf{Linear} & \textbf{kNN} & \textbf{Linear} & \textbf{kNN} \\
         \hline
         Baseline                           & 89.8 & 88.2 & 66.0 & 61.2 \\
         CutMix+Mixup~\cite{zhang2017mixup} & 90.6 & 88.7 & 66.8 & 62.3 \\
         RegionSwap~\cite{xu2022regioncl}   & 92.1 & 91.3 & 70.4 & 64.2 \\
         ResizeMix~\cite{ren2022simple}    & 94.2 & 92.1 & 77.1 & 66.7 \\
         \textbf{PatchMix (ours)}           & \textbf{96.0} & \textbf{94.6} & \textbf{78.7} & \textbf{75.4} \\
         \hline
      \end{tabular}
   }
   \caption{The effect of different mix strategies on CIFAR10 and CIFAR100 datasets. 
   The performance (\%) is measured by linear and kNN evaluation protocols.
   }
   \label{table:mix}
\end{table}

\subsubsection{The Effect of Image Number for Patch Mix}

To determine the best image number for patch mix, we evaluate the representation performance on CIFAR datasets under different image numbers for patch mix.
In Table~\ref{table:num_mix}, our method achieves the best performance under both linear and kNN evaluation protocols when the number of images for patch mix is 3.
More or fewer image number for patch mix doesn't introduce additional performance gain. 
We explain this result as follows.
Due to more images for patch mix, the number of patches from the same one image is significantly reduced.
Hence, the informative patches in mixed images also decrease, making the corresponding targets not so accurate to boost performance. 
Meanwhile, fewer images for patch mix can't well establish inter-instance similarities among natural images.

\begin{table}[ht]
   \centering
   {
      \begin{tabular}{c|cc|cc}
         \hline
           \multirow{2}{*}{\textbf{\#Mix} ($M$)}
         & \multicolumn{2}{c|}{\textbf{CIFAR10}} 
         & \multicolumn{2}{c}{\textbf{CIFAR100}}\\
         \cline{2-5} 
         & \textbf{Linear} & \textbf{kNN} & \textbf{Linear} & \textbf{kNN} \\
         \hline
         1  & 89.8 & 88.2 & 66.0 & 61.2 \\
         2  & 95.6 & 94.3 & 77.6 & 71.1 \\
         3  & \textbf{96.0} & \textbf{94.6} & \textbf{78.7} & \textbf{75.4} \\
         4  & 95.7 & 94.3 & 75.2 & 70.9 \\
         \hline
      \end{tabular}
   }
   \caption{The effect of image number for PatchMix on CIFAR10 and CIFAR100 datasets.
   The performance (\%) is measured by linear and kNN evaluation protocols.
   }
   \label{table:num_mix}
\end{table}

\begin{figure*}[t]
   \renewcommand\arraystretch{0}
   \renewcommand\tabcolsep{1pt}
   \resizebox{\linewidth}{!}
   {
   \begin{tabular}{m{3.5cm}<{\centering} m{3.5cm}<{\centering} m{5cm}<{\centering} m{5cm}<{\centering} m{5cm}<{\centering}}
      \textbf{Query Sample} & \textbf{Key Samples}
      & \textbf{DINO} & \textbf{SDMP} 
       & \textbf{PatchMix (ours)} \\

      \includegraphics[width=\linewidth]{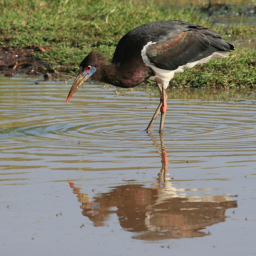}
      & \includegraphics[width=\linewidth]{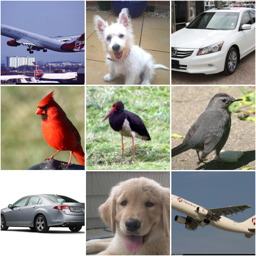}
      & \includegraphics[width=\linewidth]{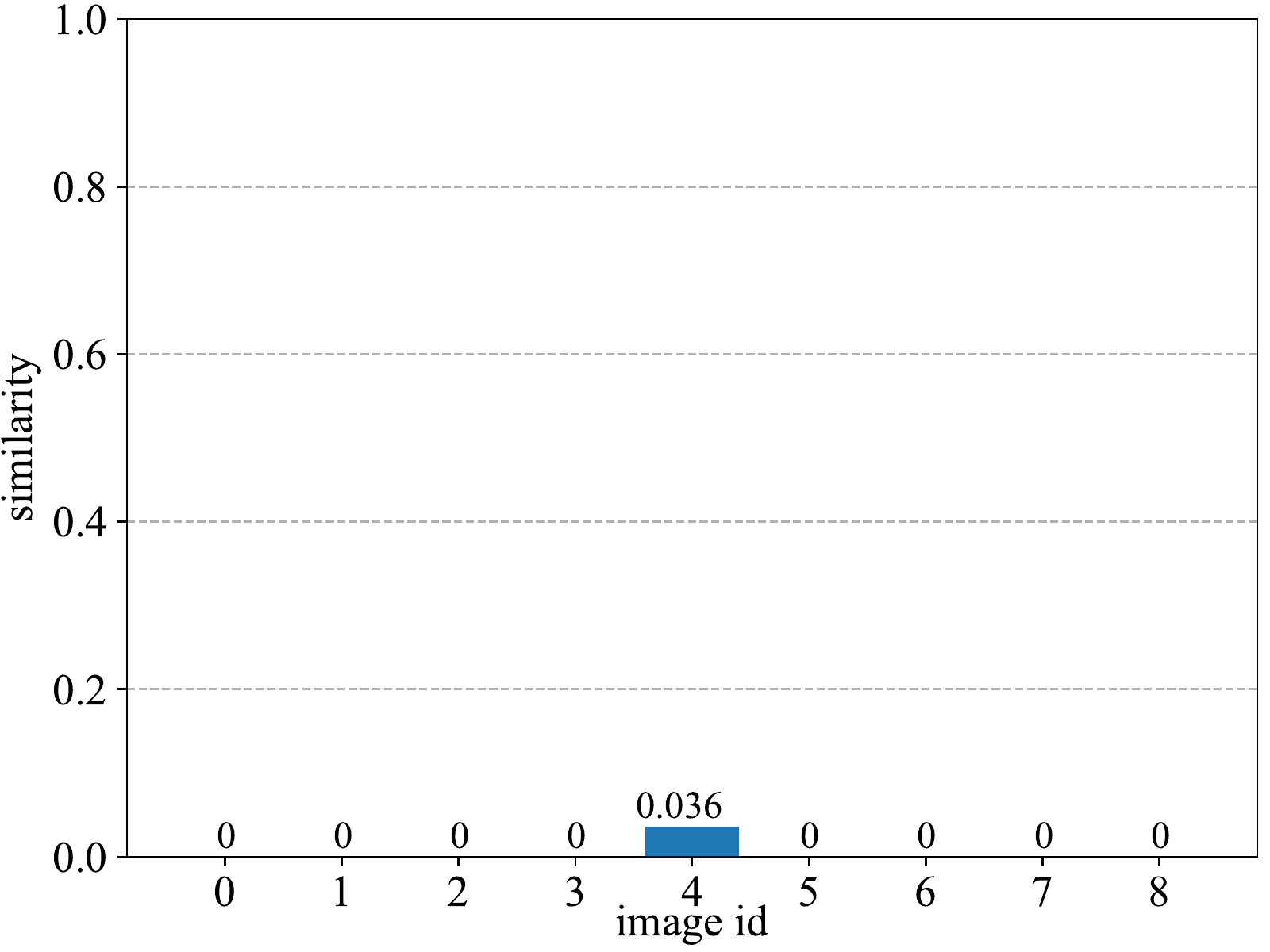}
      & \includegraphics[width=\linewidth]{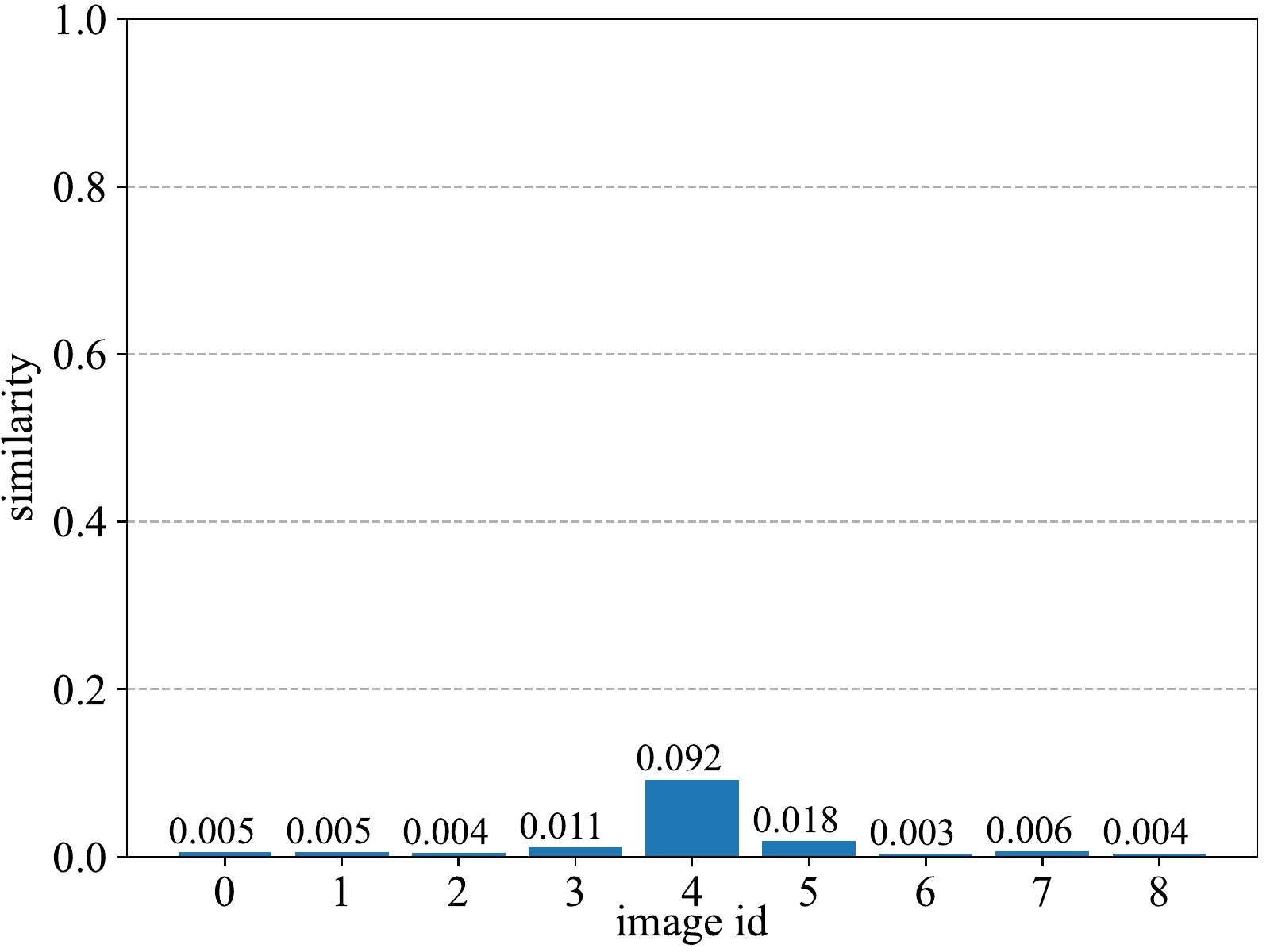}
      & \includegraphics[width=\linewidth]{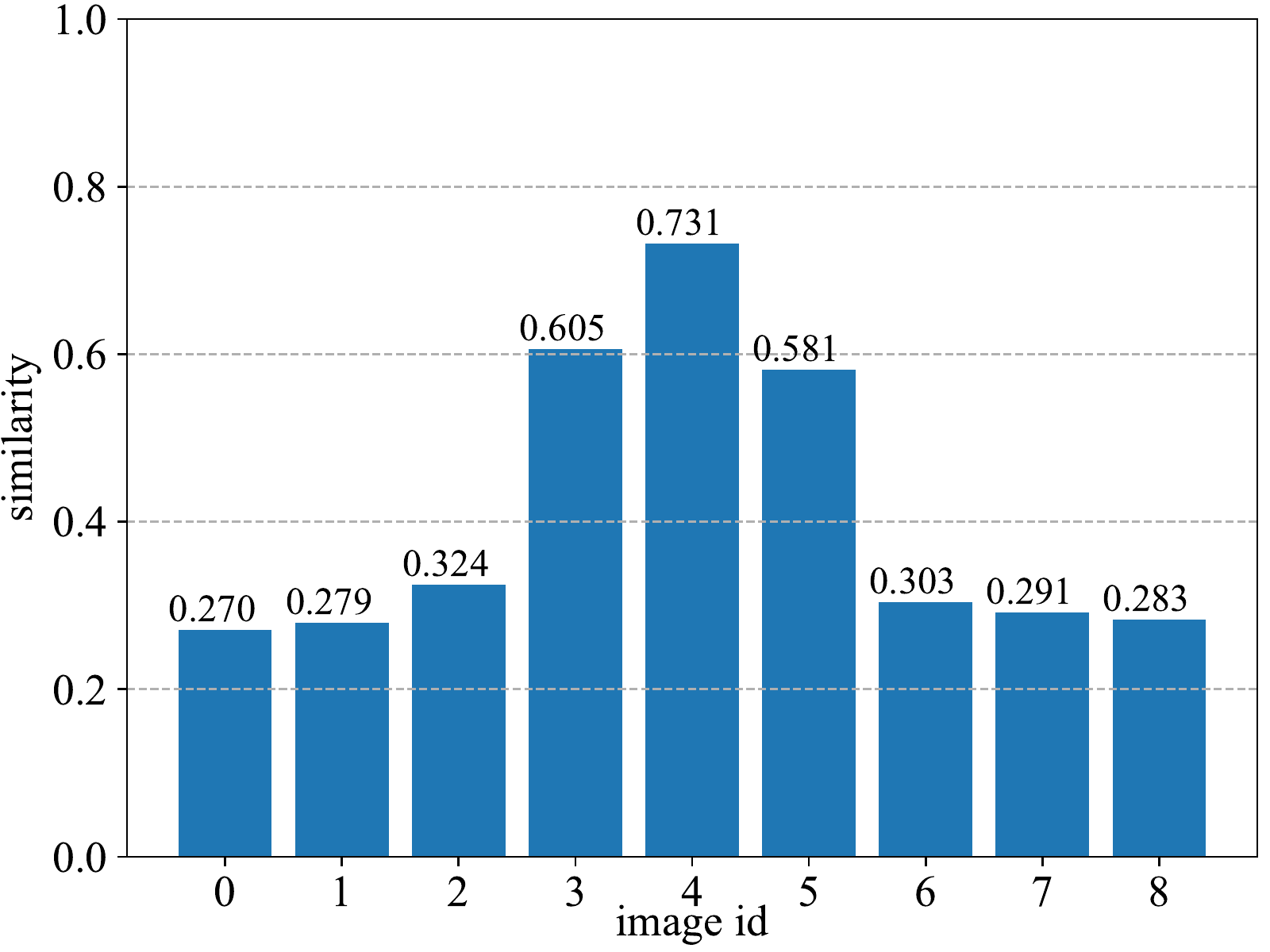} \\


      \includegraphics[width=\linewidth]{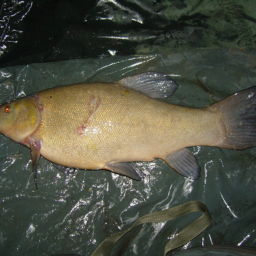}
      & \includegraphics[width=1\linewidth]{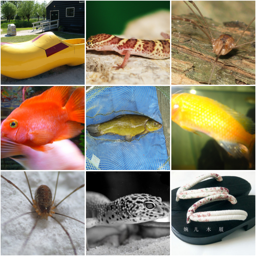}
      & \includegraphics[width=1\linewidth]{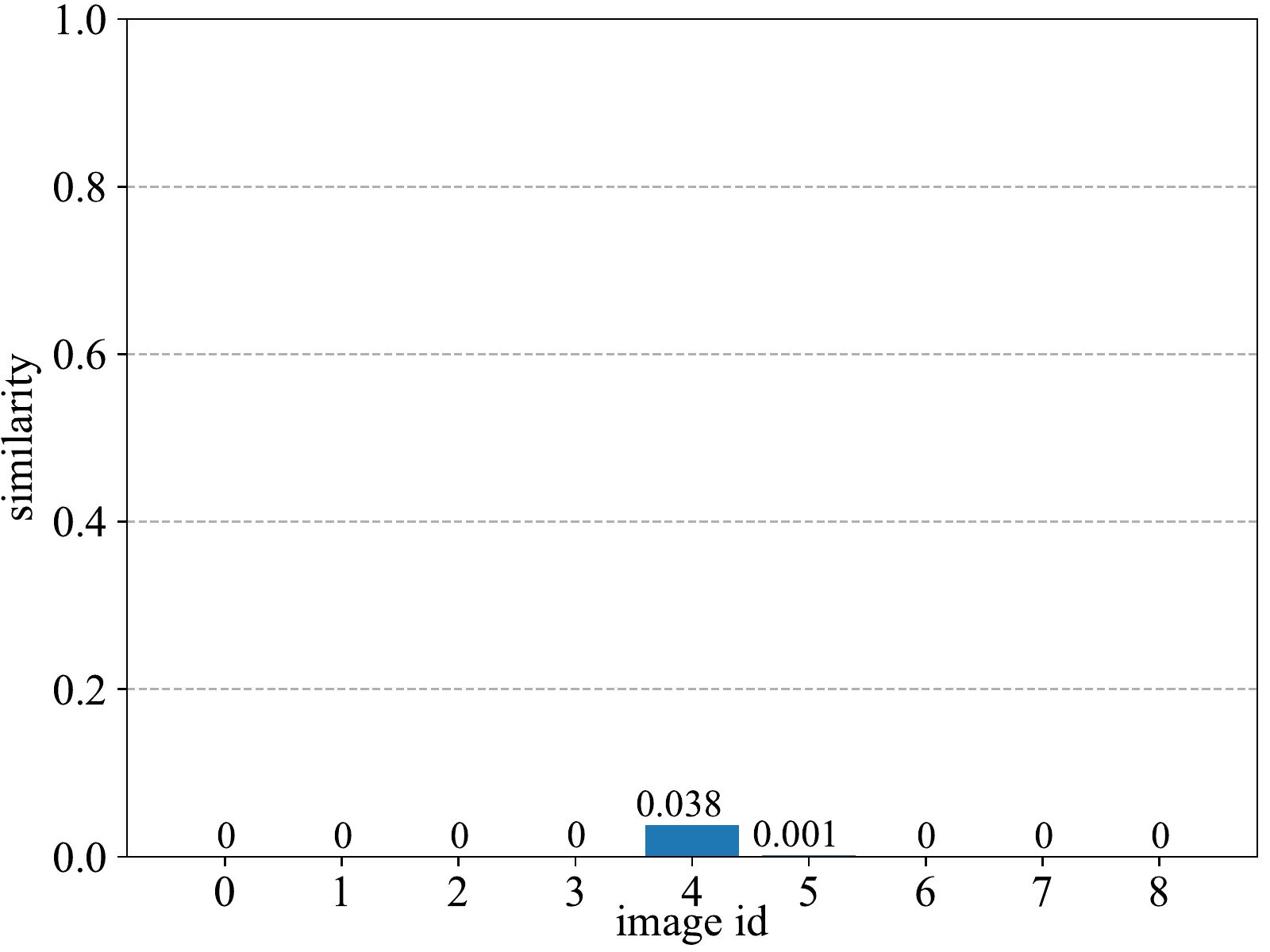}
      & \includegraphics[width=\linewidth]{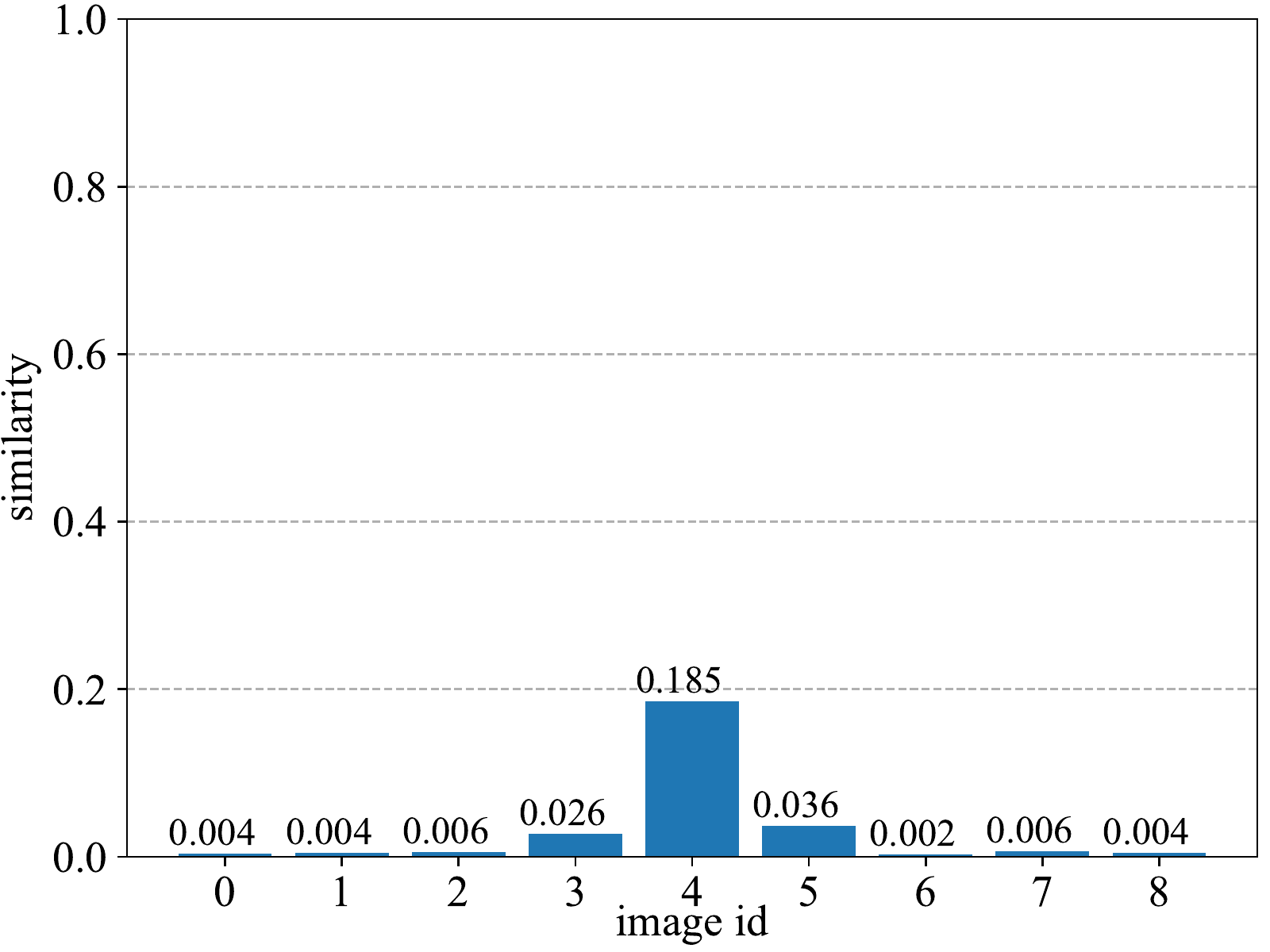}
      & \includegraphics[width=1\linewidth]{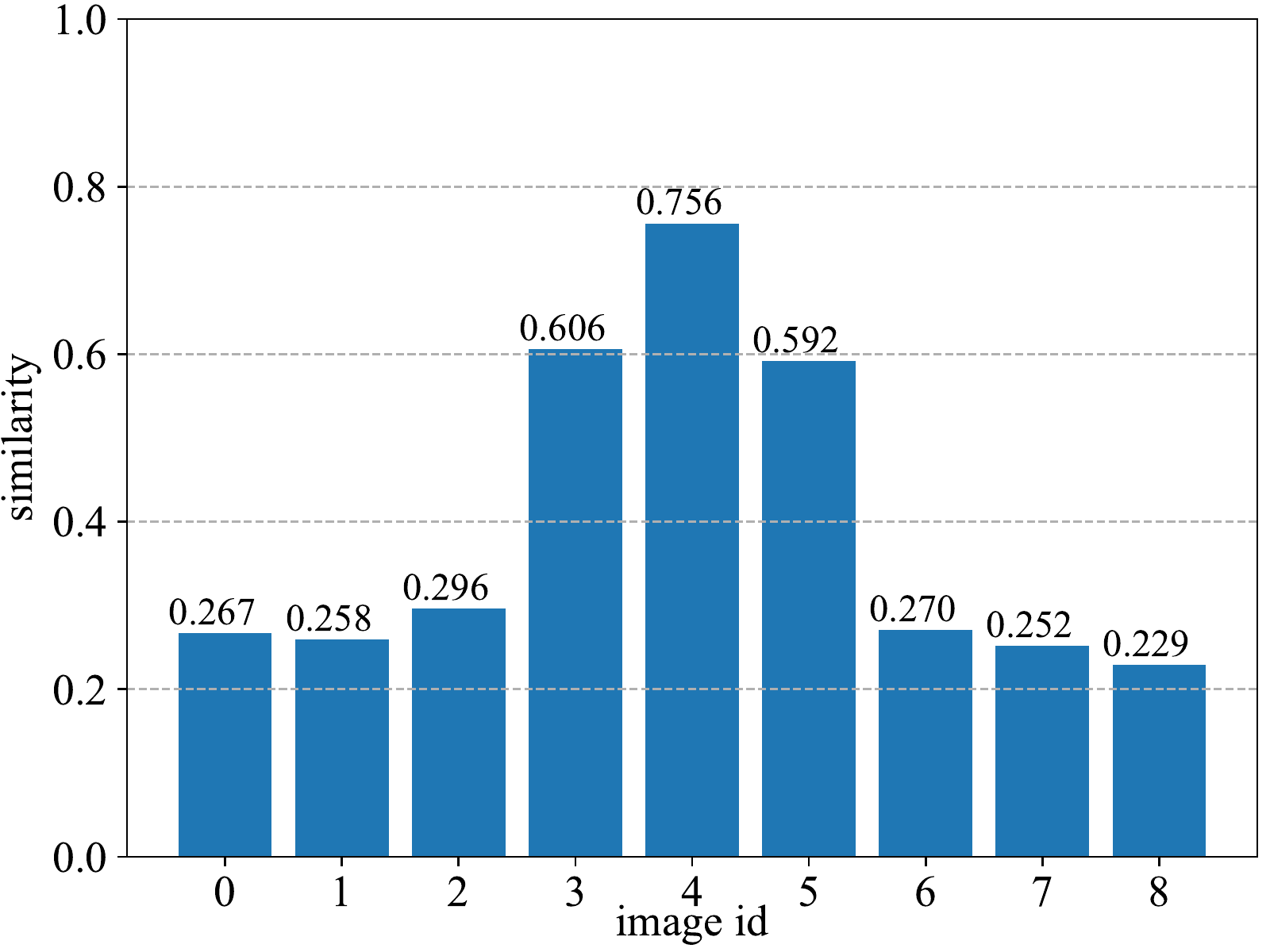} \\


      \includegraphics[width=\linewidth]{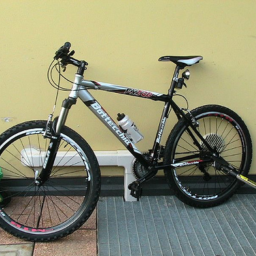}
      & \includegraphics[width=\linewidth]{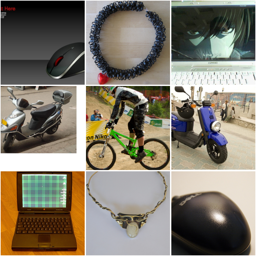}
      & \includegraphics[width=\linewidth]{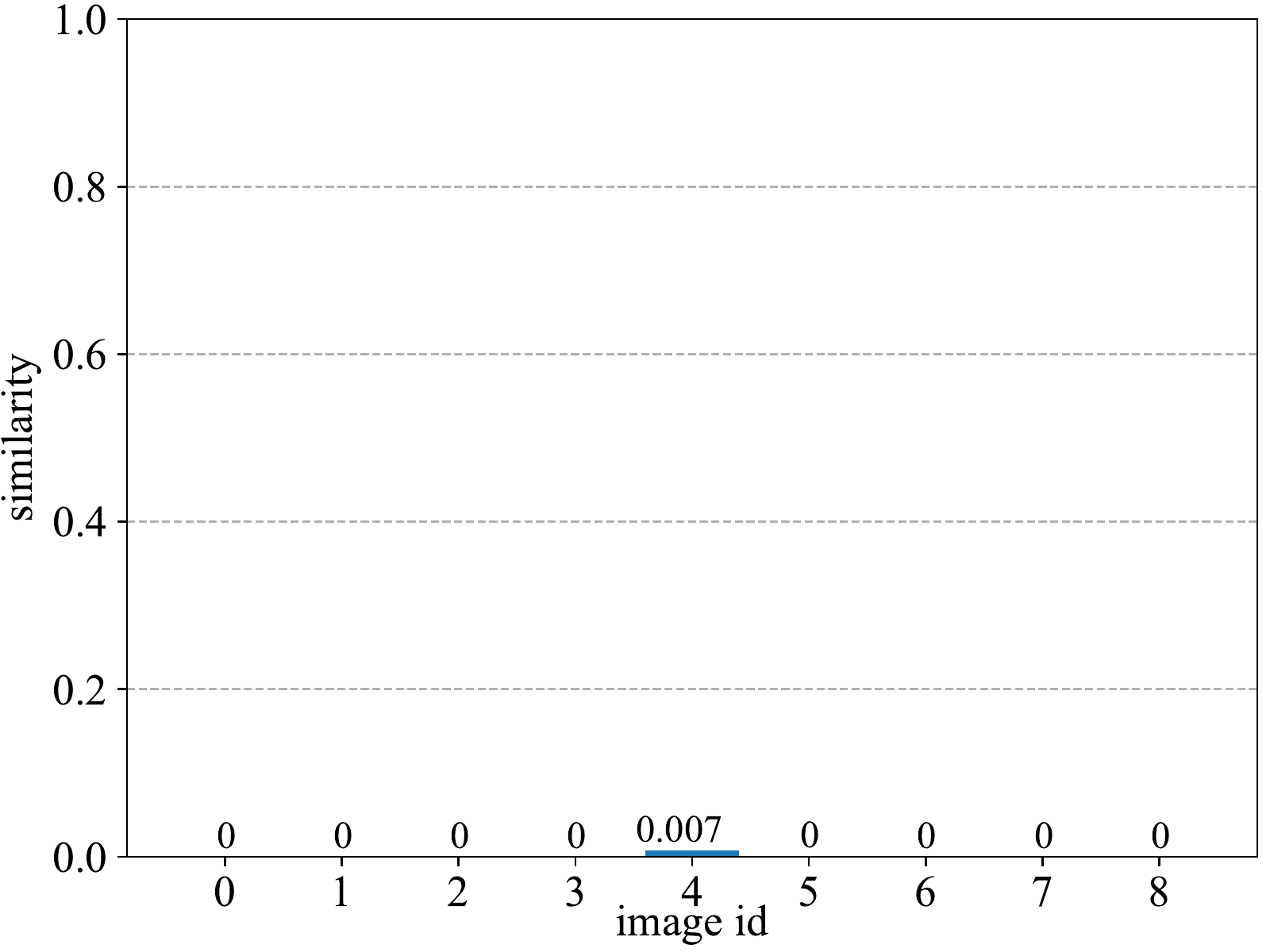}
      & \includegraphics[width=\linewidth]{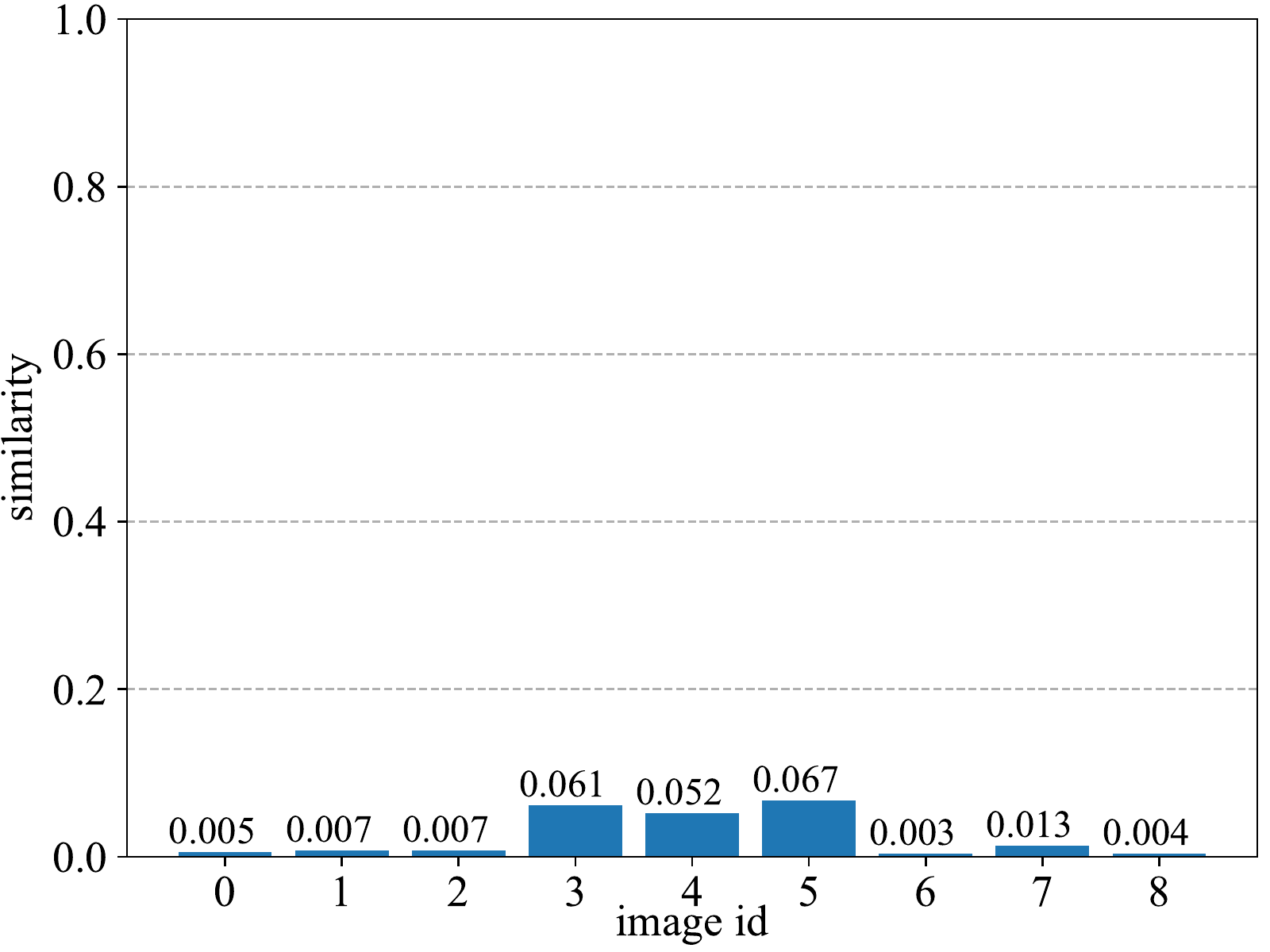}
      & \includegraphics[width=\linewidth]{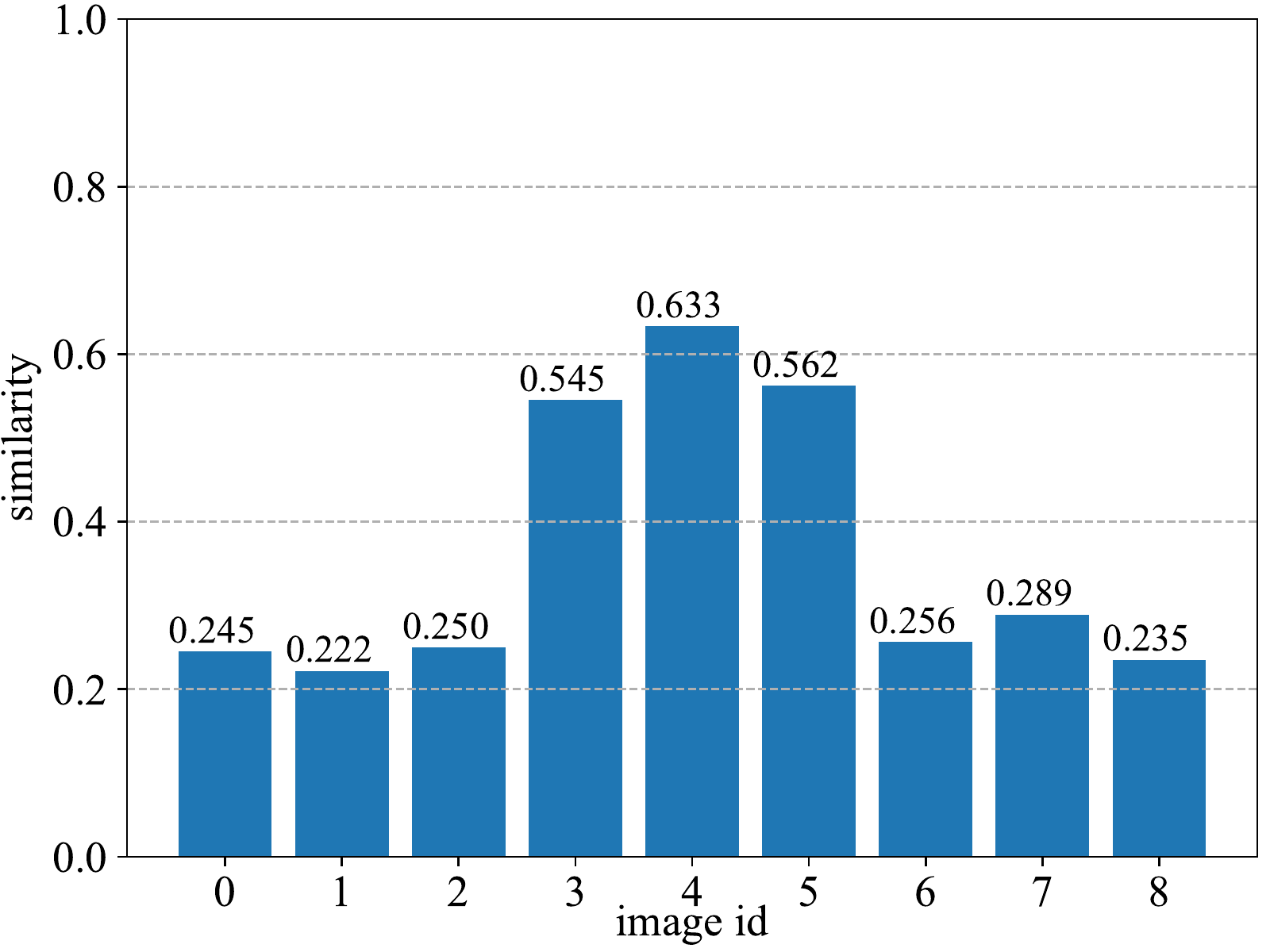} \\

   \end{tabular}
   }
   \caption{The visualization of inter-instance similarities using ViT-S/16 pretrained on ImageNet-1K dataset for 300 epochs.
   The query sample and the image with id 4 in key samples are from the same category. 
   The images with id 3 and 5 come from category similar to query sample. 
   We compare inter-instance similarities constructed by our method with two representative methods, DINO and SDMP.
   }
   \label{fig:visualization}
\end{figure*}

\subsubsection{The Effect of Loss Function}

To investigate the effectiveness of loss items in our contrastive objective, we evaluate the performance as single loss item or their combination is applied. 
The results are presented in Table~\ref{table:loss}.
First, we can observe that simply utilizing single loss item can not achieve excellent performance.
Second, when original images to original ones contrast item $\mathcal{L}_{\rm oto}$ and mixed images to original ones contrast item $\mathcal{L}_{\rm mto}$ are jointly applied, the representation performance is significantly improved, e.g. from 59.7\% to 69.2\% on CIFAR100 under the kNN evaluation protocol. 
Third, combing $\mathcal{L}_{\rm oto}$, $\mathcal{L}_{\rm mto}$ and $\mathcal{L}_{\rm mtm}$ achieves the best performance, surpassing the baseline with only $\mathcal{L}_{\rm oto}$ by 4.9\% linear accuracy and 7.6\% kNN accuracy improvement on CIFAR10, 9.0\% linear accuracy and 15.7\% kNN accuracy improvement on CIFAR100. 
We analyze the reason as follows. 
The item $\mathcal{L}_{\rm oto}$, $\mathcal{L}_{\rm mto}$ and $\mathcal{L}_{\rm mtm}$ respectively establishes image relations between natural images and natural ones, mixed images and natural ones, mixed images and mixed one. 
As all above items work, the trained model can well balance the capacity of natural image representations and multi-image relation modeling to achieve high-quality unsupervised representation performance. 

\begin{table}[ht]
   \centering
   \resizebox{\linewidth}{!}
   {
      \begin{tabular}{ccc|cc|cc}
         \hline
           \multirow{2}{*}{$\mathcal{L}_{\rm oto}$} 
         & \multirow{2}{*}{$\mathcal{L}_{\rm mto}$} 
         & \multirow{2}{*}{$\mathcal{L}_{\rm mtm}$}
         & \multicolumn{2}{c|}{\textbf{CIFAR10}} 
         & \multicolumn{2}{c}{\textbf{CIFAR100}} \\
         \cline{4-7} 
         & & & Linear & kNN & Linear & kNN \\
         \hline
         \checkmark &     -      &     -      & 91.1          & 87.0          & 69.7          & 59.7 \\
             -      & \checkmark &     -      & NA            & NA            & 61.3          & 57.2 \\
             -      &     -      & \checkmark & NA            & NA            & 52.6          & 47.4 \\
         \checkmark & \checkmark &     -      & 95.4          & 93.8          & 77.9          & 69.2 \\
         \checkmark & \checkmark & \checkmark & \textbf{96.0} & \textbf{94.6} & \textbf{78.7} & \textbf{75.4}\\
         \hline
      \end{tabular}
   }
   \caption{The effect of loss items on CIFAR10 and CIFAR100 datasets.
   The performance (\%) is measured by linear and kNN evaluation protocols.
   ``NA'' indicates that the model fails to converge. 
   }
   \label{table:loss}
\end{table}

\subsubsection{The Effect of Pretraining Epochs}

\begin{figure*}[ht]
   \setlength\tabcolsep{1pt}
   \begin{tabular}{cccc}
        \rotatebox{90}{{\scriptsize \qquad \qquad \qquad CIFAR10}} 
      & \includegraphics[width=0.33\linewidth]{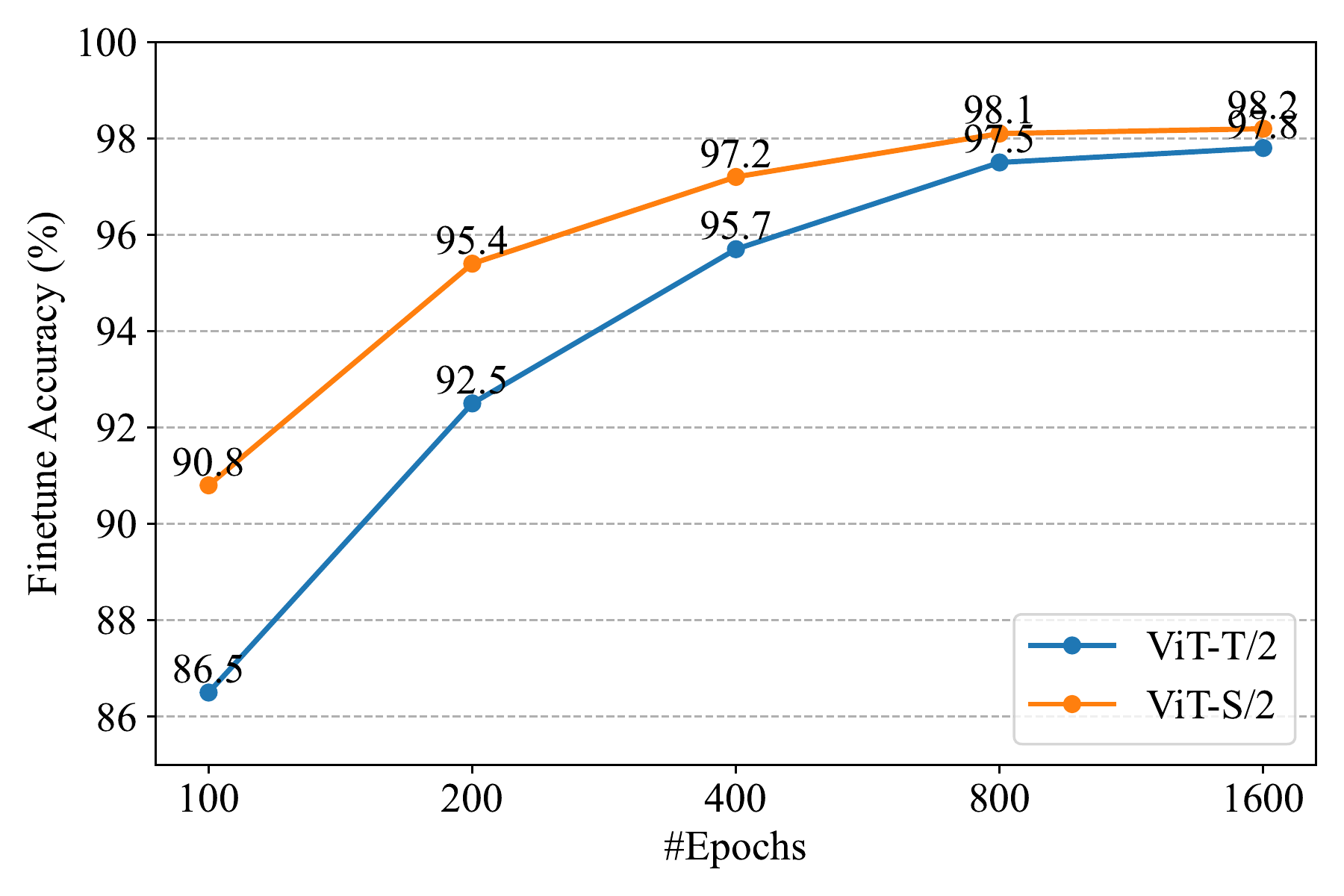}
      & \includegraphics[width=0.33\linewidth]{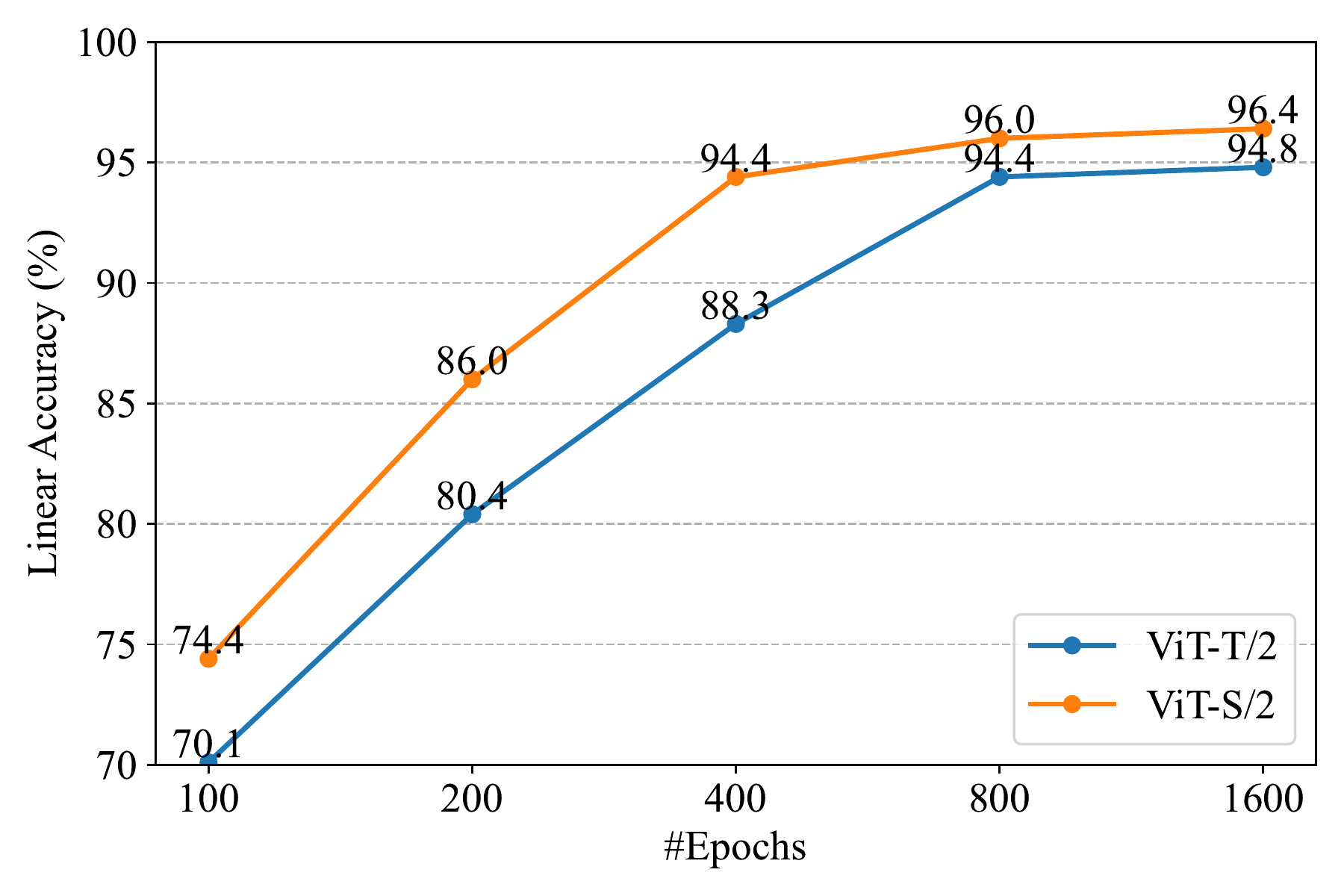}
      & \includegraphics[width=0.33\linewidth]{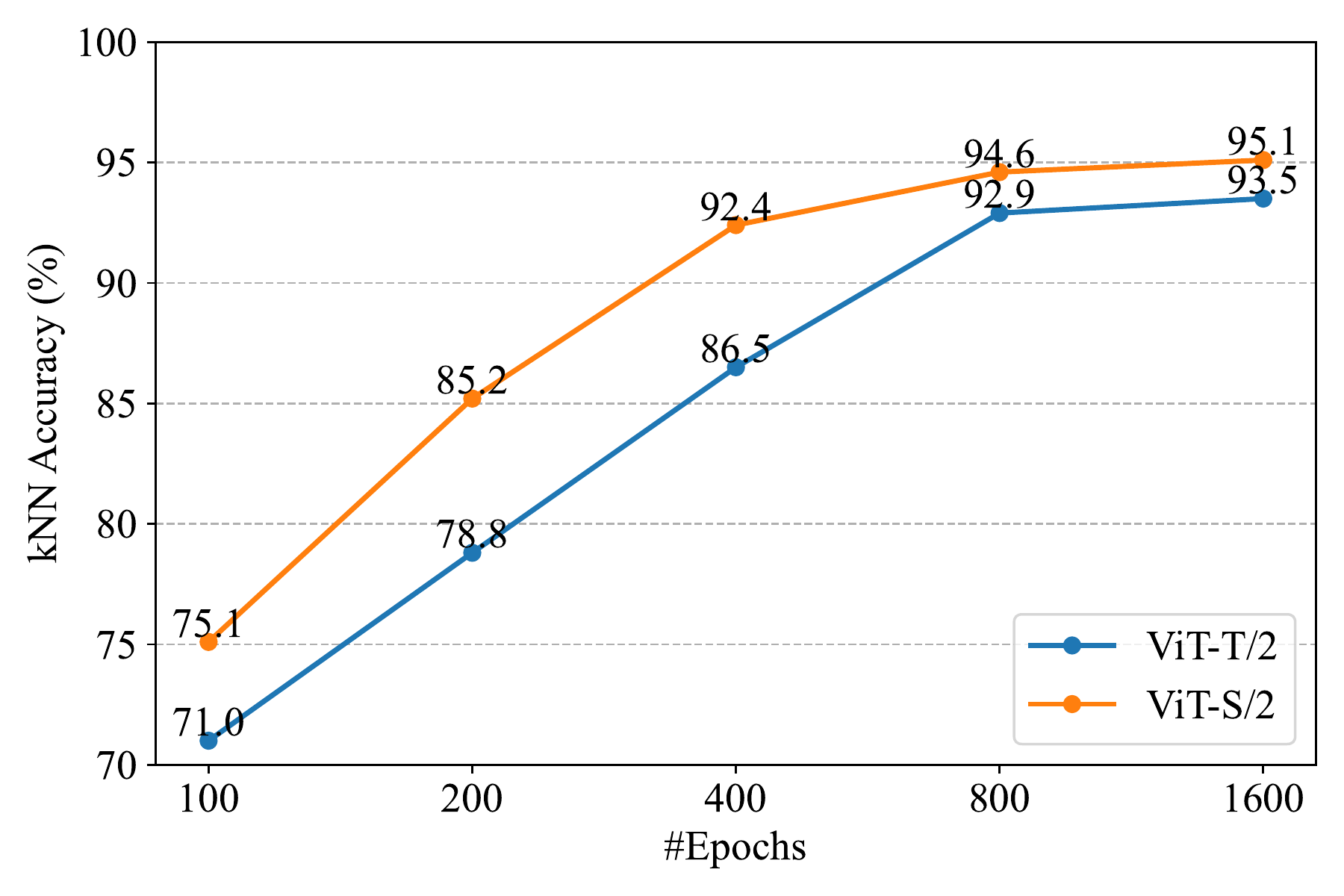} \\
      \rotatebox{90}{{\scriptsize \qquad \qquad \qquad  CIFAR100}} 
      & \includegraphics[width=0.33\linewidth]{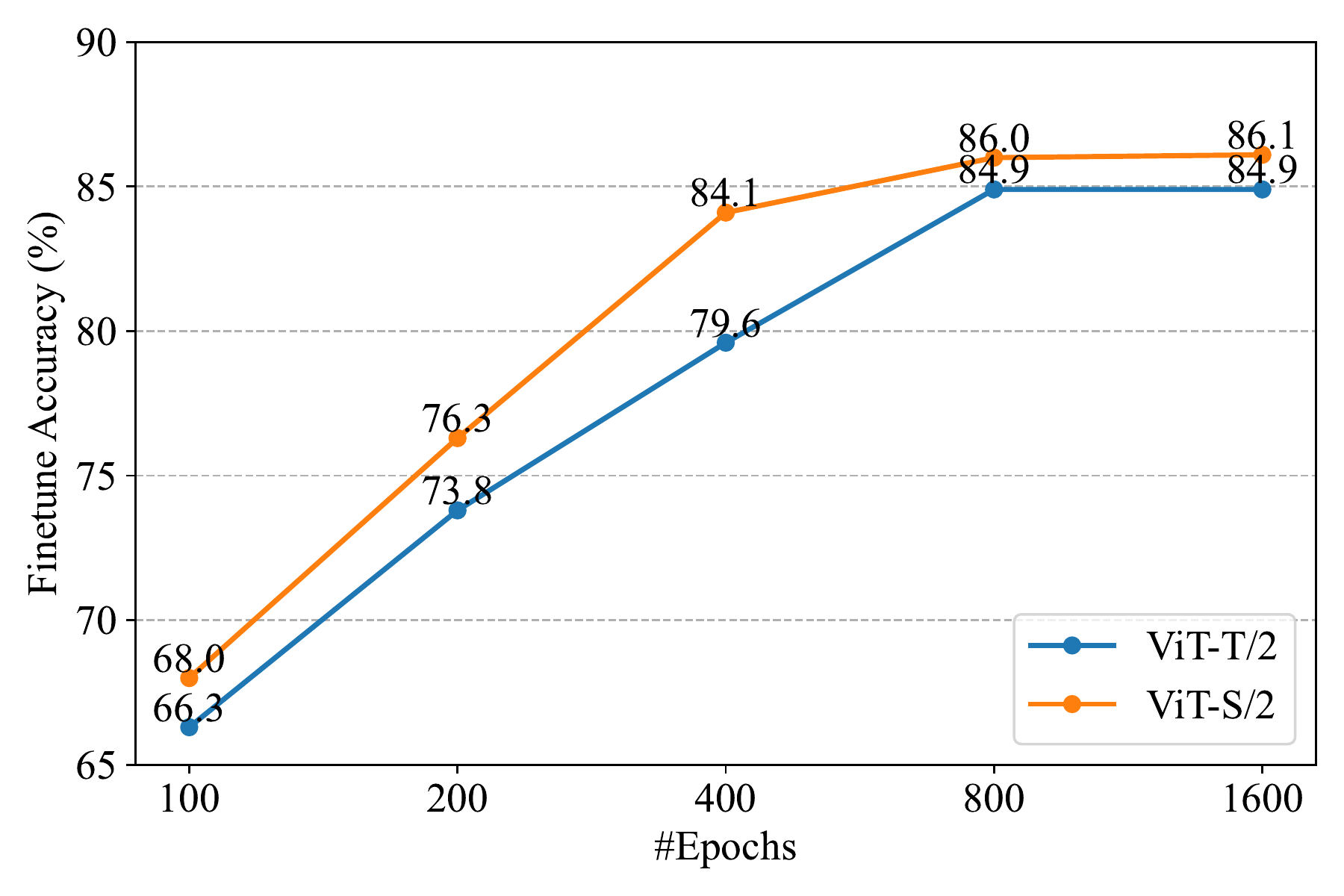} 
      & \includegraphics[width=0.33\linewidth]{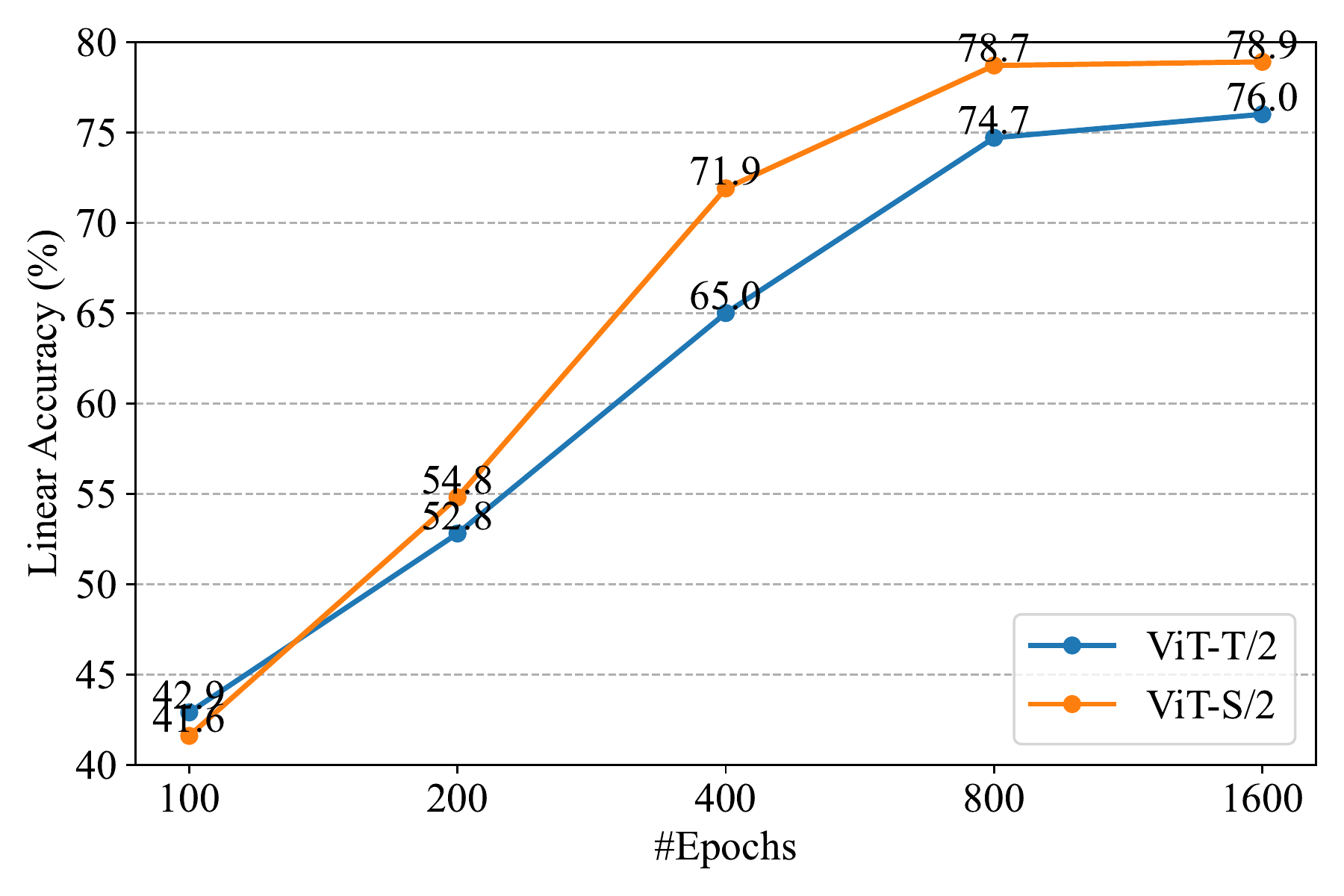}
      & \includegraphics[width=0.33\linewidth]{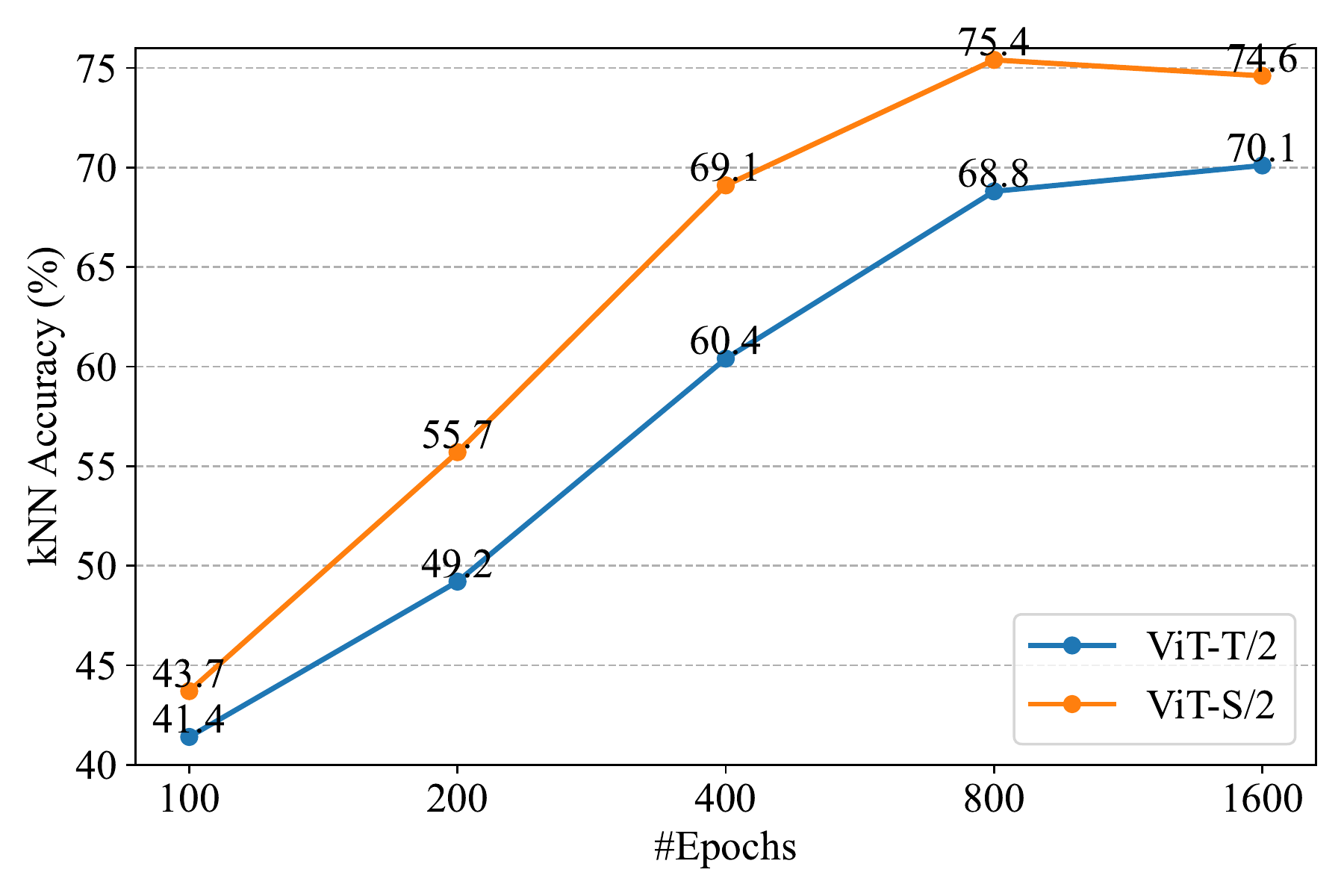} \\
      & {\small (a) finetune accuracy}
      & {\small (b) linear accuracy}
      & {\small (c) kNN accuracy} \\
   \end{tabular}
   \caption{Pretraining with different numbers of epochs on CIFAR10 and CIFAR100 dataset.
   The accuracies are evaluated under the finetune protocol, linear protocol and kNN protocol, respectively. 
   }
   \label{fig:epochs}
\end{figure*}

To validate the effect of training epochs during self-supervised pretraining, we conduct experiments with different numbers of pretraining epochs on CIFAR10 and CIFAR100, respectively. 
As shown in Figure~\ref{fig:epochs}, we report performance on both ViT-T/2 and ViT-S/2 under all three evaluation protocols, finetune evaluation protocol, linear evaluation protocol and kNN evaluation protocol. 
We can observe that the proposed method achieves better performance with longer training schedule, especially on CIFAR100.
Meanwhile, the performance improvement progressively reaches to saturation with the increment of pretraining epochs. 
Training the model using our proposed PatchMix for 800 epochs can achieve the balance between performance and time consumption.

\subsection{Inter-Instance Similarity Visualization}
To evaluate inter-instance similarities constructed by our method, we visualize the similarity distribution among multiple images. 
Specifically, we follow the similarity metric adopted in kNN evaluation protocol and normalize this metric as $\frac{\expb{\similar{a}{b}/\tau^{\prime}}}{\expb{1/\tau^{\prime}}} \in (0, 1]$, where the temperature coefficient $\tau^{\prime}=0.07$. 

In Figure~\ref{fig:visualization}, we measure the similarities between query sample and key samples using the outputs of backbone encoder, ViT-S/16 pretrained on ImageNet-1K for 300 epochs. 
For DINO and SDMP, the distribution of similarities to key samples is significantly sharp and the similarity peak concentrates on the sample with the same category as the query image.
SDMP achieves higher similarity on the category-related samples than DINO, but also only a slight one.
In contrast, our proposed PatchMix establishes significantly rich similarity relations among natural images, not only the corresponding category but also the related categories. 
This experimental result demonstrates that our method can effectively improve the generalization of unsupervised representations.
More results can be found in the supplementary material. 

\section{Discussion}
In this section, we analyze the properties of the proposed PatchMix as follows. 
\begin{itemize}
   \item The proposed PatchMix constructs hybrid image, which includes parts from several image instances, to simulate rich inter-instance similarities among natural images. 
   The model pretrained by PatchMix can effectively capture the inter-instance similarities, especially for the images from the related categories as Figure~\ref{fig:visualization}, thus improving the generalization ability of representations among different instances and significantly outperforming the previous unsupervised learning methods as Table~\ref{table:imagenet}. 
   \item Our PatchMix also presents excellent performance on small-scale datasets, such as CIFAR10 and CIFAR100, as Table~\ref{table:cifar}. 
   Despite data-hungry architecture of ViT, the model pretrained by PatchMix significantly alleviates potential overfitting and representation degeneration without additional training data, such as ImageNet-1K. 
   \item Due to the similarities introduced by local patches, our PatchMix encourages the model to capture local structures of images. 
   Hence, the pretrained model presents the excellent transferability on downstream tasks, which are sensitive to local structures, such as object detection and instance segmentation in Table~\ref{table:coco}.
\end{itemize}

\section{Conclusion}
In this paper, we address monotonous similarity issue suffered by contrastive learning methods. 
To this end, we propose a novel image mix strategy, PatchMix, which mixes multiple images in patch level. 
The mixed image contains massive local components from multiple images and efficiently simulates rich similarities among natural images in an unsupervised manner.
To model rich inter-instance similarities among images, the contrasts between mixed images and original ones, mixed images to mixed ones, and original images to original ones are conducted to optimize the ViT model. 
Experimental results illustrate that our method significantly improves the quality of unsupervised representations, achieving state-of-the-art performance on image classification task of ImageNet-1K, CIFAR10 and CIFAR100 datasets, object detection and instance segmentation tasks of COCO dataset.
Extensive experiments support that our proposed PatchMix can effectively model the rich similarities among natural images and improve the generalization of unsupervised representations on various downstream tasks.

In future work, we plan to explore more general and accurate inter-instance modeling pretext task to further improve the representation quality of contrastive learning.


{\small
\bibliographystyle{ieee_fullname}
\bibliography{paper.bbl}
}

\appendix
\renewcommand{\appendixname}{Appendix~\arabic{section}}

\section{Patch Mix}
In this section, we supplement more explanations about patch mix strategy. 
As shown in Figure~\ref{fig:patchmix} (a), we mix the given image $x_i^{\rm shuffle} = \sequence{G_{im}}{m=0}{M-1}$ by rearranging the indices of the first dimension $u(i,m) = (i+m) \bmod N$ as $x_i^{\rm smix} = \{G_{u(i,m)m}\}_{m=0}^{M-1}$. 
When the index $(i+m)$ overflows the boundary along batch dimension, we conduct modular operation on index $(i+m)$ as $\left( (i+m) \bmod N \right)$ to cyclically mix the image patch group $G_{im}$ in the image batch $x^{\rm shuffle}$. 

\begin{figure*}[ht]
   \centering
   \subfloat[patch mix using 2-d indices]{{\includegraphics[width=0.5\linewidth]{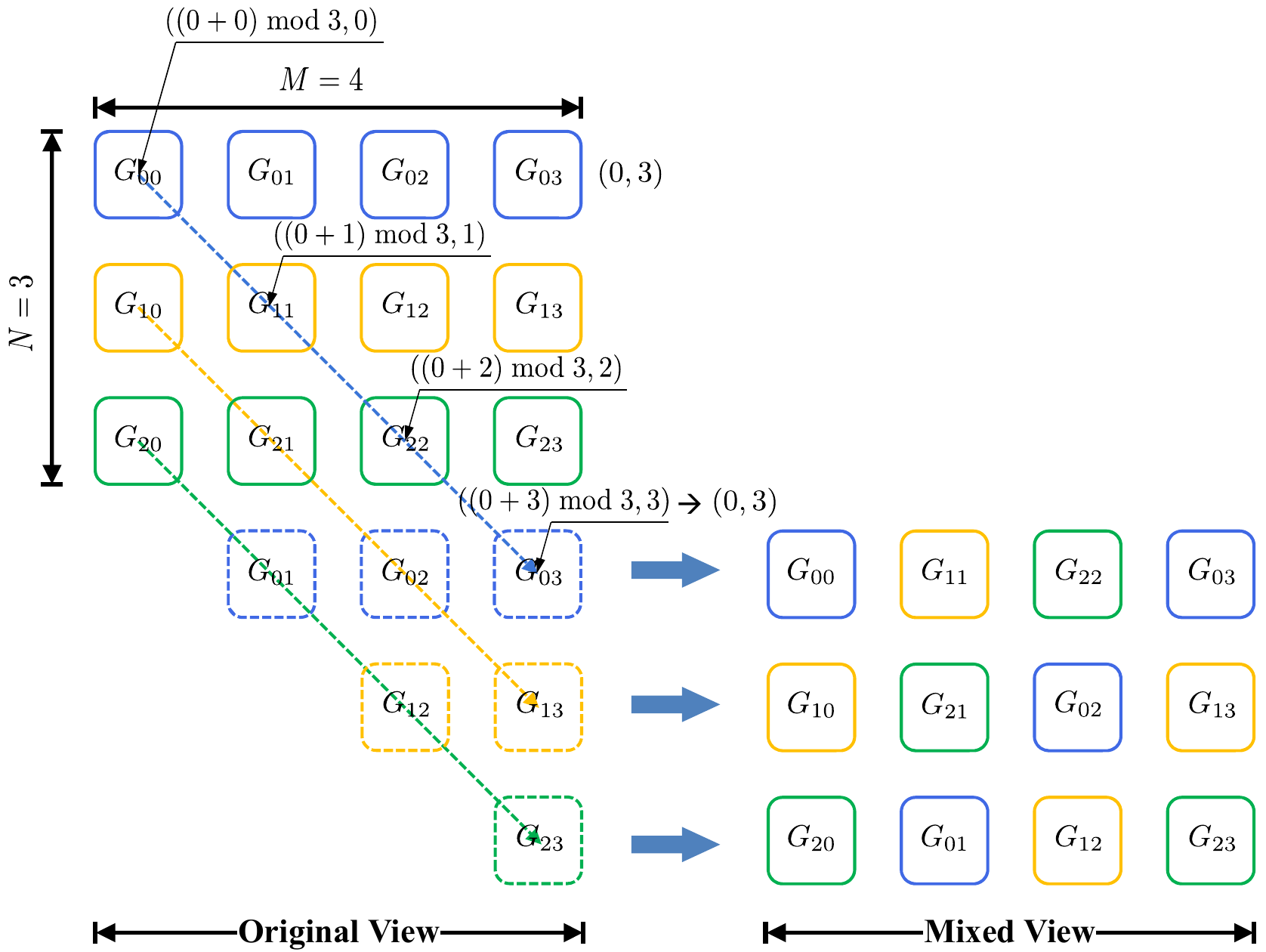}}}
   \hfill
   \subfloat[patch mix using 1-d indices (flatten) ]{{\includegraphics[width=0.5\linewidth]{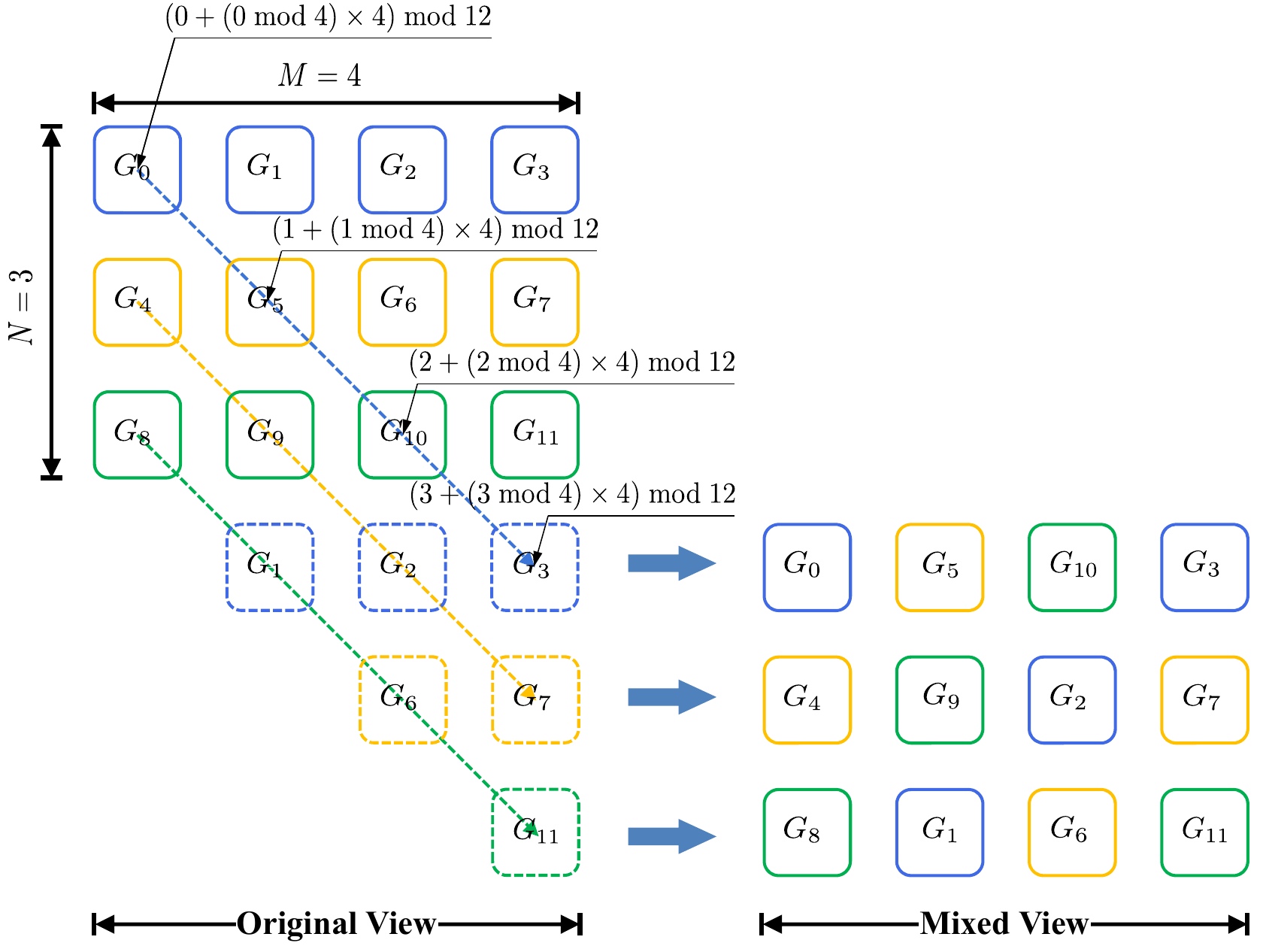}}}
   \caption{The illustration for patch mix strategy, where the mix number $M = 4$ and the batch size $N=3$.
   }
   \label{fig:patchmix2}
\end{figure*}

To fit our patch mix strategy into efficient implement of tensor operation, we flatten the indices of patch groups in image batch 
$x^{\rm shuffle} = \sequence{\sequence{G_{im}}{m=0}{M-1}}{i=0}{N-1}$ as $l = \sequence{\sequence{i \cdot M + m}{m=0}{M-1}}{i=0}{N-1} = \sequence{i}{i=0}{M\cdot N - 1}$ as Figure~\ref{fig:patchmix} (b). 
To mix the image patch group $G_{im}$, a similar index transformation on 1-d indices as $u(i, m)$ is implemented by
\begin{equation}\label{eq:mix_index2}
   q = (l + (l \bmod M) \cdot M) \bmod L, 
\end{equation}
where $L = N \cdot M$ denotes the number of image patch groups in image batch. 
This index transformation can be efficiently implemented by tensor operation.
Our proposed patch mix strategy can be efficiently implemented by $x^{\rm smix} = \sindex{x^{\rm shuffle}}{q}$.

\begin{figure*}[ht]
   \centering
   \includegraphics[width=0.9\linewidth]{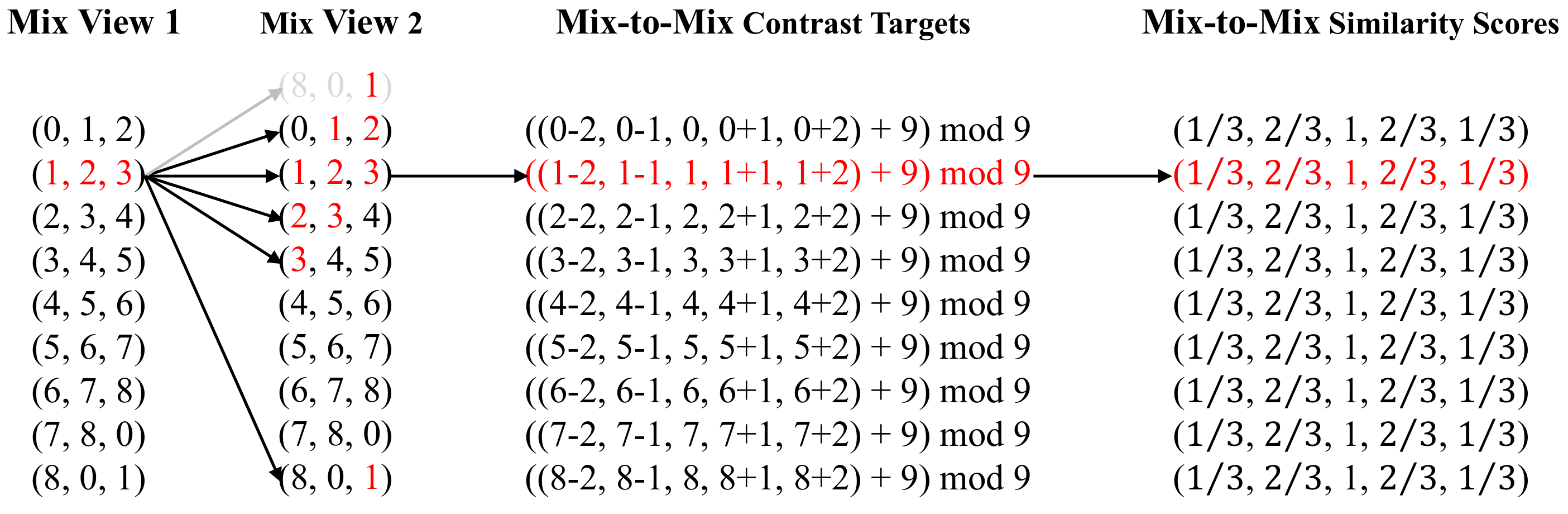}
   \caption{The illustration of mix-to-mix targets and mix-to-mix similarity scores, where the mix nuber $M=3$ and the batch size $N = 9$.}
   \label{fig:mix2mix}
\end{figure*}

We present the computataion of mix-to-mix targets and mix-to-mix similarity scores in Figure~\ref{fig:mix2mix}. 
For example, we analyze the mix-to-mix targets for the second sample in view 1, including the patches from image (1, 2, 3).
The mixed samples from view 2, which contains the patches from image (1, 2, 3), includes 5 samples (8, 0, 1), (0, 1, 2), (1, 2, 3), (2, 3, 4) and (3, 4, 5).
Hence, the mix-to-mix target between view 1 and view 2 is $((0-2, 0-1, 0, 0+1, 0+2) + 9) \bmod 9 = (7, 8, 0, 1, 2)$.
The general formula for mix-to-mix targets can be written as 
\begin{equation}\label{eq:mix_mix2}
   y^{\rm mtm} = \sequence{\sequence{j}{j=(i-M+1+N) \bmod N}{(i+M-1) \bmod N}}{i=0}{N-1},
\end{equation}
which contains $(2M-1)$ positive samples for view 1.
As shown in Figure~\ref{fig:mix2mix}, the number of overlapping patches between positive samples and the sample in view 1 are different, 
$(\frac{1}{3}, \frac{2}{3}, 1, \frac{2}{3}, \frac{1}{3})$ for image $(7, 8, 0, 1, 2)$, respectively.
More general formula for mix-to-mix scores can be obtained by
\begin{equation}\label{eq:mix_mix_weight2}
   \omega^{\rm mtm} = \sequence{\sequence{1 - \frac{\left| M-j-1 \right|}{M} }{j=0}{2M-2}}{i=0}{N-1},
\end{equation}
which provides more inter-instance similarities for contrastive learning.

\section{Network Structures}

As shown in Table~\ref{table:backbone}, we give the details of backbones used in our experiments. 
For ImageNet-1K with input size $224 \times 224$, we adopt standard vision transformer architectures, ViT-Small and ViT-Base, where the patch size for tokenization is $16 \times 16$. 
For CIFAR10 and CIFAR100 with input size $32 \times 32$, we modify the patch size of standard vision transformer architectures from $16 \times 16$ to $2 \times 2$, to adapt the small input images. 
We also introduce a more lightweight vision transformer architecture, ViT-Tiny, which only has half head number and half token dimension of ViT-Small.

\begin{table*}[ht]
\centering
{
   \begin{tabular}{ccccccc}
      \toprule[1pt]
      \textbf{Dataset} & \textbf{Network} & \textbf{Patch Size} & \textbf{\#Blocks} & \textbf{\#Heads} & \textbf{Token Dim} & \textbf{\#Params (M)} \\ 
      \toprule[1pt]
      ImageNet-1K & ViT-Small & 16 & 12 & 6   & 384 & 21.6 \\ 
                  & ViT-Base  & 16 & 12 & 12  & 768 & 85.7 \\ 
      \hline
      CIFAR       & ViT-Tiny  & 2  & 12 & 3   & 192 & 5.4  \\ 
                  & ViT-Small & 2  & 12 & 6   & 384 & 21.3 \\ 
                  & ViT-Base  & 2  & 12 & 12  & 768 & 85.1 \\ 
      \bottomrule[1pt]

   \end{tabular}
}

\caption{The structure of visual transformer backbones. 
``\#Blocks'' denotes the number of standard transformer blocks in backbone. 
``Token Dim'' denotes the dimension of visual token vector.
}
\label{table:backbone}
\end{table*}

\begin{table*}[ht]
   \centering
   {
      \begin{tabular}{c|c|l|l}
         \hline
         \textbf{Dataset} & \textbf{Layer} & \makecell[c]{\textbf{Projection Head}} & \makecell[c]{\textbf{Prediction Head}} \\
         \hline
         ImageNet-1K & 1 & Linear (4096) + BN + ReLU & Linear (4096) + BN + ReLU\\
                     & 2 & Linear (4096) + BN + ReLU & Linear (256) + BN\textsuperscript{*} \\
                     & 3 & Linear (256) + BN\textsuperscript{*}              & \\
         \hline
         CIFAR       & 1 & Linear (4096) + BN + ReLU  & Linear (4096) + BN + ReLU\\
                     & 2 & Linear (4096) + BN + ReLU  & Linear (256) + BN \textsuperscript{*} \\
                     & 3 & Linear (256) + BN\textsuperscript{*}              & \\
         \hline
      \end{tabular}
   }
   \caption{The structure of projection and prediction heads. 
   ``Linear ($m$)'' denotes linear layer with output size $m$.
   ``BN'' and ``ReLU'' denote batch normalization and rectify linear unit operation, respectively. 
   ``BN\textsuperscript{*}'' denotes batch normalization without learnable parameters.
   }
   \label{table:head}
\end{table*}

As shown in Table~\ref{table:head}, we further present the structure of projection and prediction head adopted during self-supervised pretraining. 
For ImageNet-1K, there are 3 linear layers in projection head and 2 linear layers in prediction heads.
The first two linear layers are followed by batch normalization and rectify linear unit in turn, and the output sizes of them are both 4096. 
The last linear of both heads are followed by only batch normalization, and output sizes of them are 256.
For CIFAR10 and CIFAR100, the configurations of projection and prediction head are consistent with the ones of ImageNet-1K.

\section{Data Augmentations}

As shown in Table~\ref{table:augmentation}, we describe the parameters of data augmentations used during self-supervised pretraining. 
For ImageNet-1K, 6 data augmentation techniques are applied to the input images, including random crop and resize, horizontal flip, color jittering, gray scale, Gaussian blurring, as well as solarization. 
The same data augmentation techniques are applied to CIFAR10 and CIFAR100, except area of the crop.

\begin{table*}[!h]
   \centering
   \resizebox{\linewidth}{!}
   {
      \begin{tabular}{l|l|cc|cc}
         \hline
         \multicolumn{1}{l|}{} & \multicolumn{1}{l|}{} & \multicolumn{2}{c|}{\textbf{ImageNet-1K}} & \multicolumn{2}{c}{\textbf{CIFAR}} \\
         \makecell[c]{\textbf{Augmentation}} & \makecell[c]{\textbf{Parameter}} & \textbf{Aug.} $\mathcal{T}_1(\cdot)$ & \textbf{Aug.} $\mathcal{T}_2(\cdot)$ & \textbf{Aug.} $\mathcal{T}_1(\cdot)$ & \textbf{Aug.} $\mathcal{T}_2(\cdot)$ \\ 
         \hline
         random crop and resize   & area of the crop                    & [0.05, 1.0] & [0.05, 1.0] & [0.1, 1.0] & [0.1, 1.0] \\
                                  & aspect ratio of the crop            & [$\frac{3}{4}$, $\frac{4}{3}$] & [$\frac{3}{4}$, $\frac{4}{3}$] & [$\frac{3}{4}$, $\frac{4}{3}$] & [$\frac{3}{4}$, $\frac{4}{3}$] \\
         \hline
         random horizontal flip   & horizontal flip probability         & 0.5 & 0.5 & 0.5 & 0.5 \\
         \hline
         random color jittering   & color jittering probability         & 0.8 & 0.8 & 0.8 & 0.8 \\
                                  & max brightness adjustment intensity & 0.4 & 0.4 & 0.4 & 0.4 \\
                                  & max contrast adjustment intensity   & 0.4 & 0.4 & 0.4 & 0.4 \\
                                  & max saturation adjustment intensity & 0.2 & 0.2 & 0.2 & 0.2 \\
                                  & max hue adjustment intensity        & 0.1 & 0.1 & 0.1 & 0.1 \\
         \hline
         random gray scale        & color dropping probability          & 0.2 & 0.2 & 0.2 & 0.2 \\
         \hline
         random Gaussian blurring & Gaussian blurring probability       & 1.0 & 0.1 & 1.0 & 0.1 \\
                                  & sigma of Gaussian blurring          & [0.1, 2.0] & [0.1, 2.0] & [0.1, 2.0] & [0.1, 2.0] \\
         \hline
         random solarization      & solarization probability            & 0.0 & 0.2 & 0.0 & 0.2 \\
         \hline
      \end{tabular}
   }
   \caption{The parameters of data augmentations applied during self-supervised training.
   ``[·, ·]'' denotes the range for uniform sampling.
   }
   \label{table:augmentation}
\end{table*}

\section{Inter-Instance Similarity Visualization}

\begin{figure*}[ht]
   \renewcommand\arraystretch{0}
   \renewcommand\tabcolsep{1pt}
   \resizebox{\linewidth}{!}
   {
   \begin{tabular}{m{3.5cm}<{\centering} m{3.5cm}<{\centering} m{5cm}<{\centering} m{5cm}<{\centering} m{5cm}<{\centering}}
      \textbf{Query Sample} & \textbf{Key Samples} & \textbf{DINO} & \textbf{SDMP} & \textbf{PatchMix (ours)} \\

   \includegraphics[width=\linewidth]{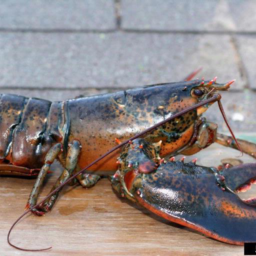}
   & \includegraphics[width=\linewidth]{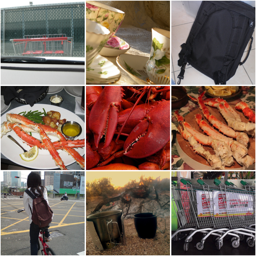}
   & \includegraphics[width=\linewidth]{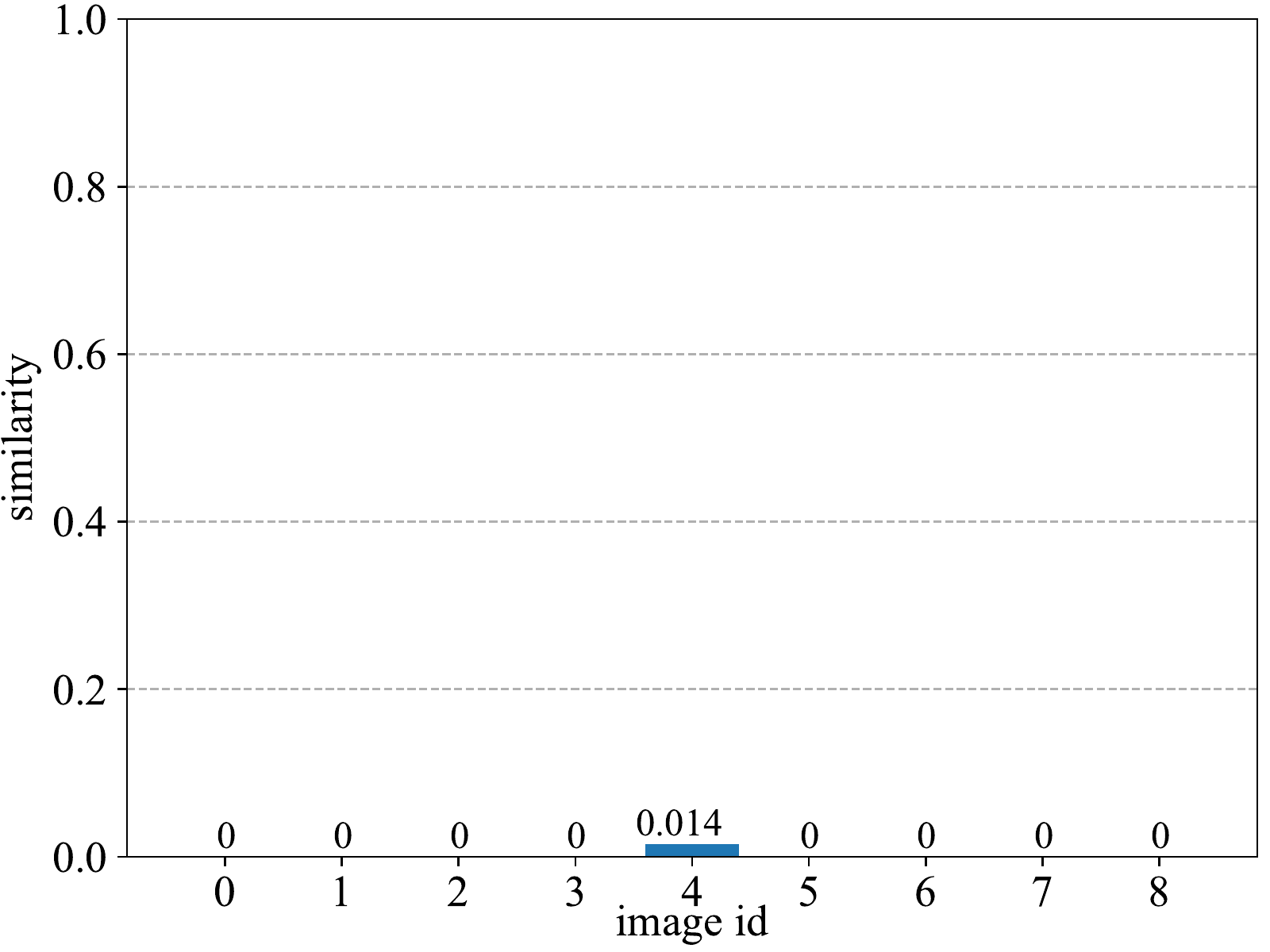}
   & \includegraphics[width=\linewidth]{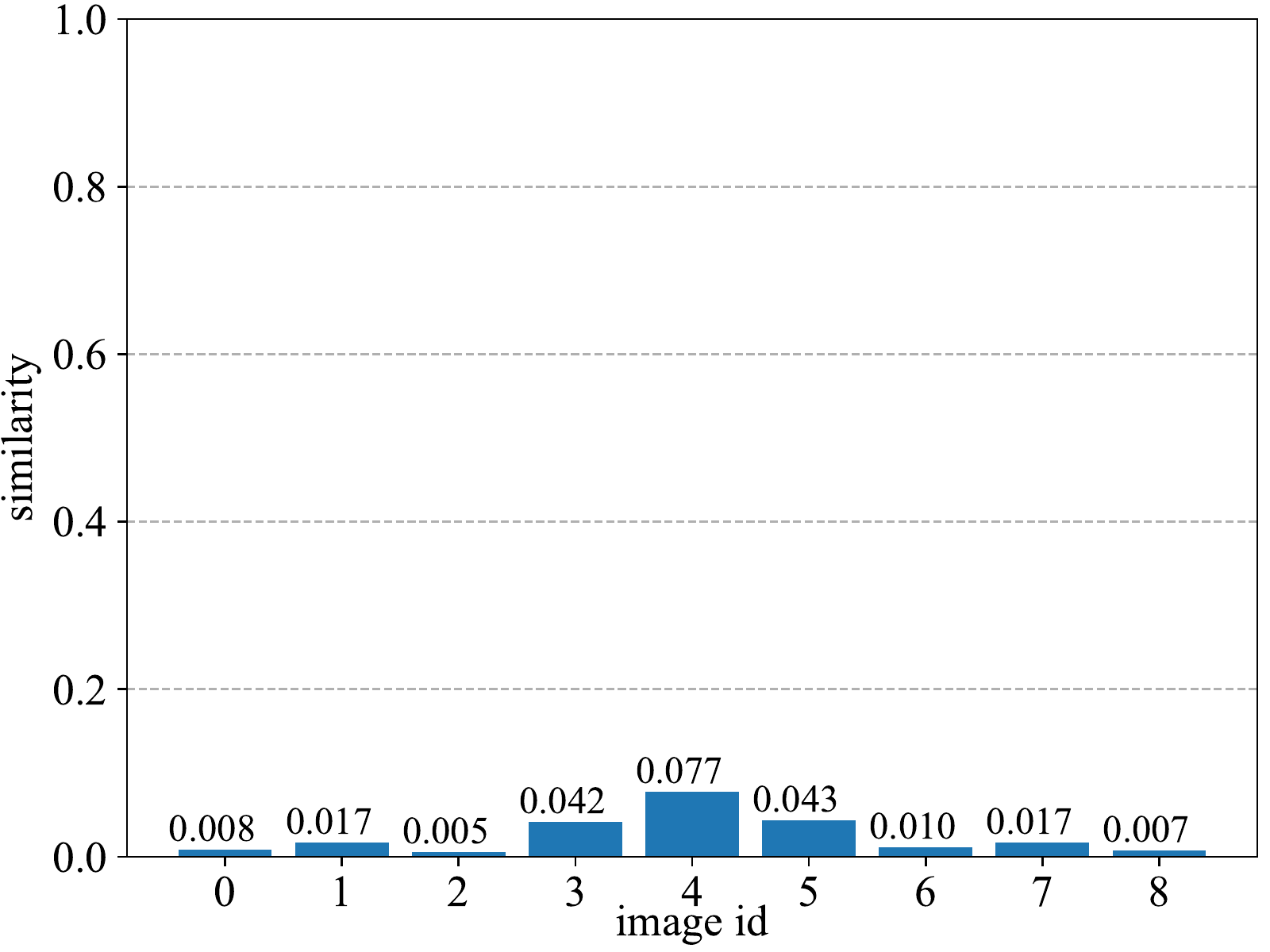}
   & \includegraphics[width=\linewidth]{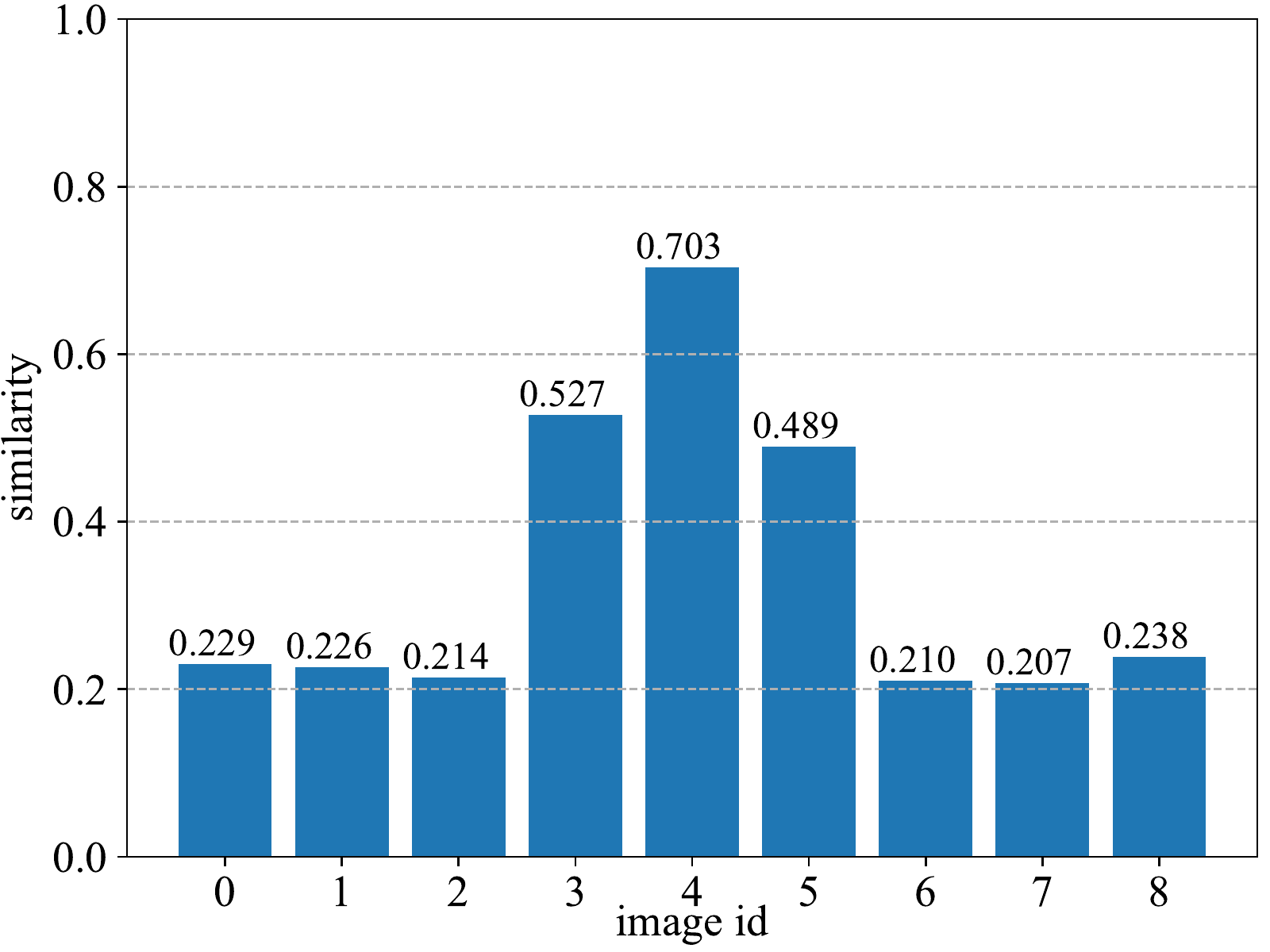} \\
  
   \includegraphics[width=\linewidth]{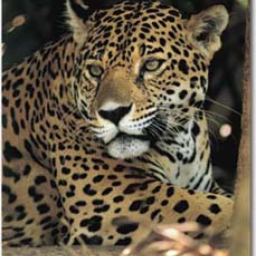}
   & \includegraphics[width=\linewidth]{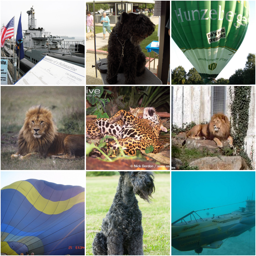}
   & \includegraphics[width=\linewidth]{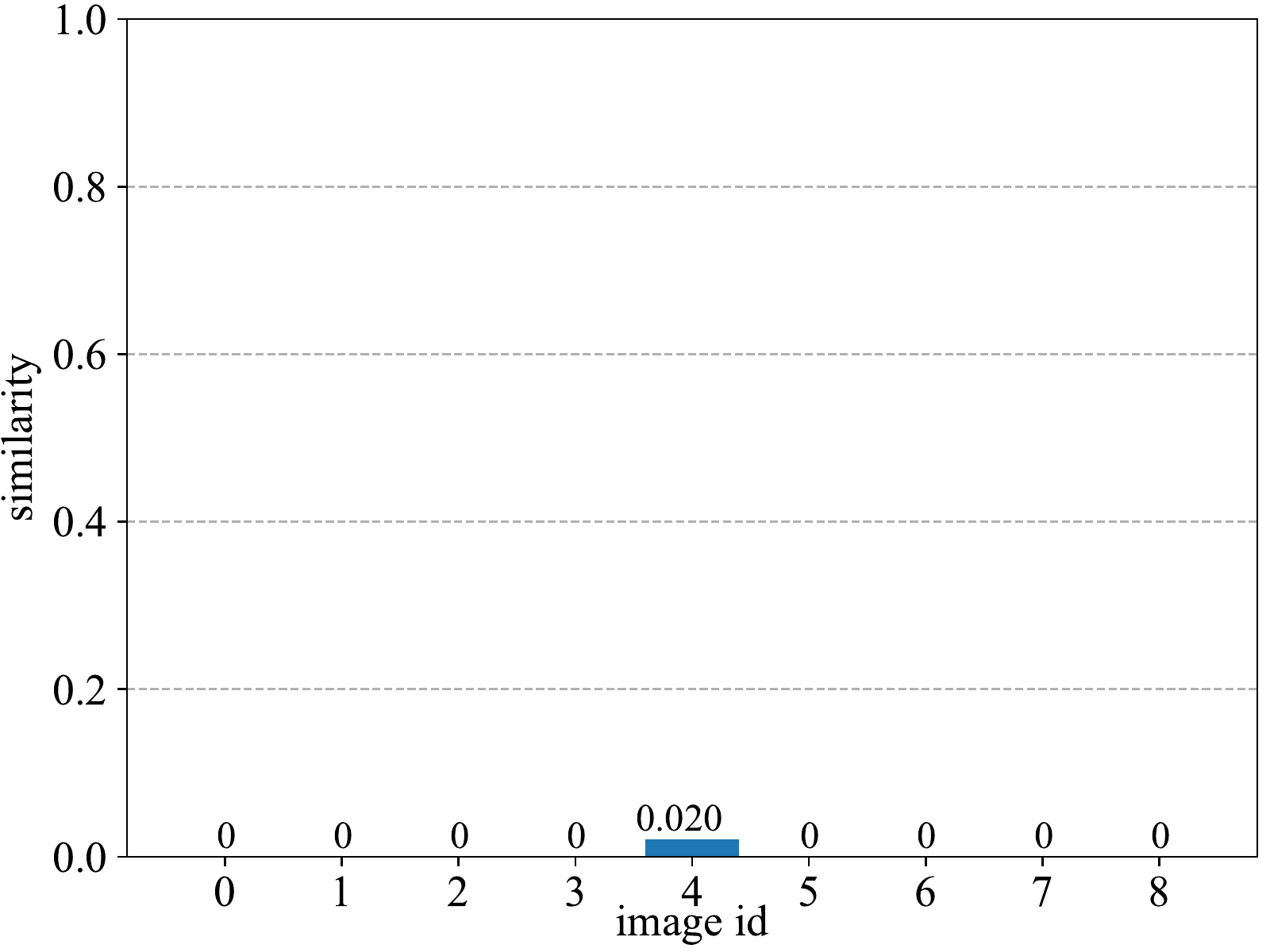}
   & \includegraphics[width=\linewidth]{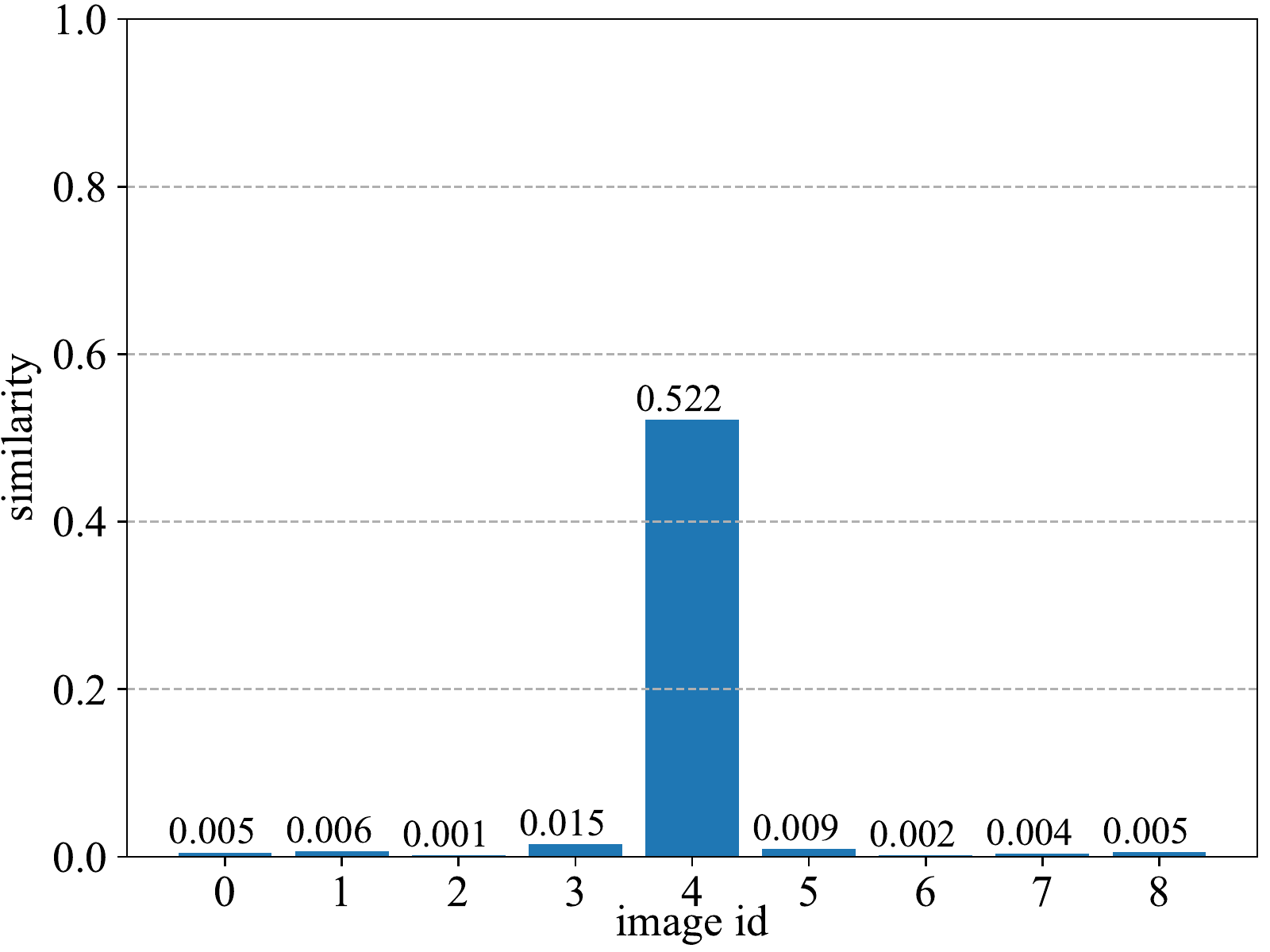}
   & \includegraphics[width=\linewidth]{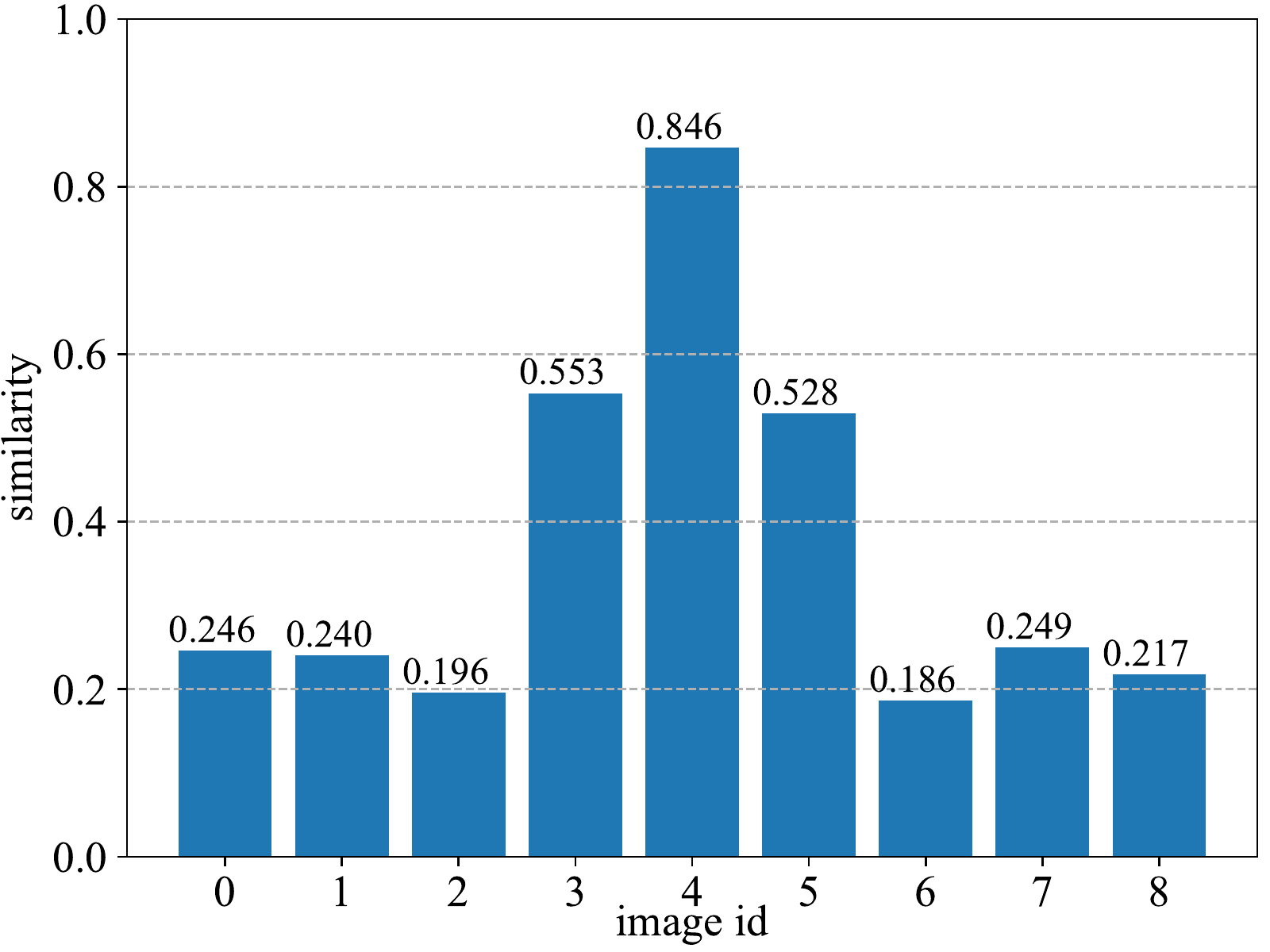} \\

   \includegraphics[width=\linewidth]{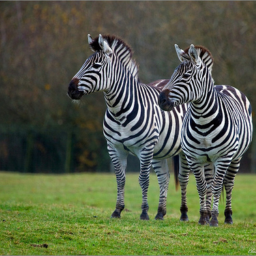}
   & \includegraphics[width=\linewidth]{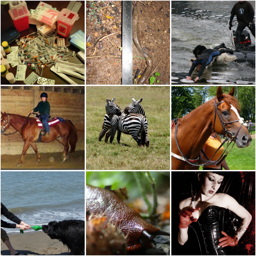}
   & \includegraphics[width=\linewidth]{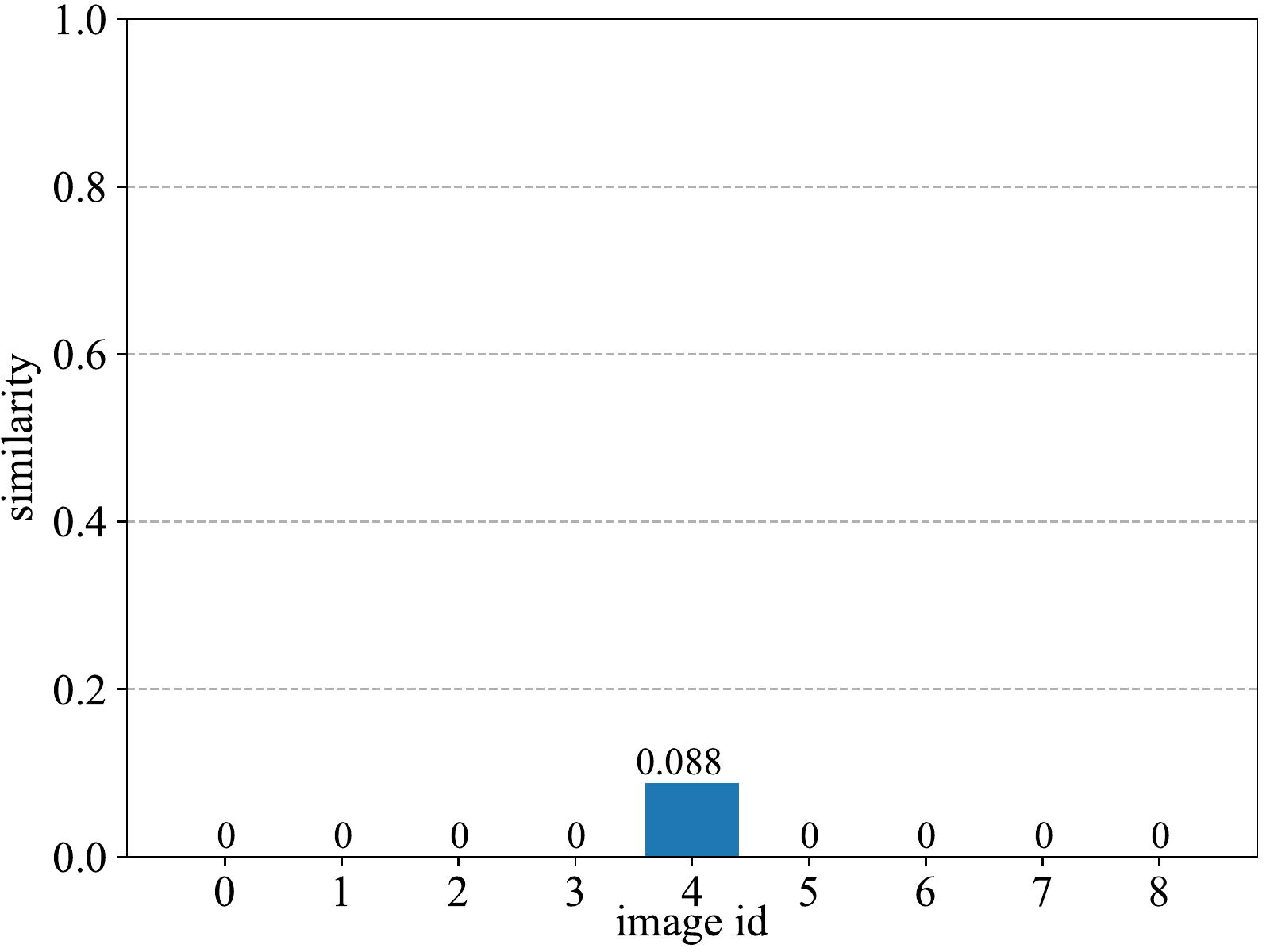}
   & \includegraphics[width=\linewidth]{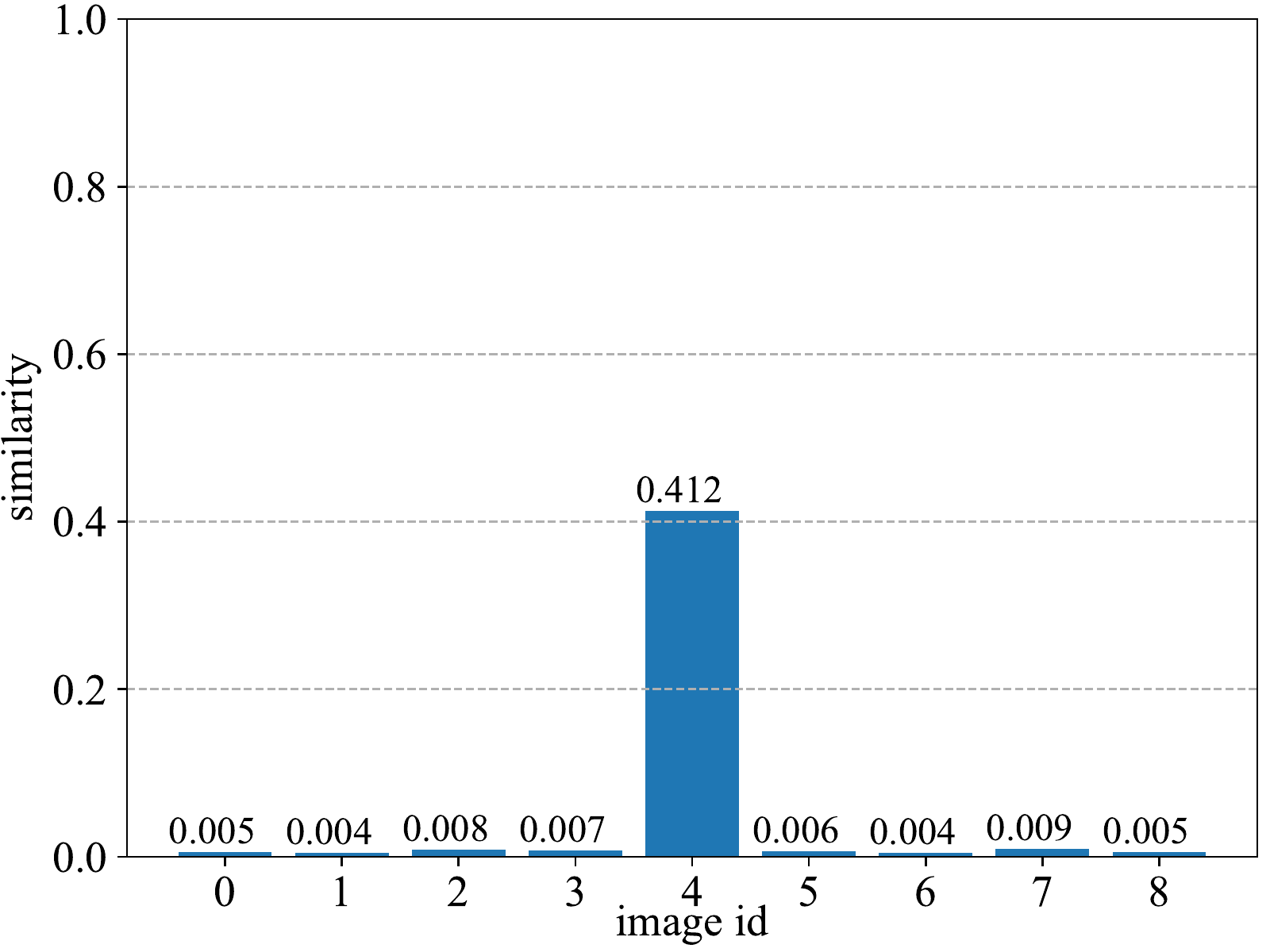}
   & \includegraphics[width=\linewidth]{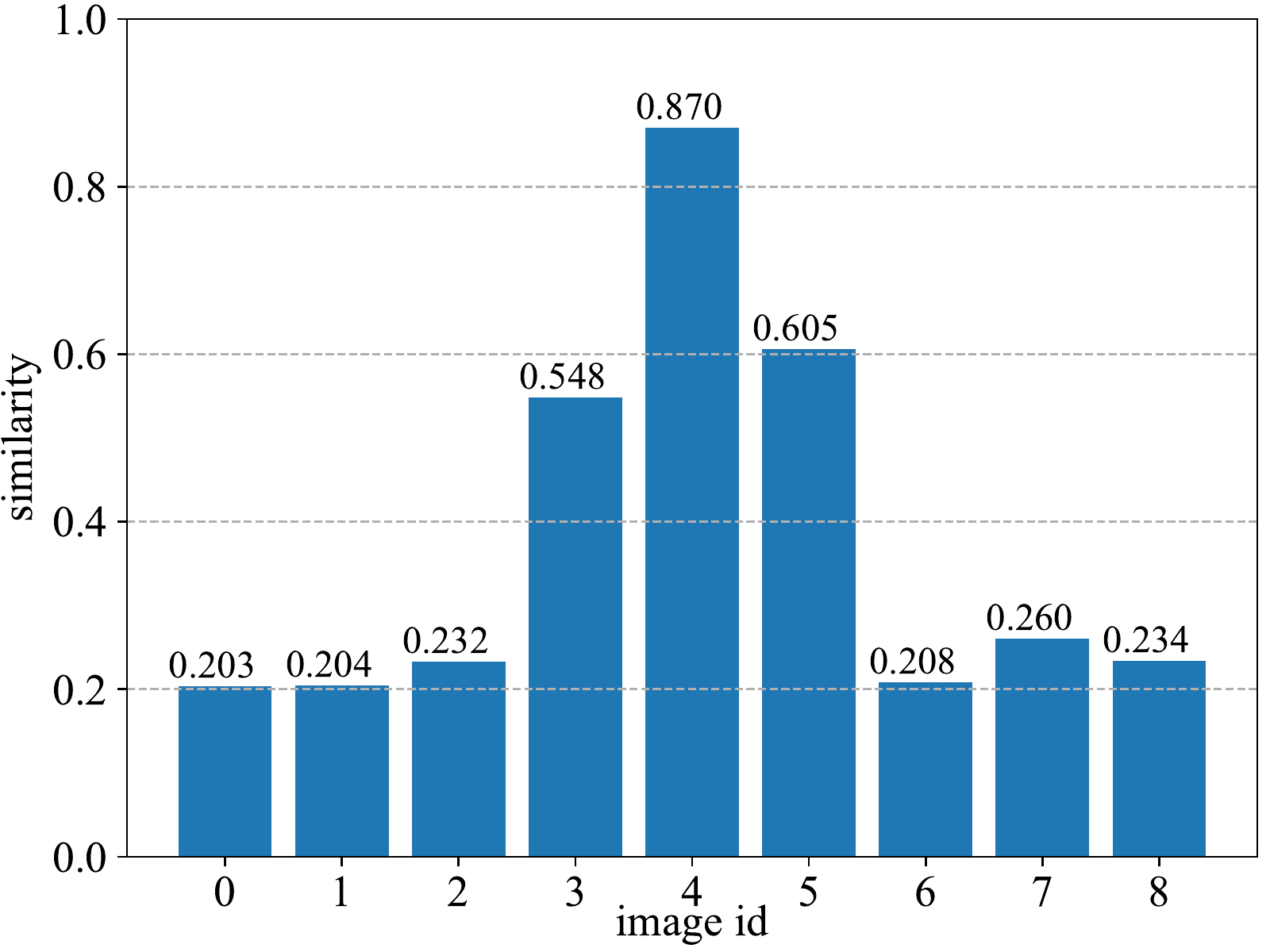} \\
  
    \includegraphics[width=\linewidth]{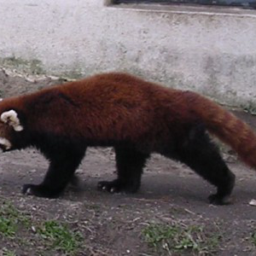}
   & \includegraphics[width=\linewidth]{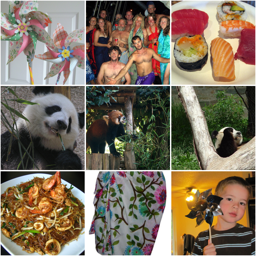}
   & \includegraphics[width=\linewidth]{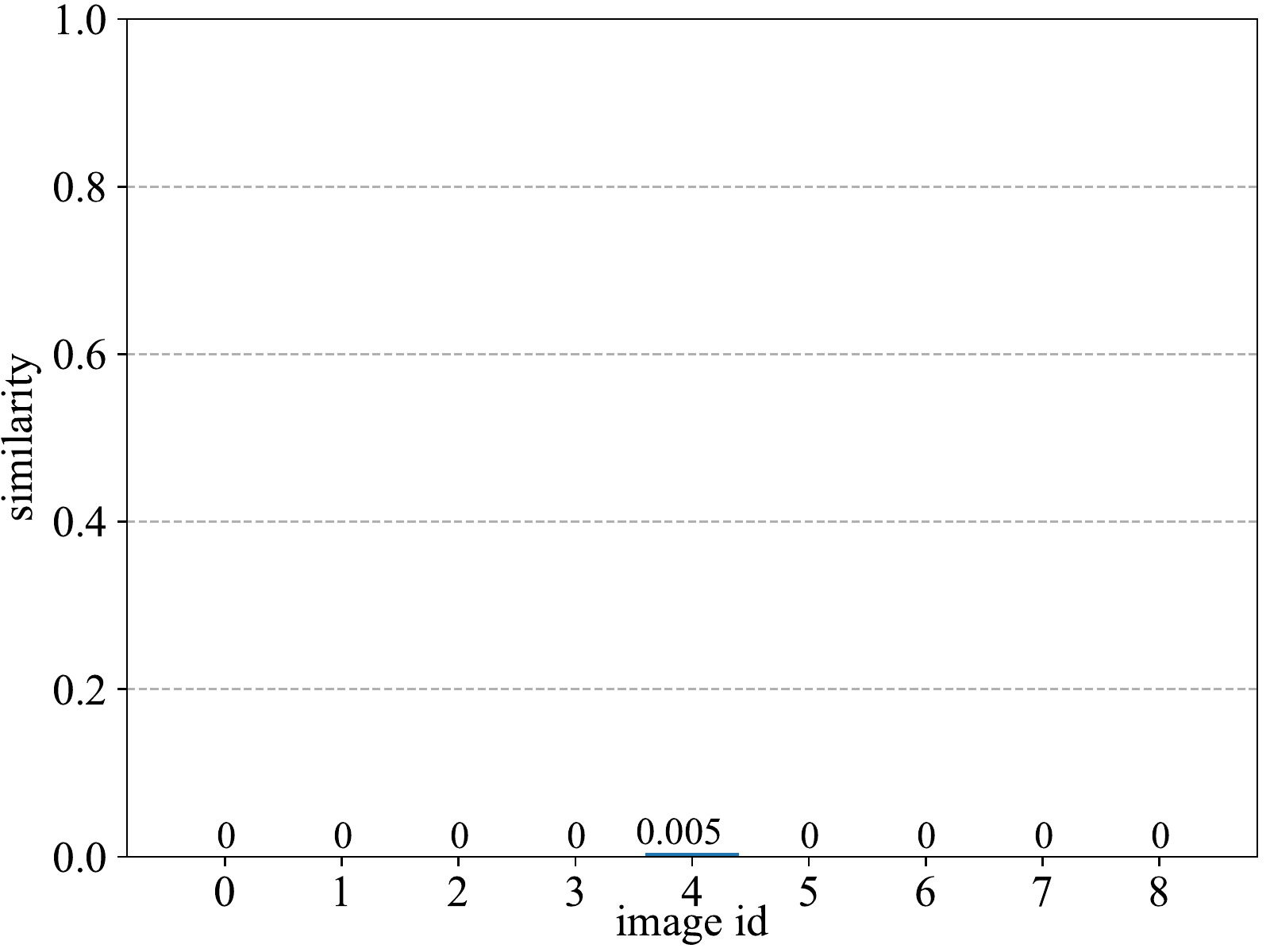}
   & \includegraphics[width=\linewidth]{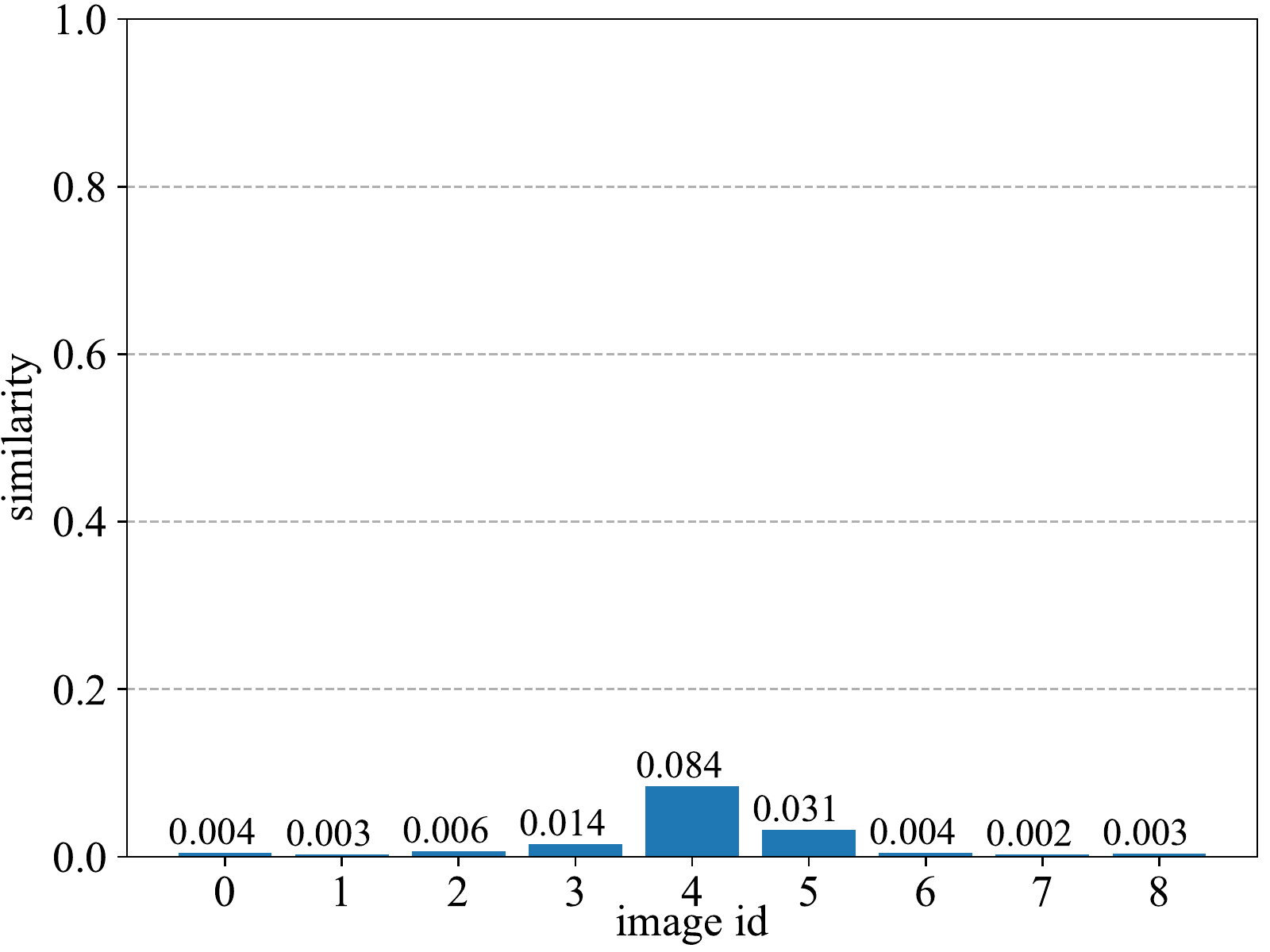}
   & \includegraphics[width=\linewidth]{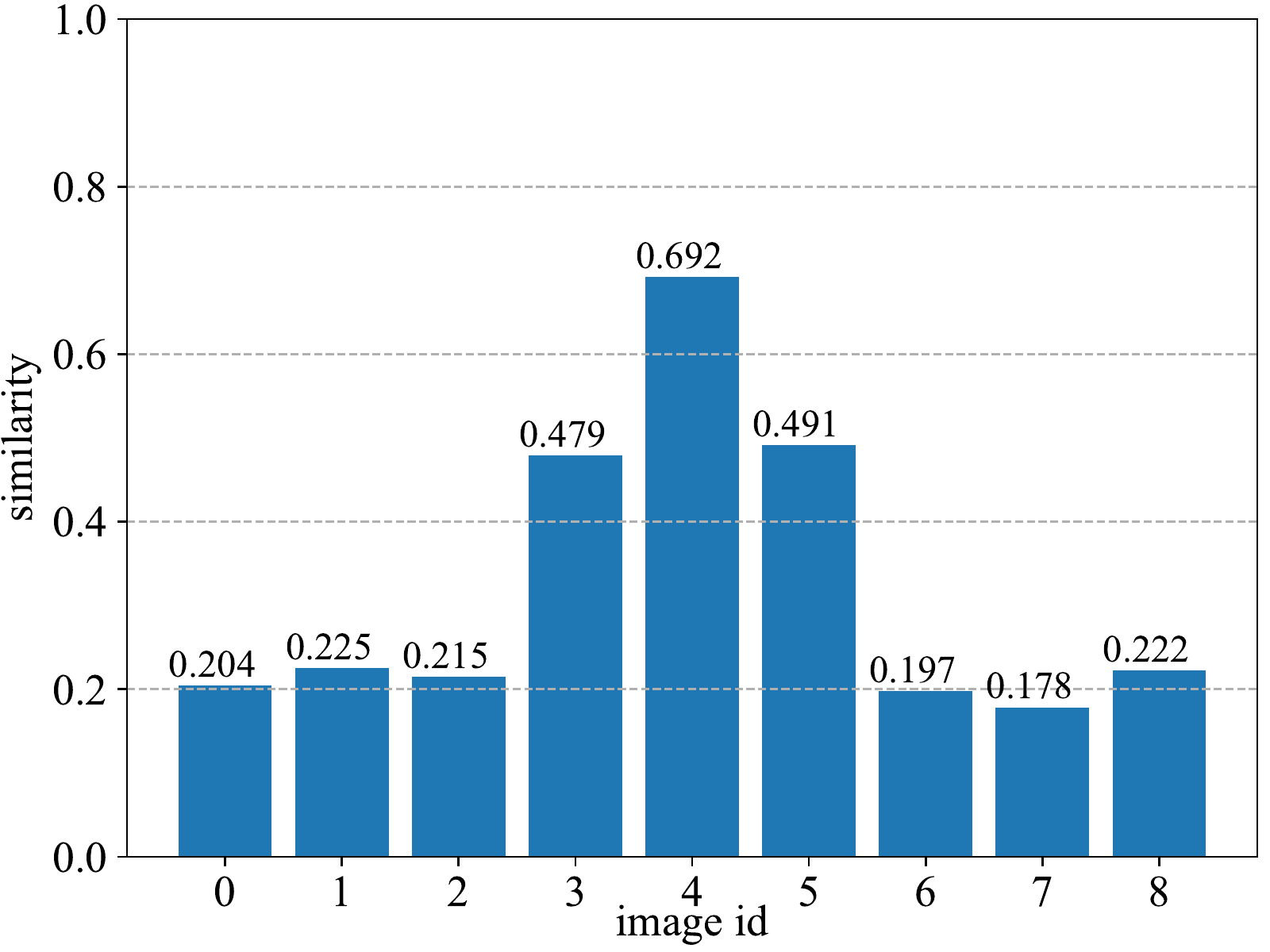} \\

   \includegraphics[width=\linewidth]{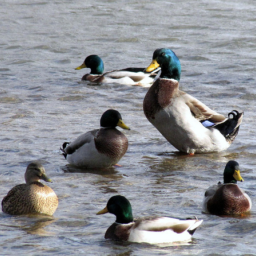}
   & \includegraphics[width=\linewidth]{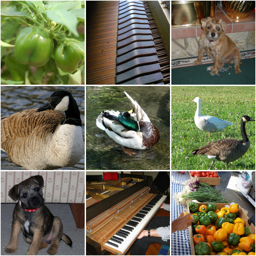}
   & \includegraphics[width=\linewidth]{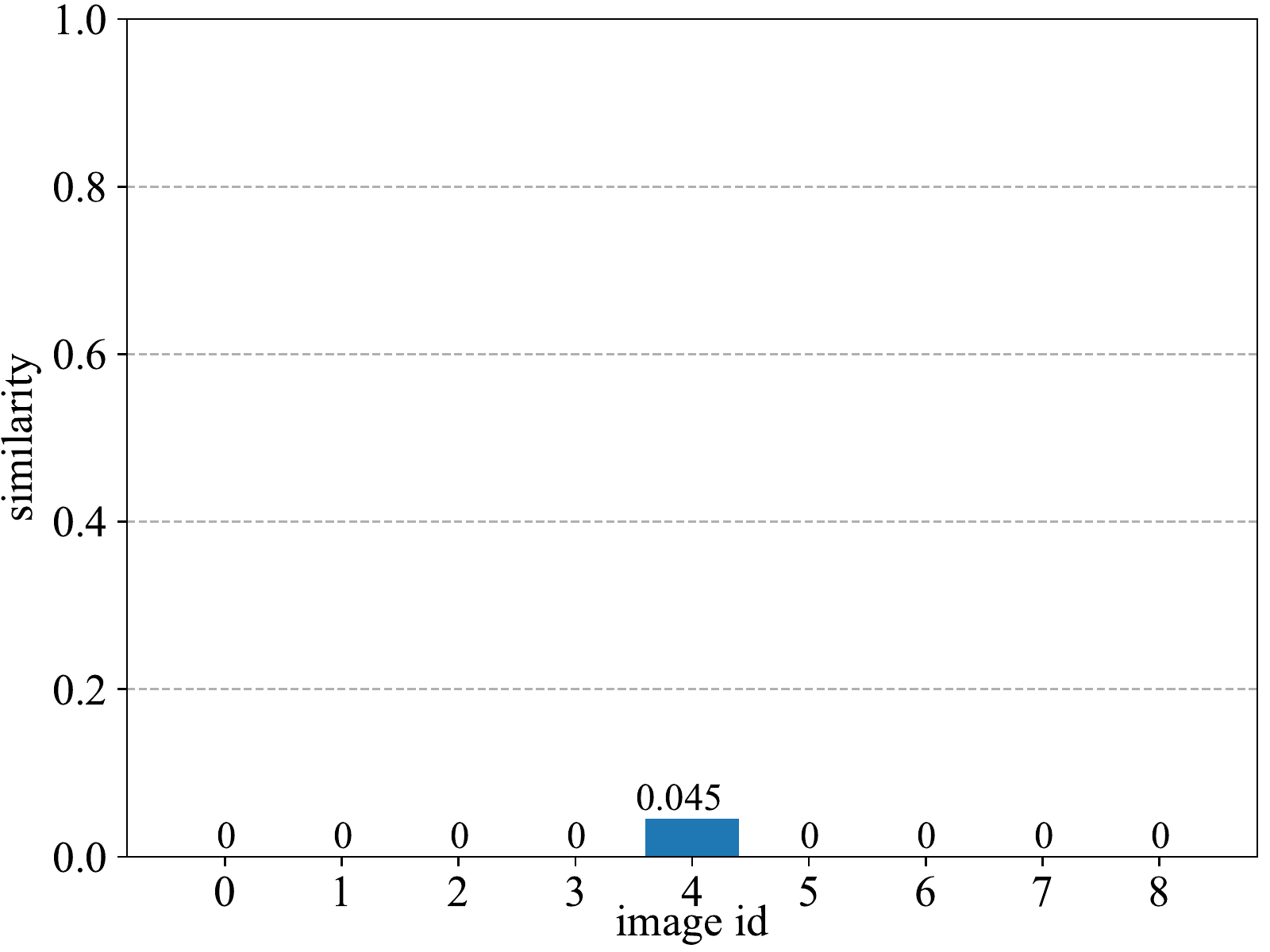}
   & \includegraphics[width=\linewidth]{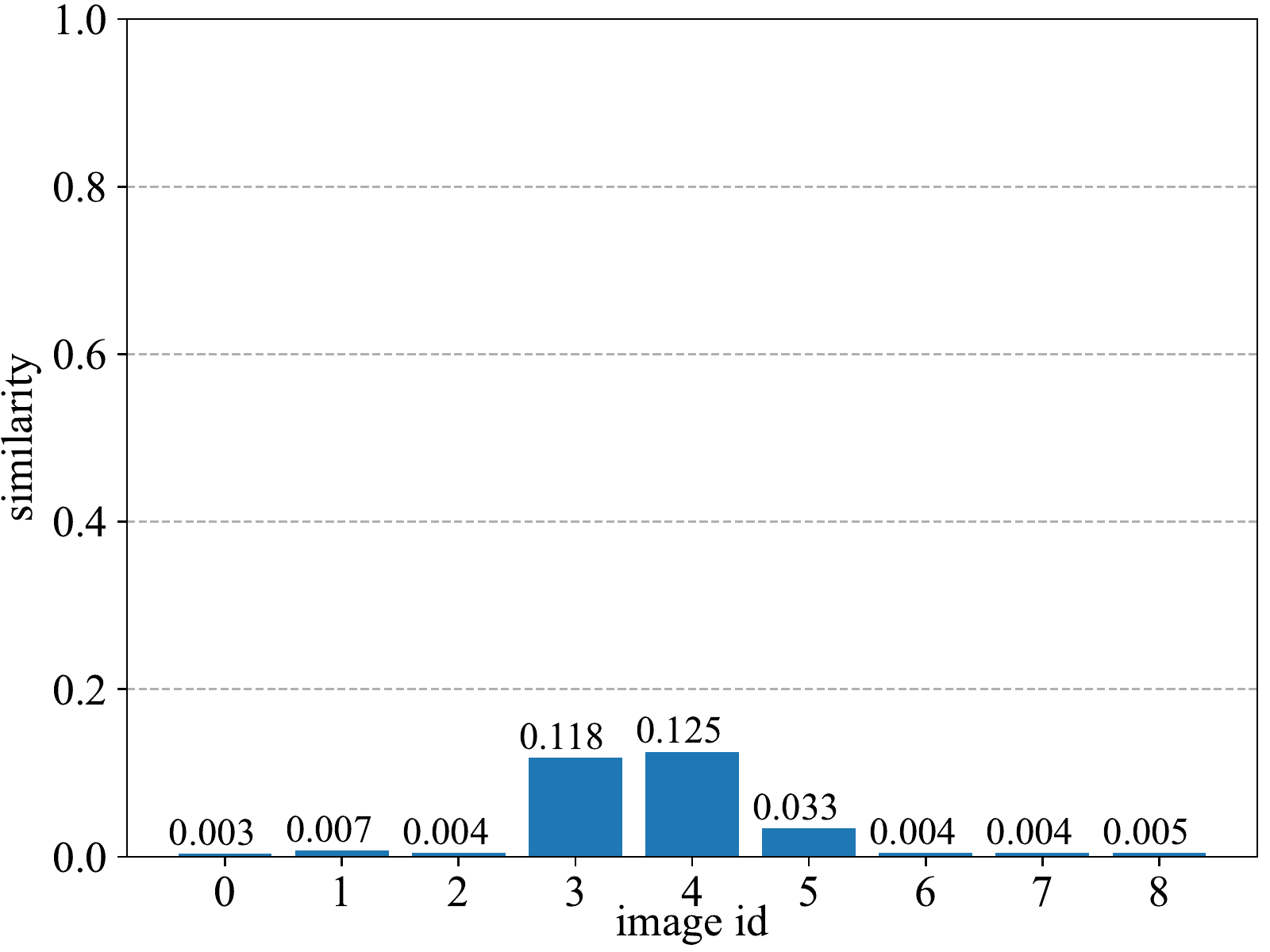}
   & \includegraphics[width=\linewidth]{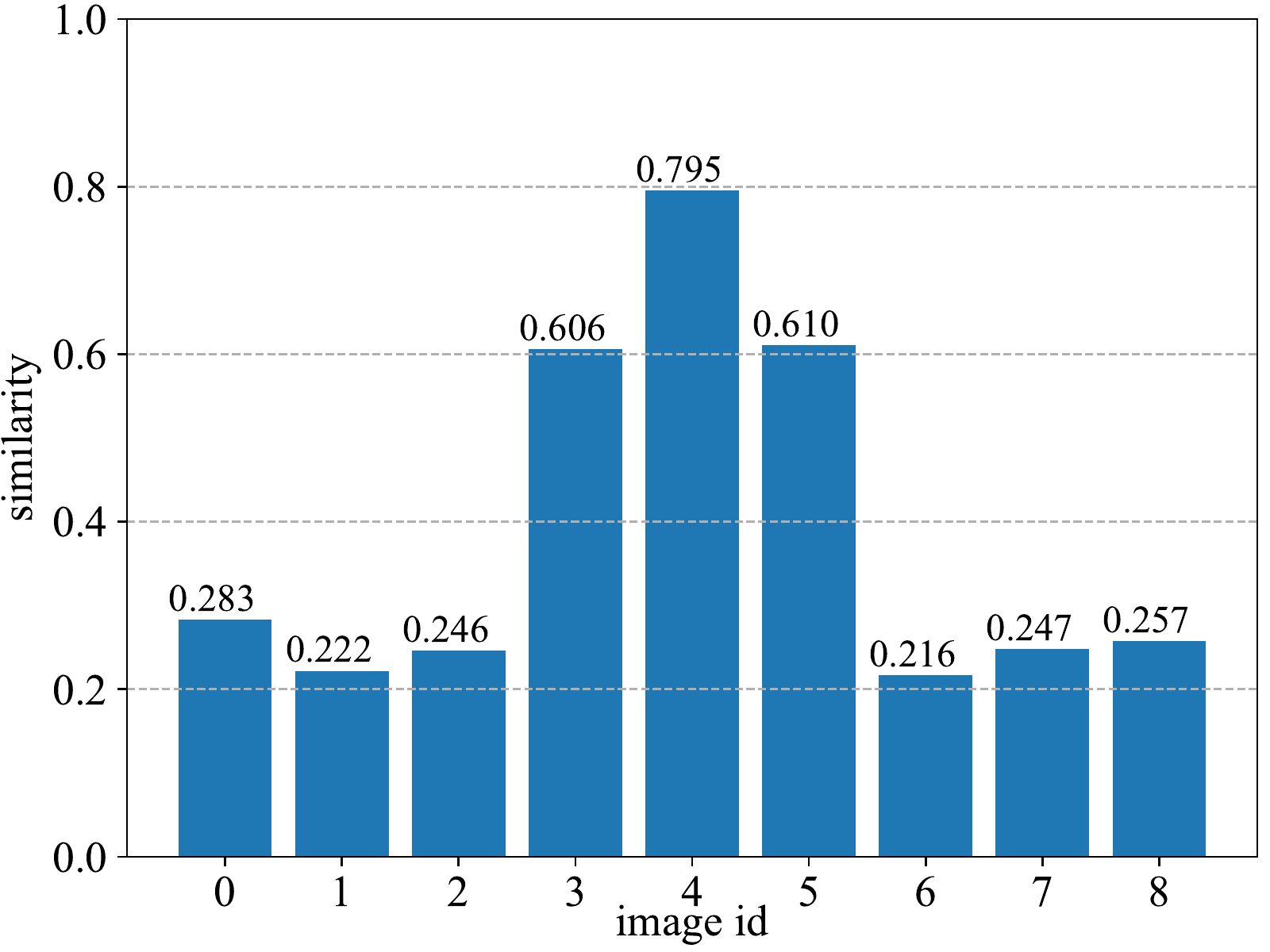} \\

   \includegraphics[width=\linewidth]{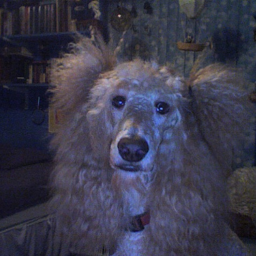}
   & \includegraphics[width=\linewidth]{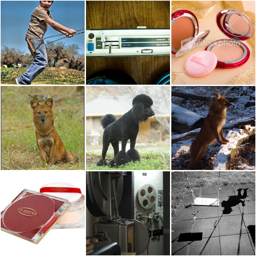}
   & \includegraphics[width=\linewidth]{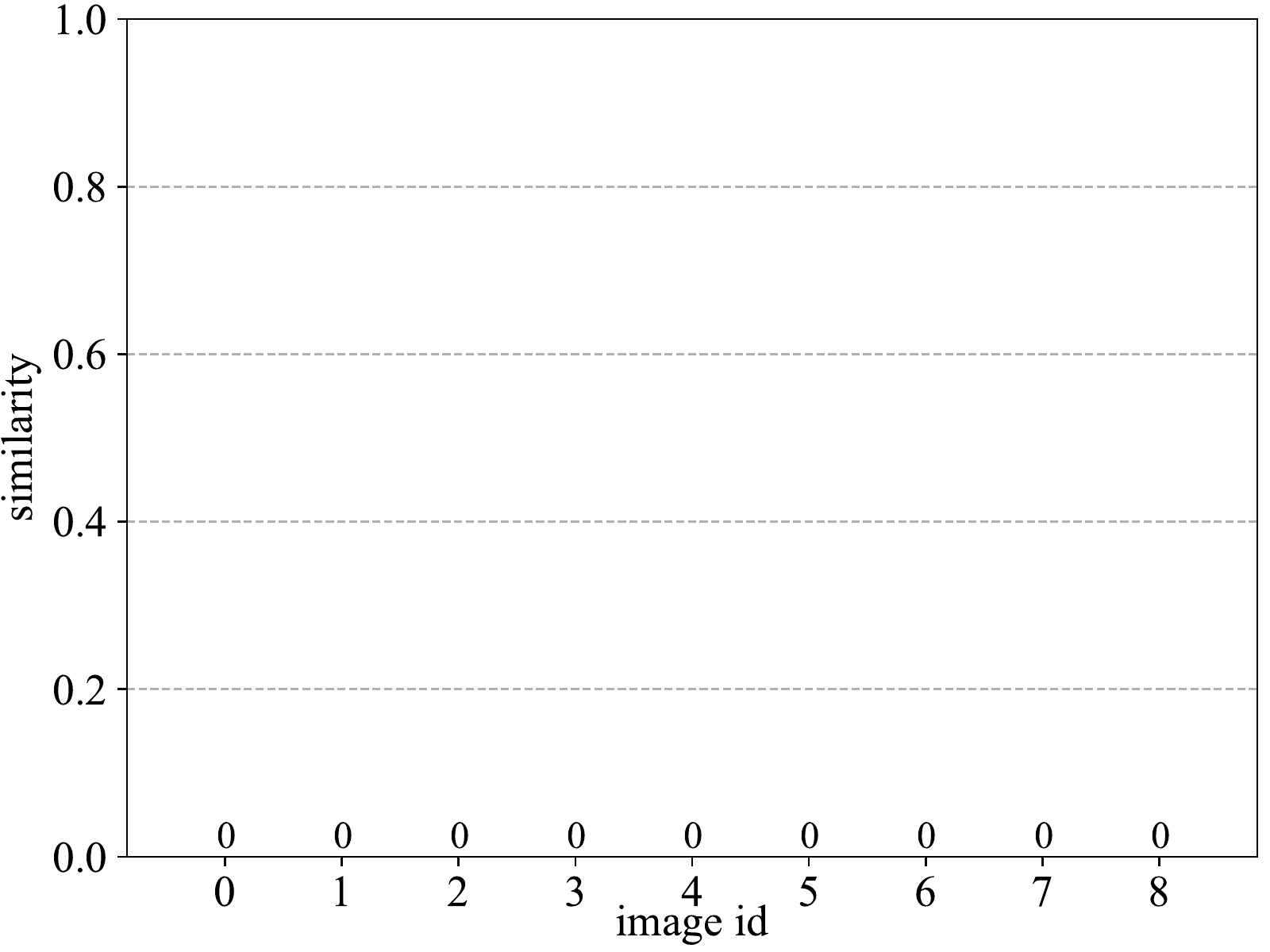}
   & \includegraphics[width=\linewidth]{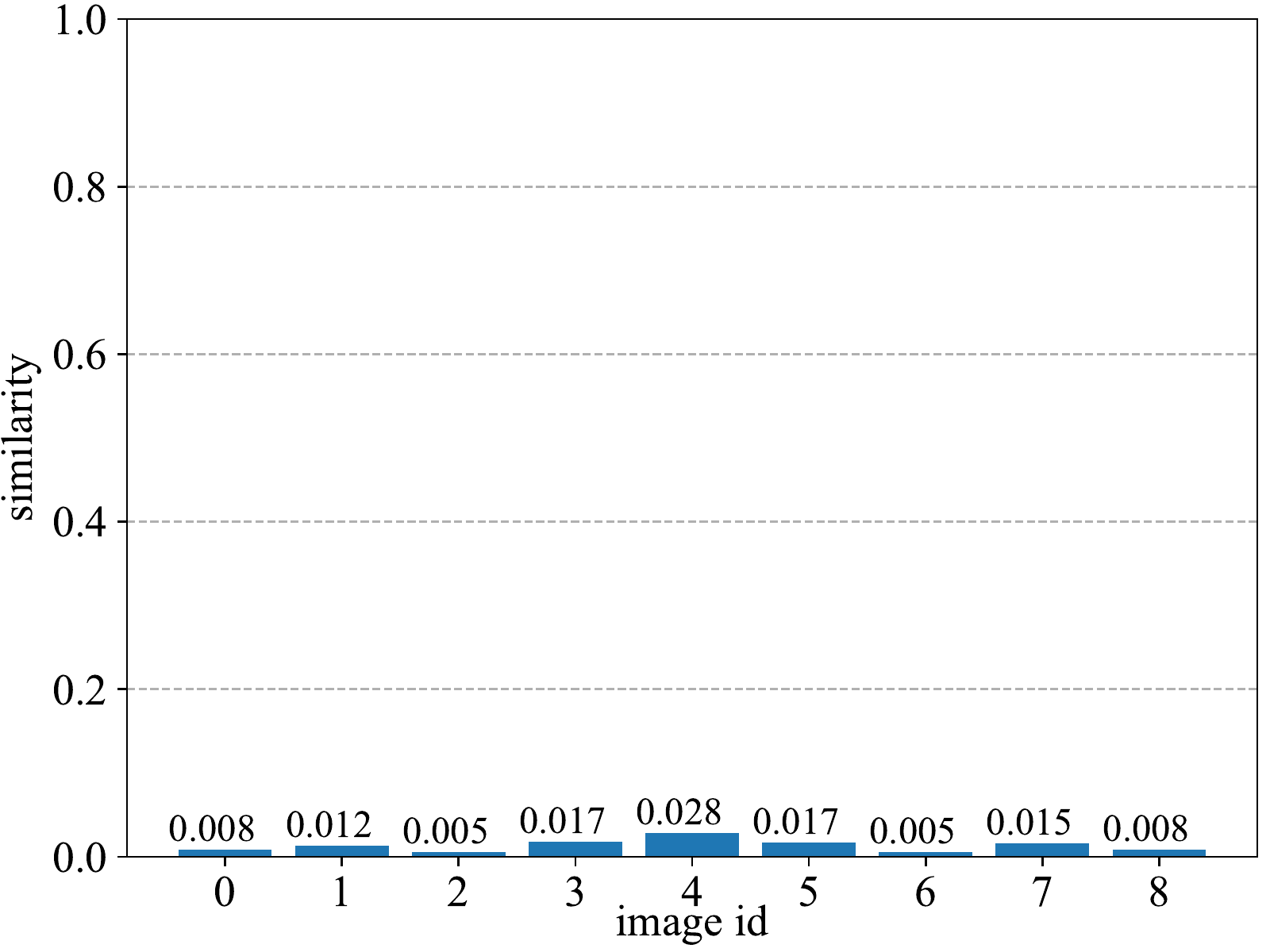}
   & \includegraphics[width=\linewidth]{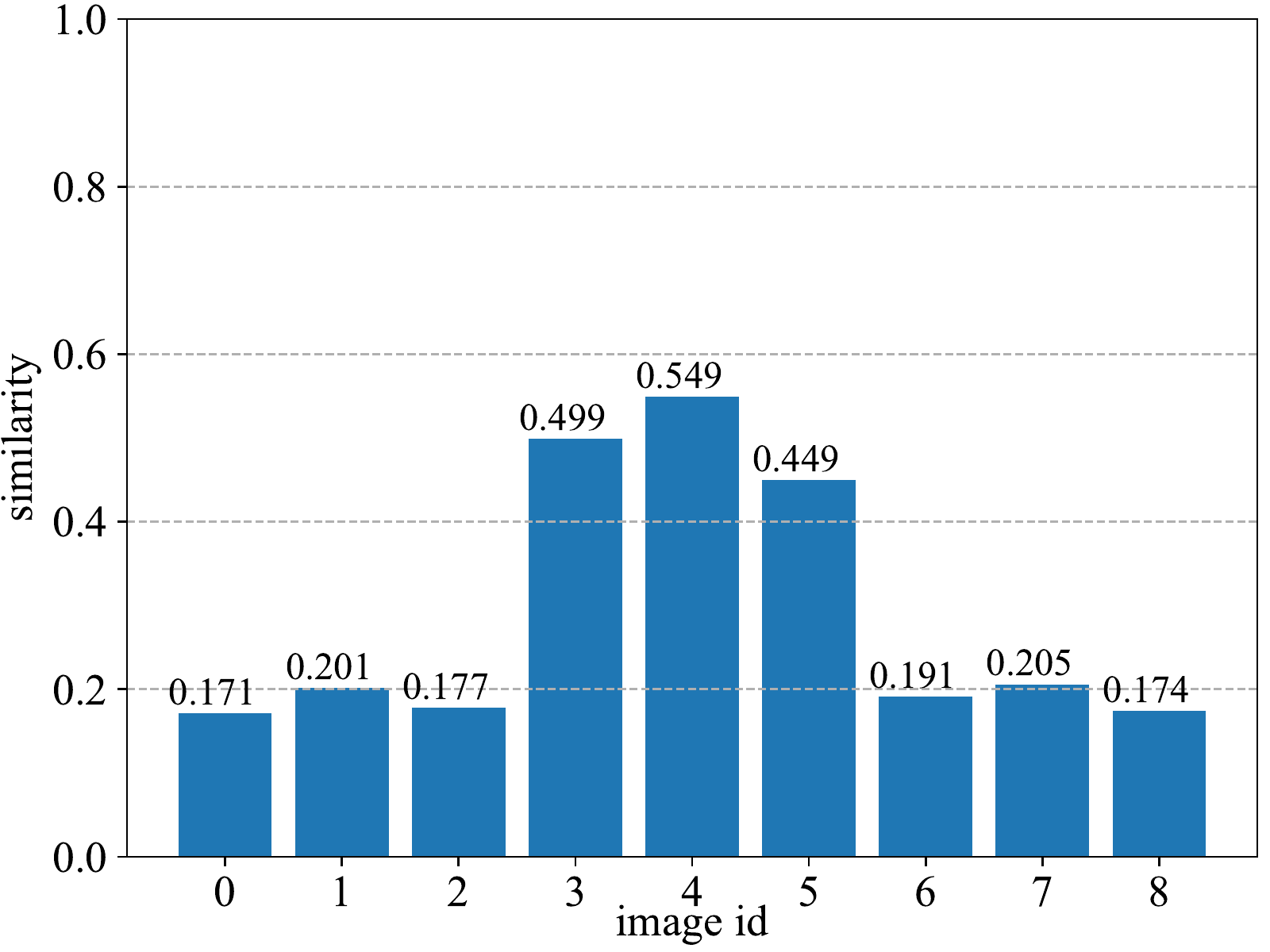} \\

   \includegraphics[width=\linewidth]{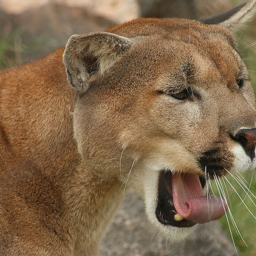}
   & \includegraphics[width=\linewidth]{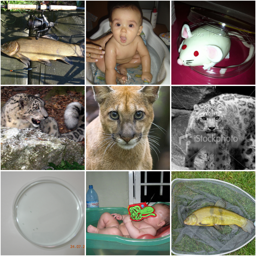}
   & \includegraphics[width=\linewidth]{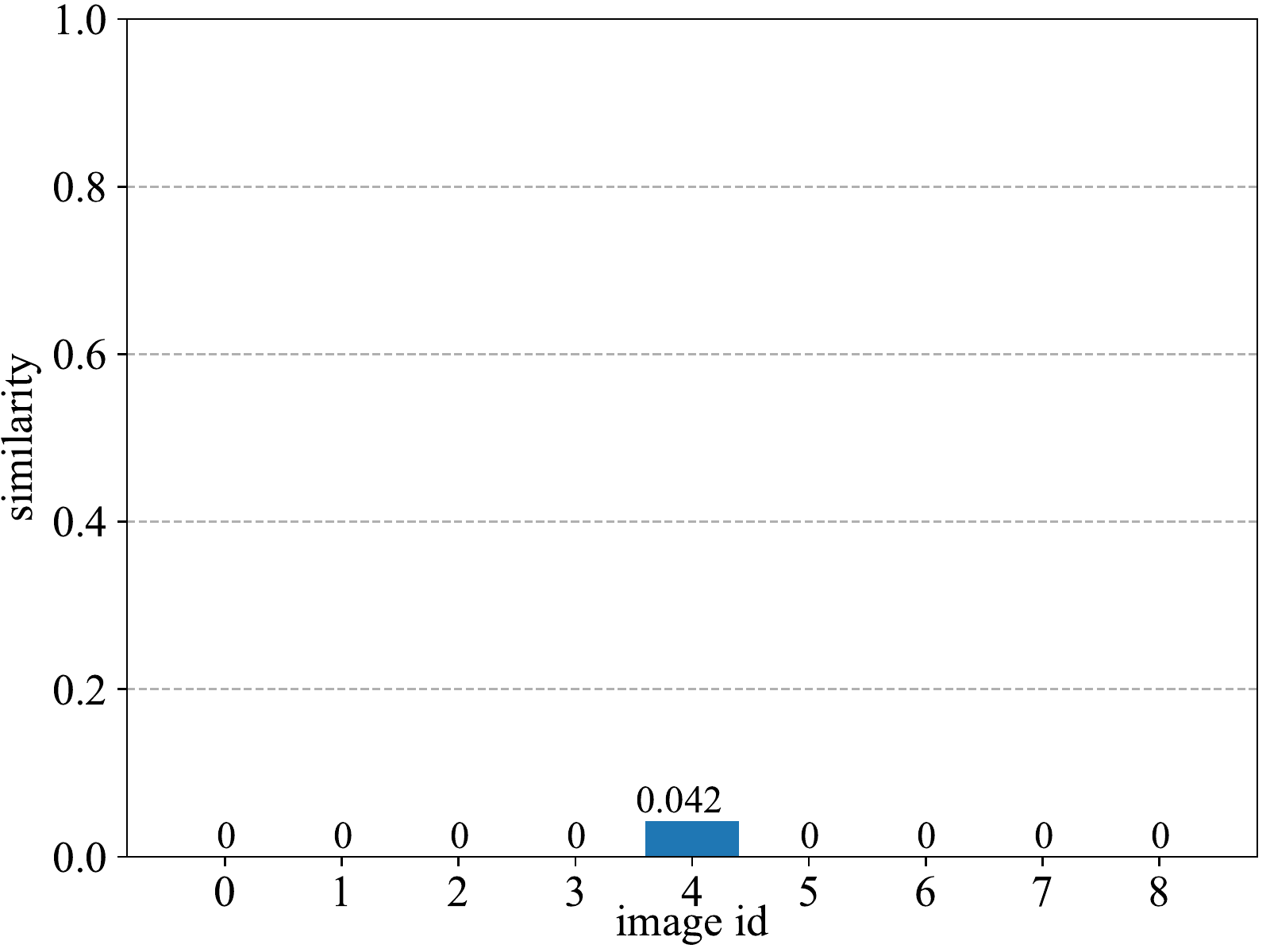}
   & \includegraphics[width=\linewidth]{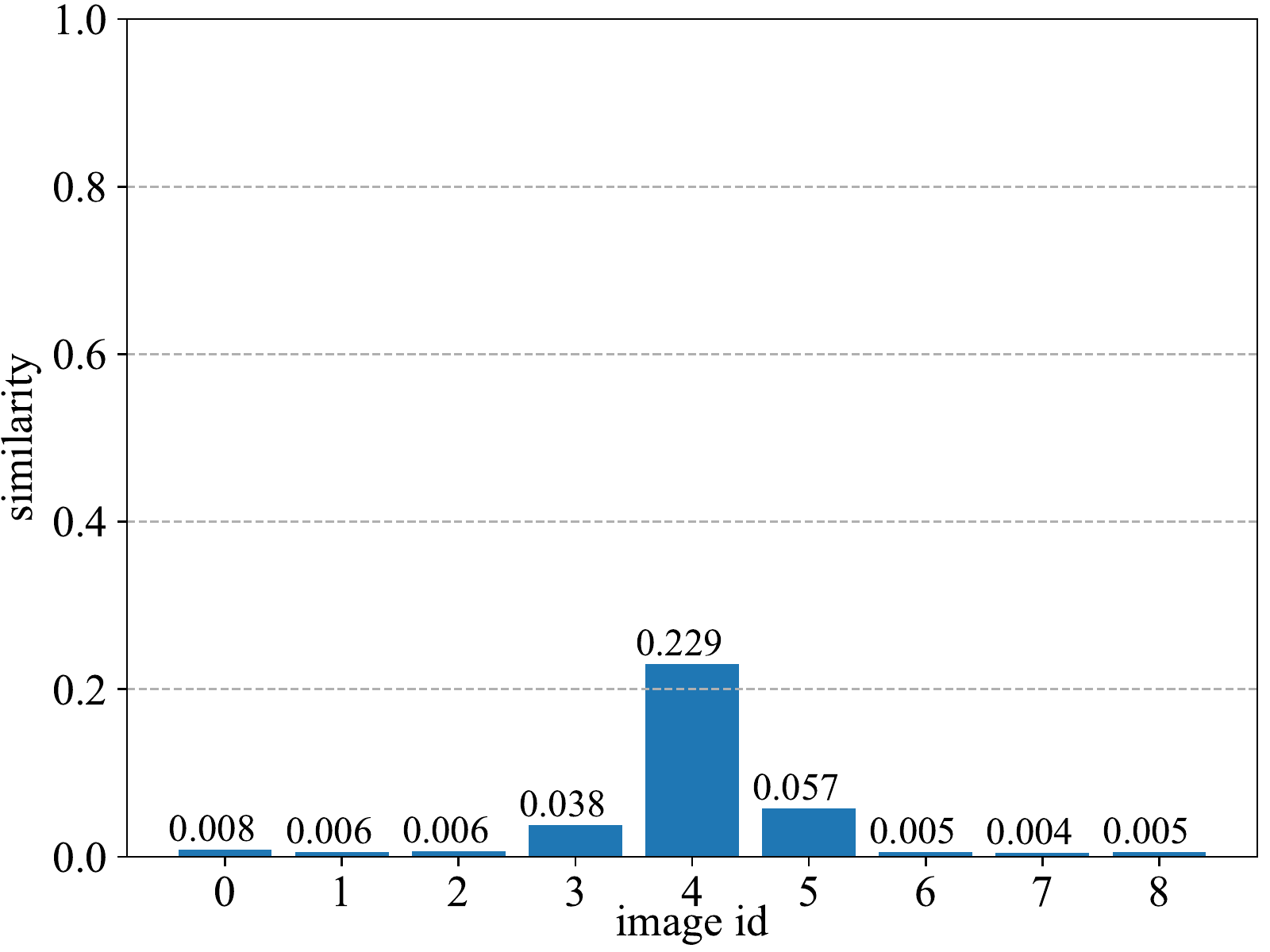}
   & \includegraphics[width=\linewidth]{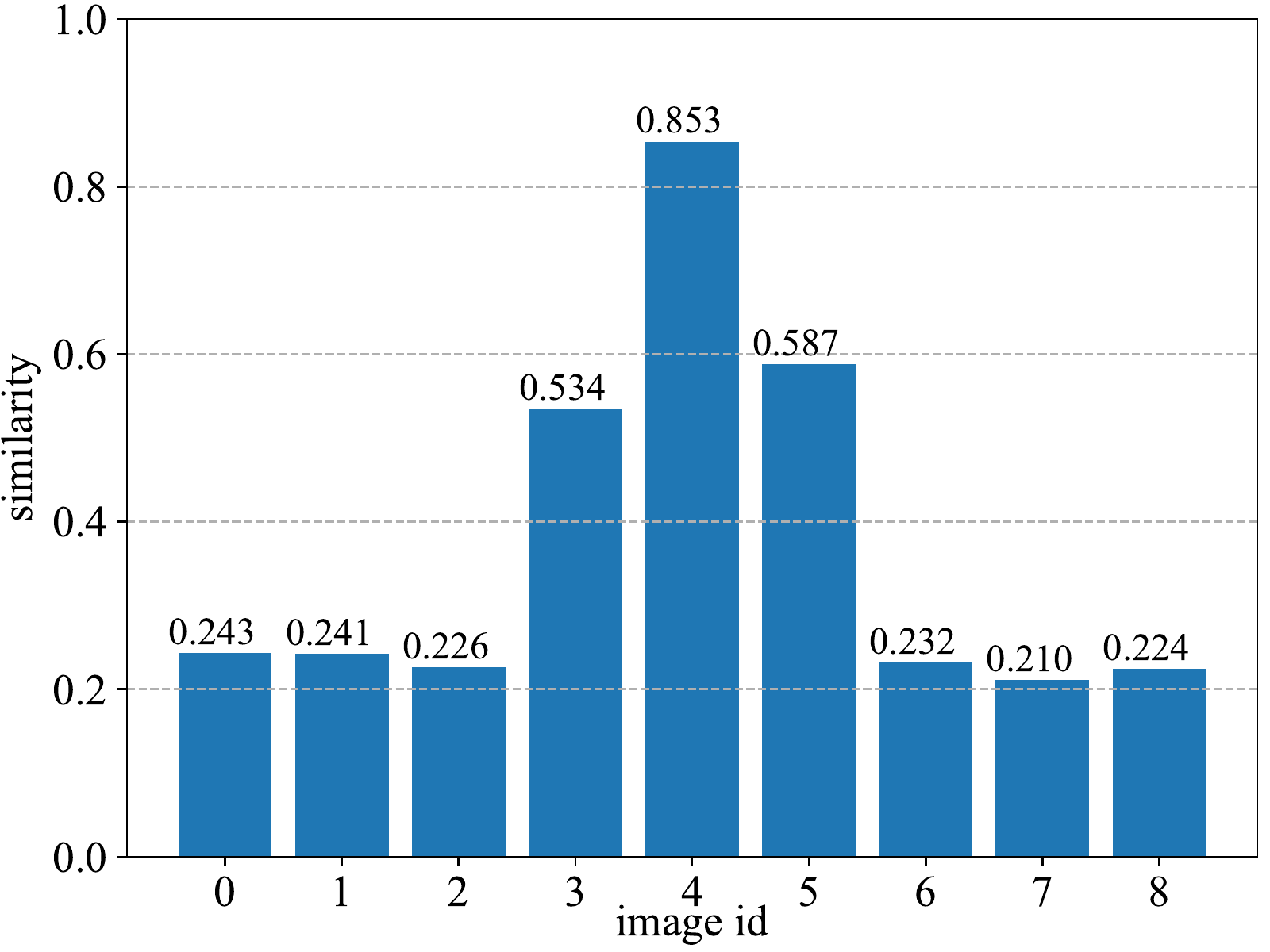} \\

\end{tabular}
}
   \caption{The visualization of inter-instance similarities on ImageNet-1K.
   The query sample and the image with id 4 in key samples are from the same category. 
   The images with id 3 and 5 come from category similar to query sample.
   }
   \label{fig:visualization1}
\end{figure*}

\begin{figure*}[ht]
   \renewcommand\arraystretch{0}
   \renewcommand\tabcolsep{1pt}
   \resizebox{\linewidth}{!}
   {
   \begin{tabular}{m{3.5cm}<{\centering} m{3.5cm}<{\centering} m{5cm}<{\centering} m{5cm}<{\centering} m{5cm}<{\centering}}
      \textbf{Query Sample} & \textbf{Key Samples} & \textbf{DINO} & \textbf{SDMP} & \textbf{PatchMix (ours)} \\

   \includegraphics[width=\linewidth]{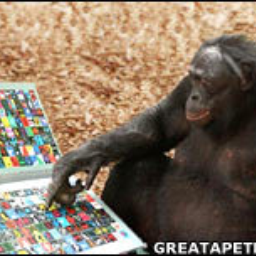}
   & \includegraphics[width=\linewidth]{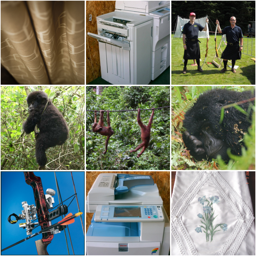}
   & \includegraphics[width=\linewidth]{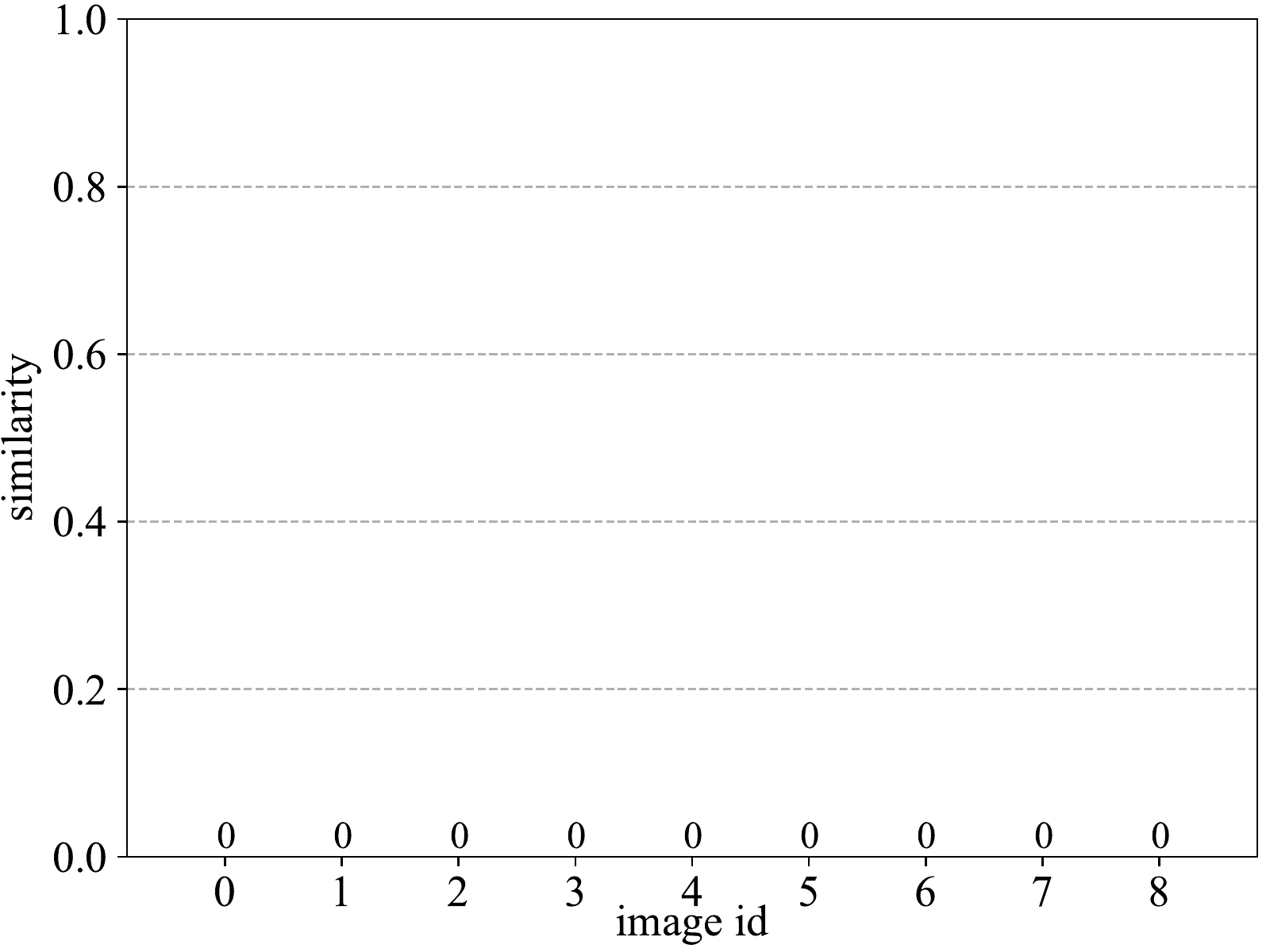}
   & \includegraphics[width=\linewidth]{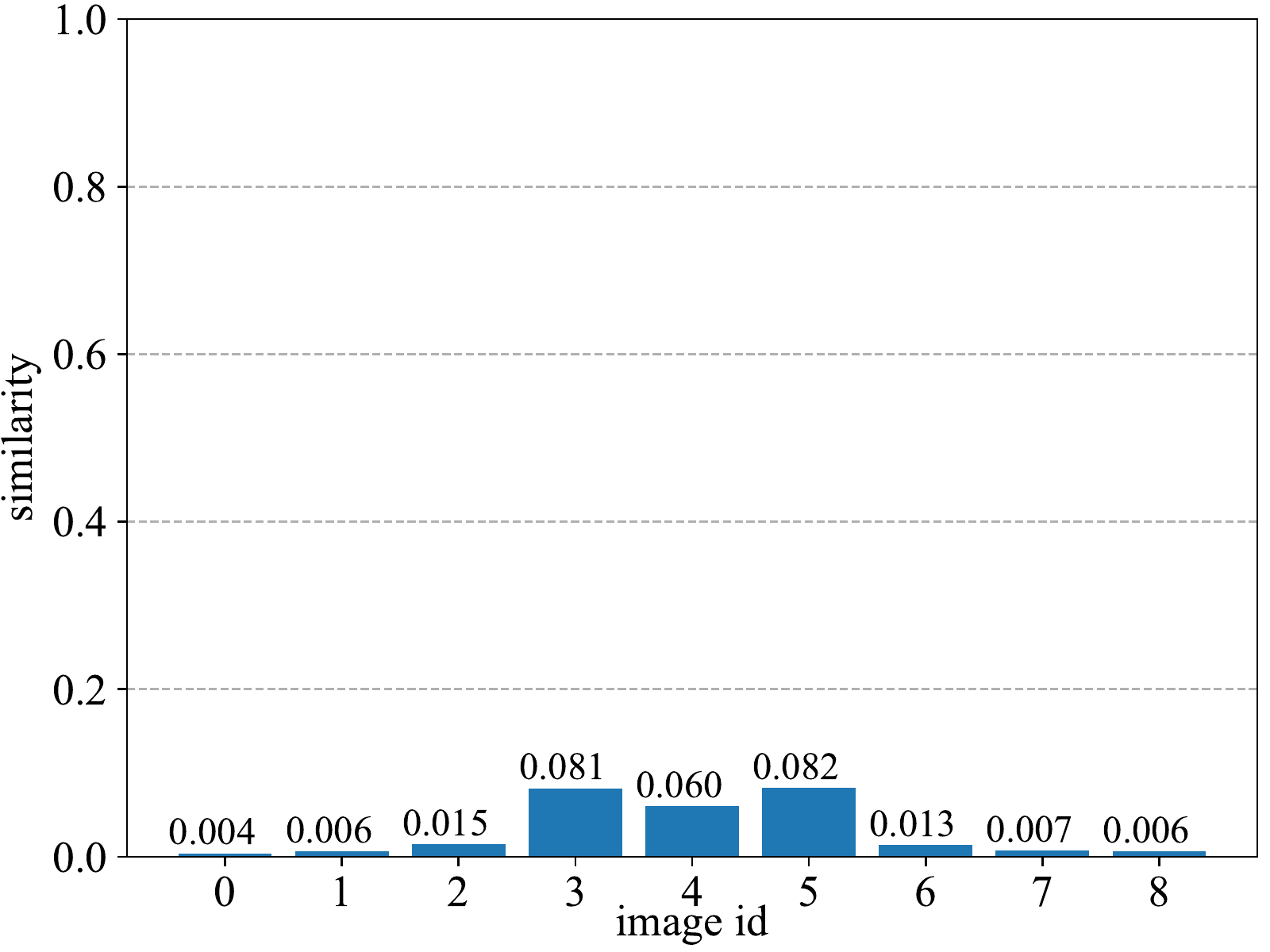}
   & \includegraphics[width=\linewidth]{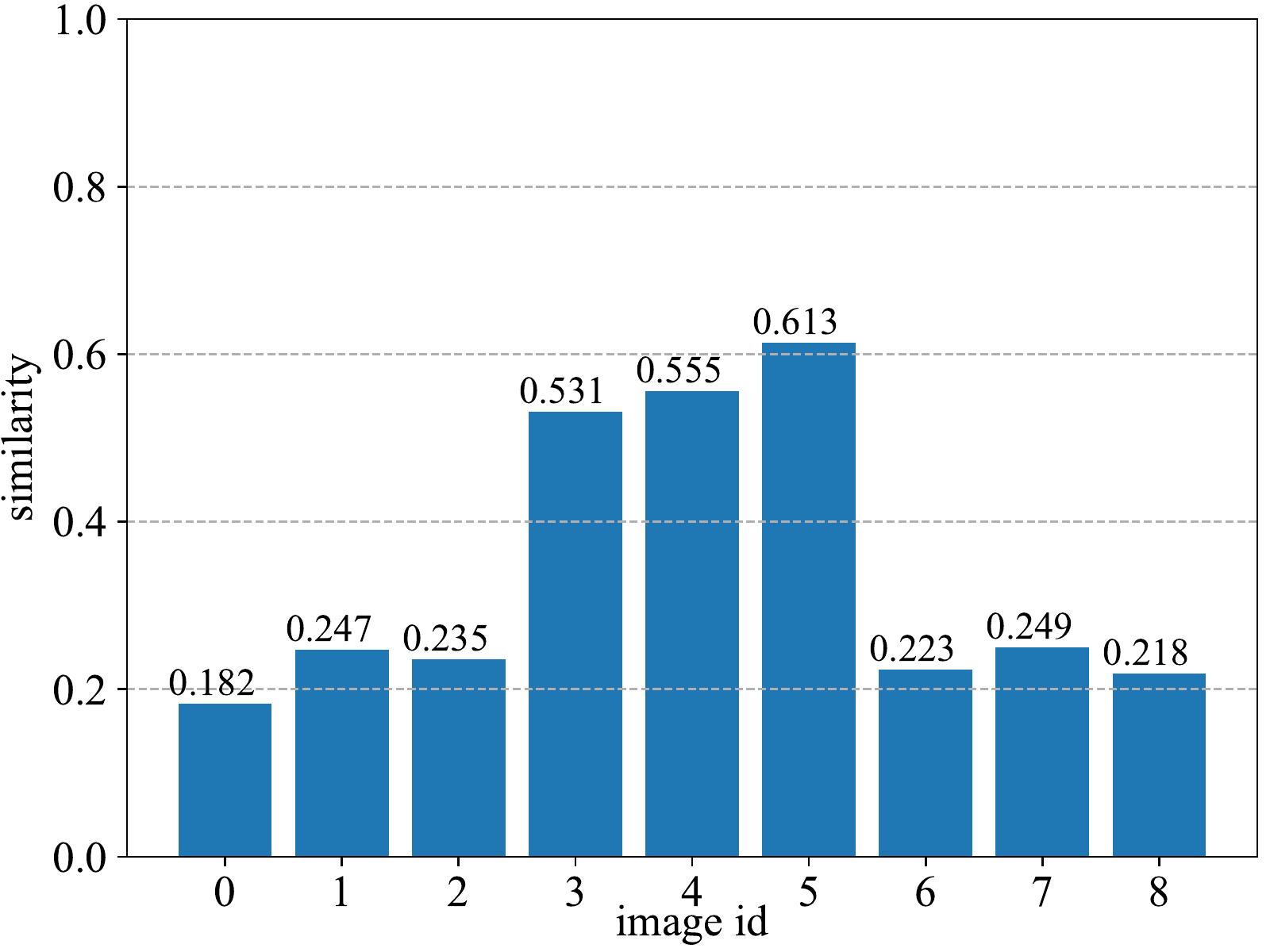} \\

   \includegraphics[width=\linewidth]{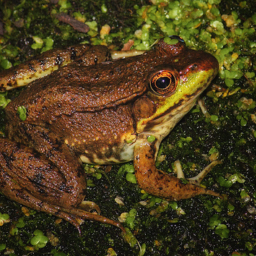}
   & \includegraphics[width=\linewidth]{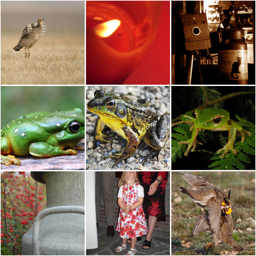}
   & \includegraphics[width=\linewidth]{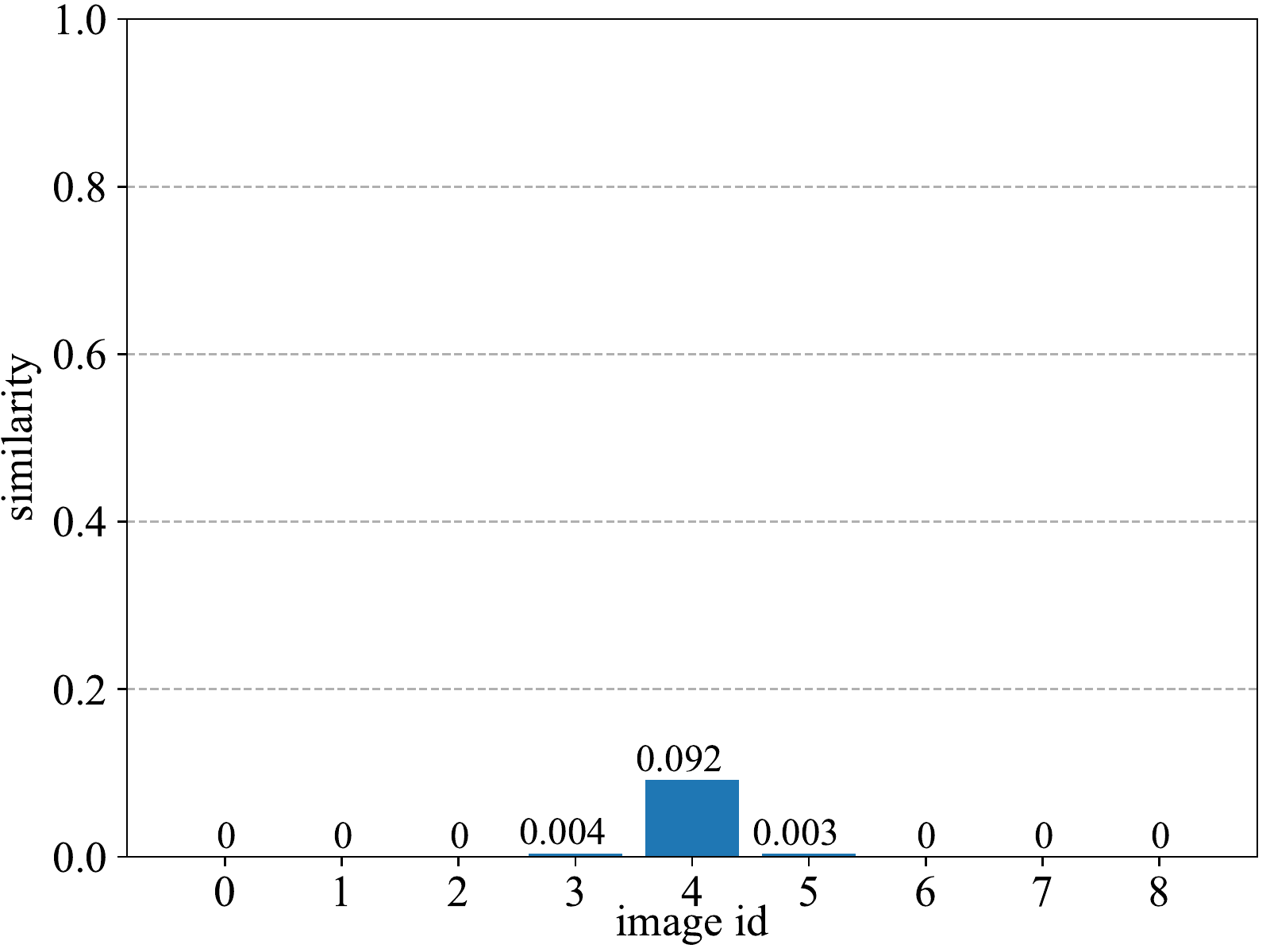}
   & \includegraphics[width=\linewidth]{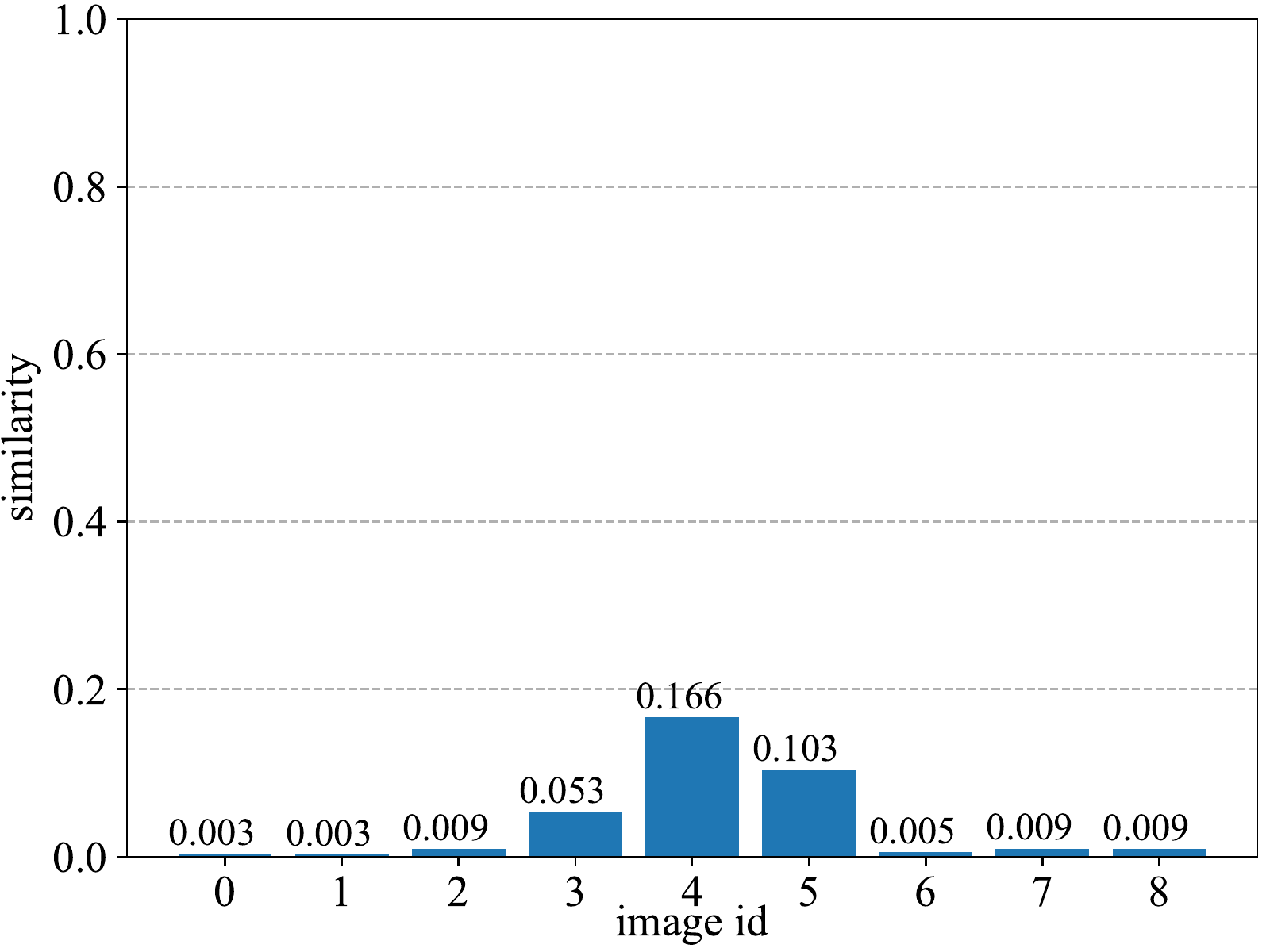}
   & \includegraphics[width=\linewidth]{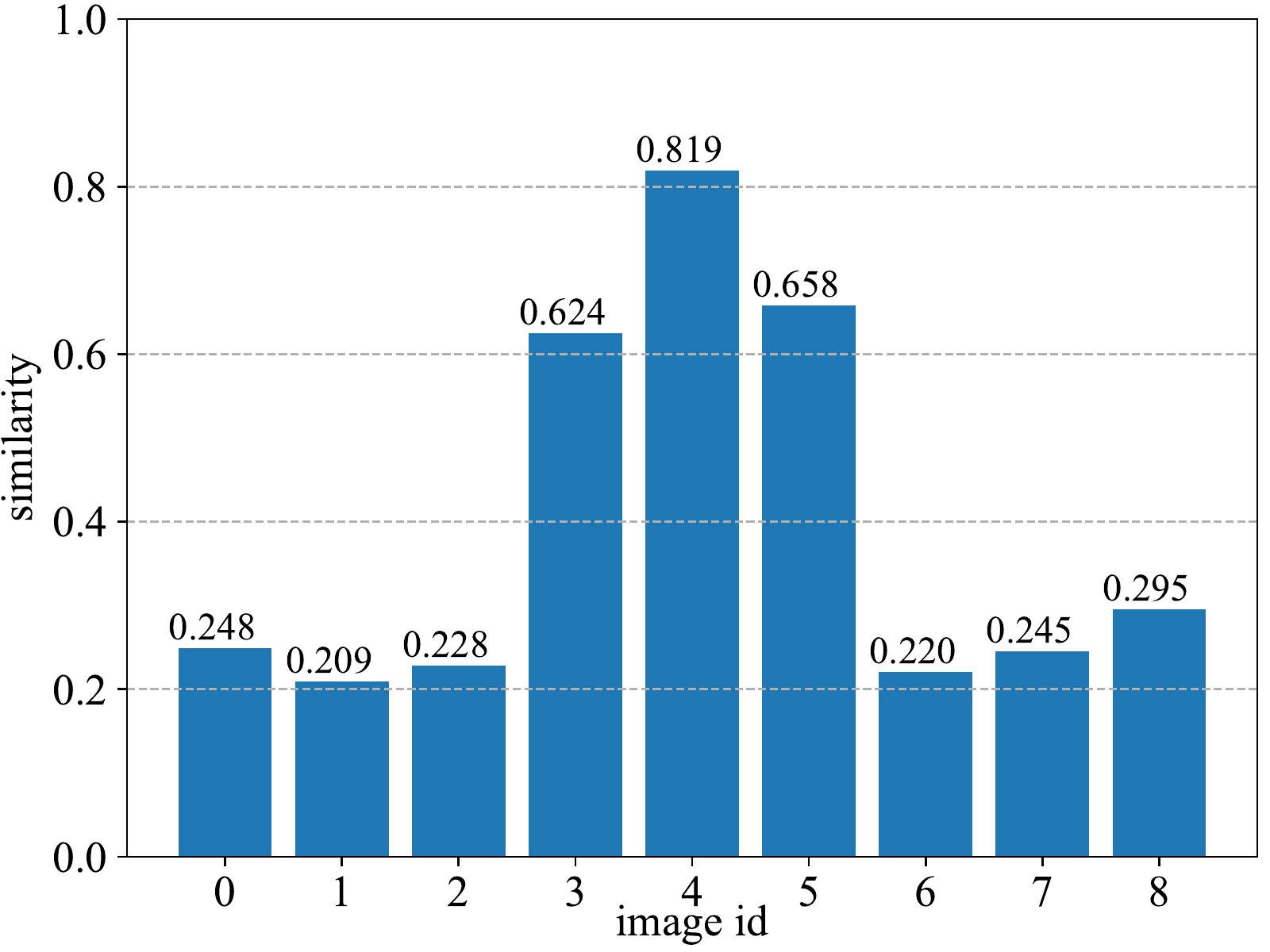} \\

   \includegraphics[width=\linewidth]{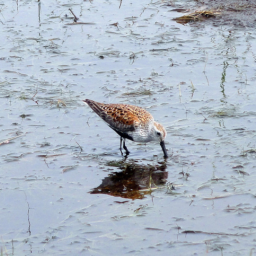}
   & \includegraphics[width=\linewidth]{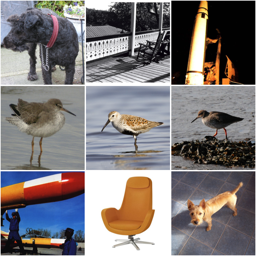}
   & \includegraphics[width=\linewidth]{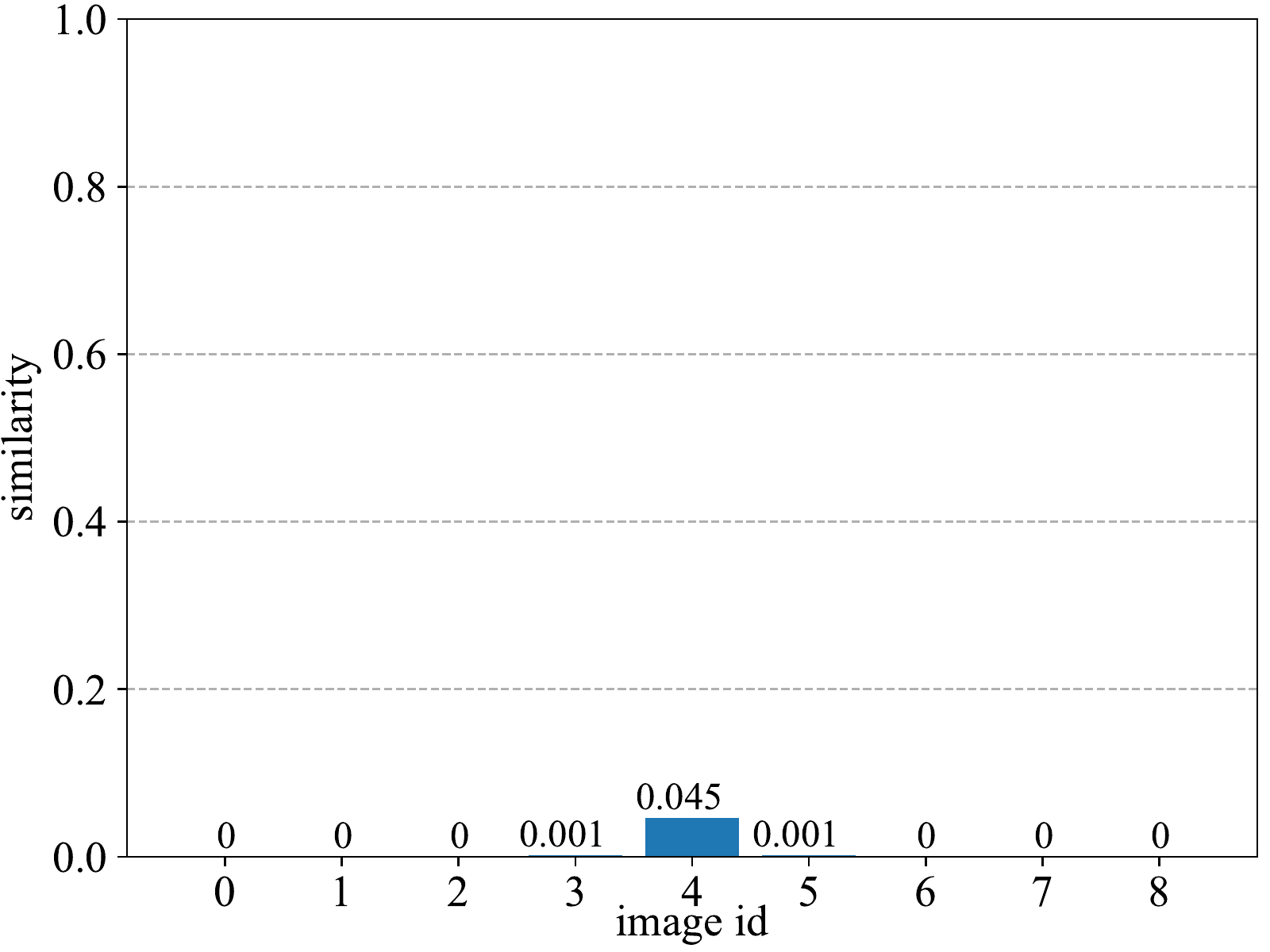}
   & \includegraphics[width=\linewidth]{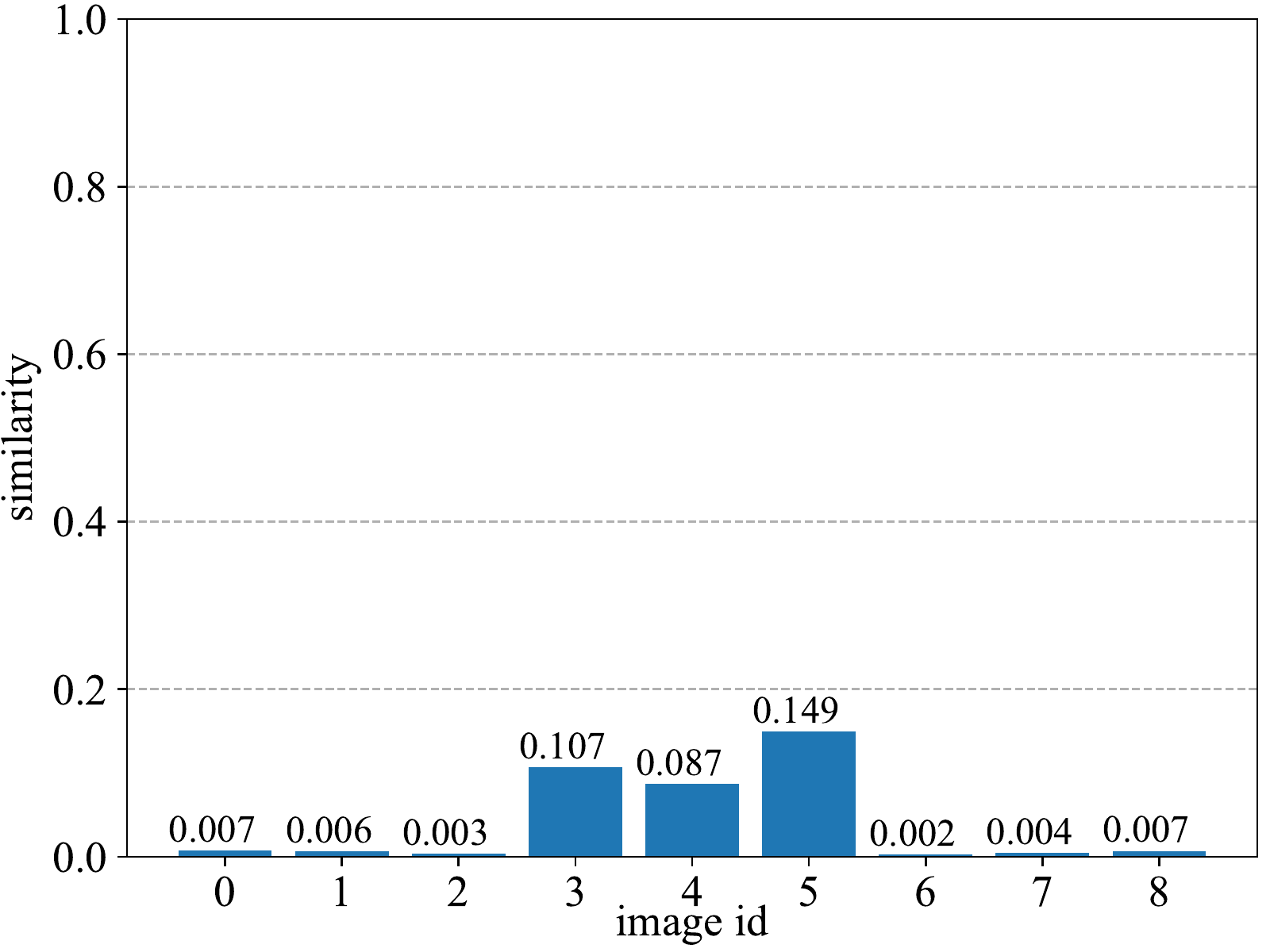}
   & \includegraphics[width=\linewidth]{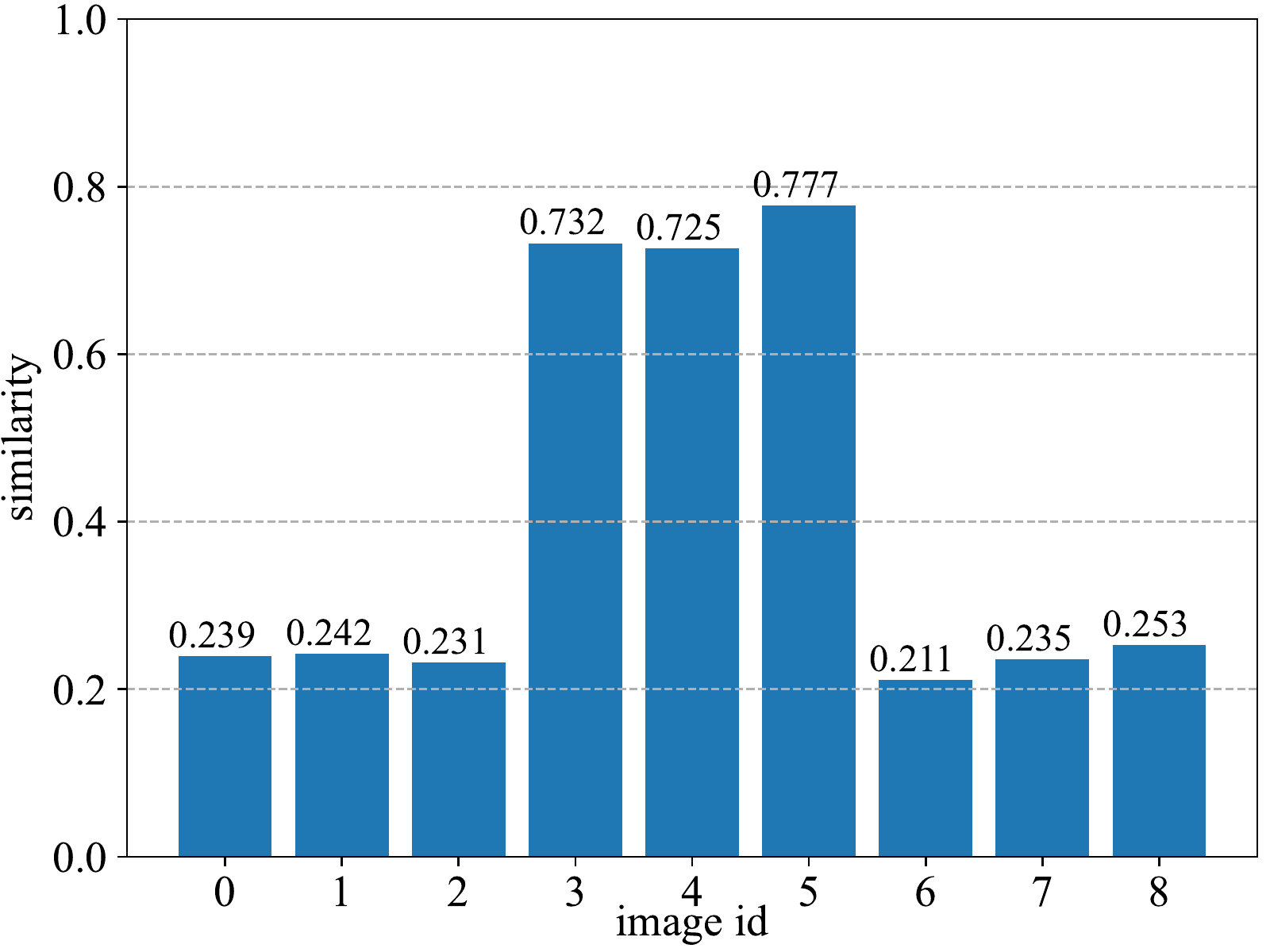} \\
 
   \includegraphics[width=\linewidth]{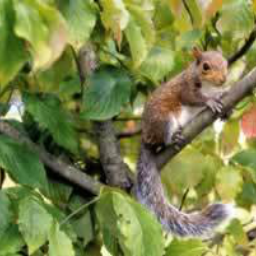}
   & \includegraphics[width=\linewidth]{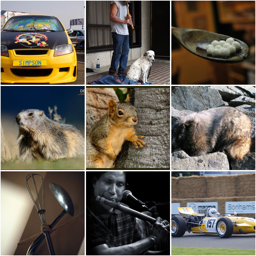}
   & \includegraphics[width=\linewidth]{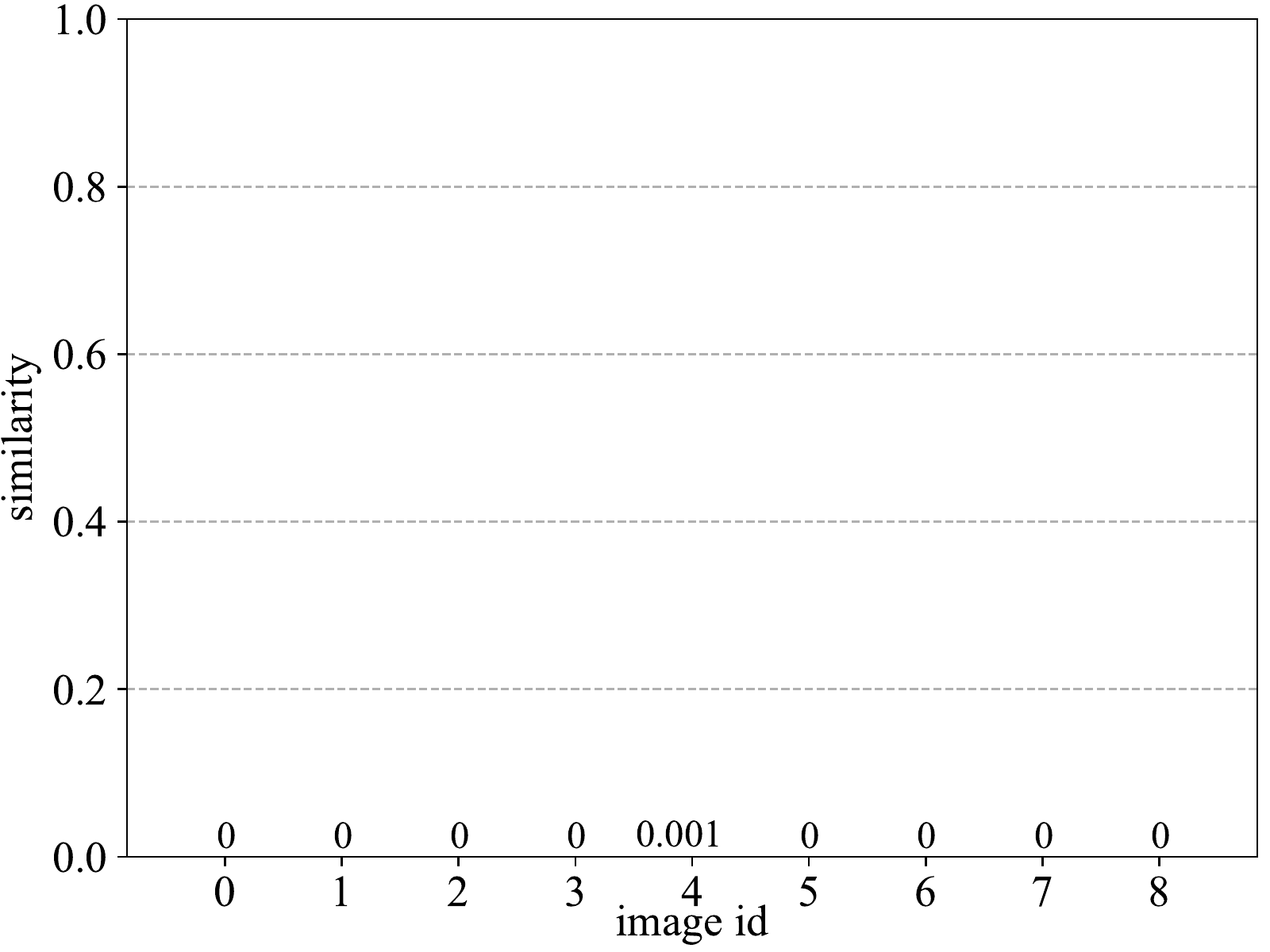}
   & \includegraphics[width=\linewidth]{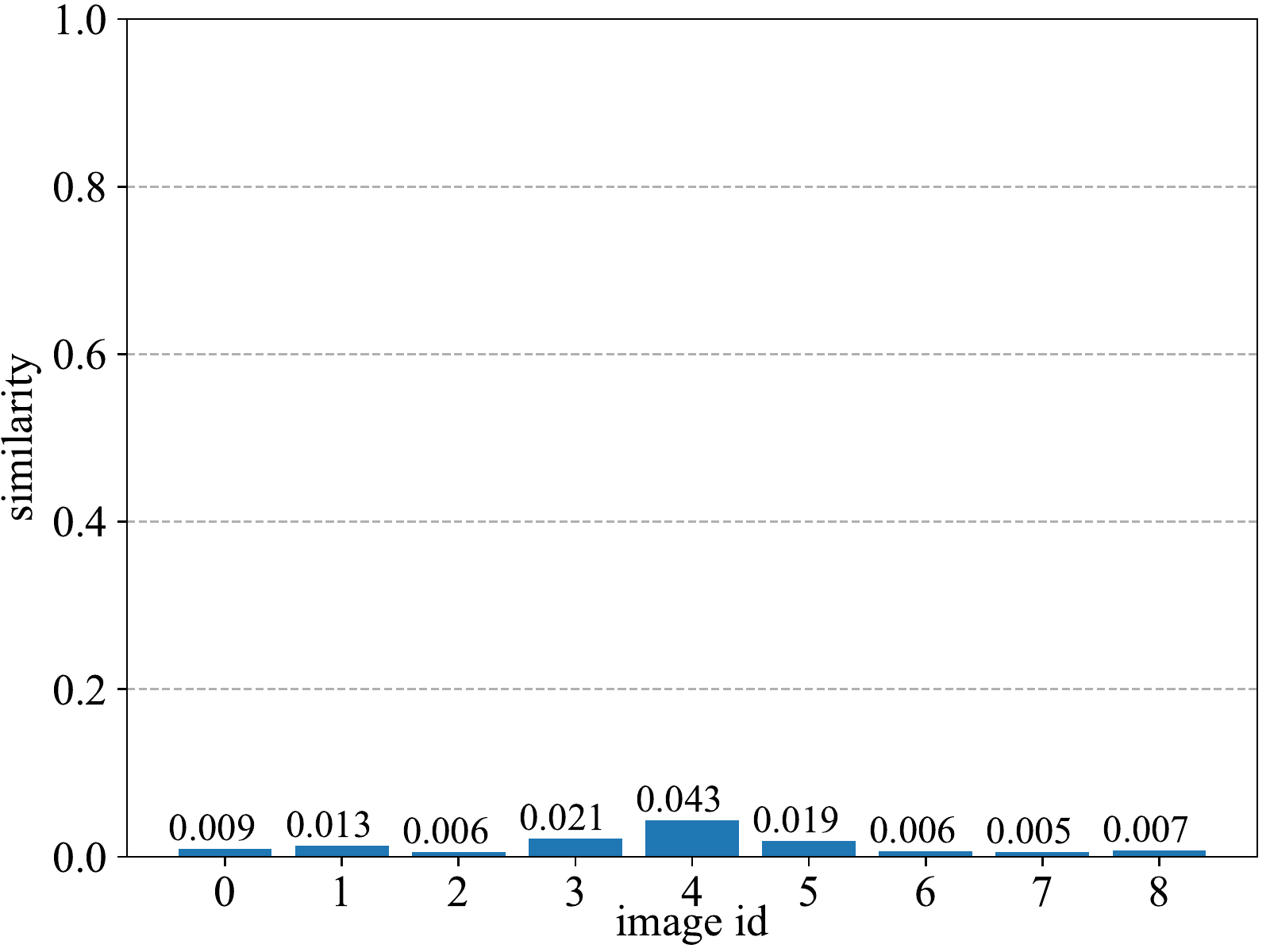}
   & \includegraphics[width=\linewidth]{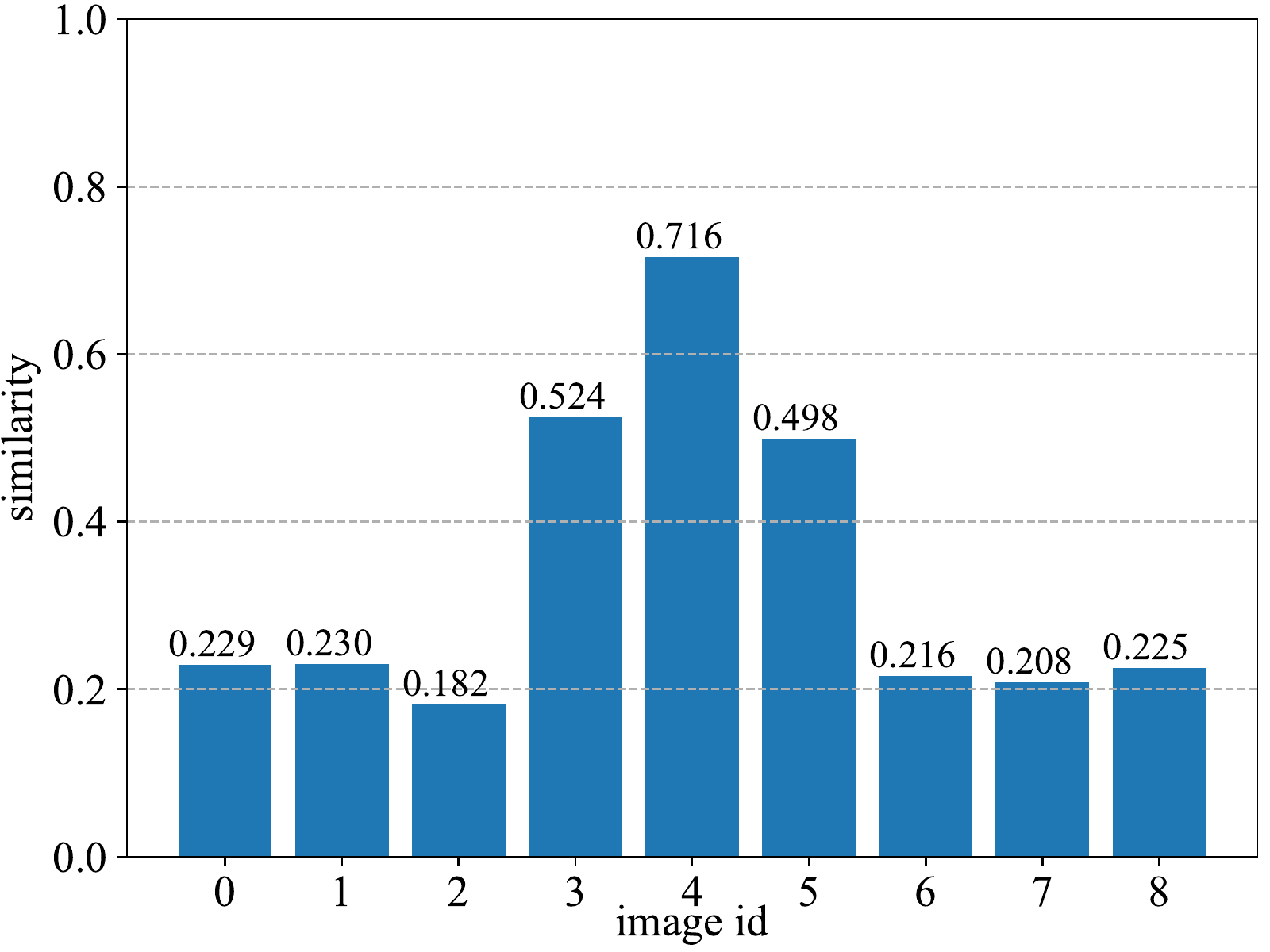} \\

   \includegraphics[width=\linewidth]{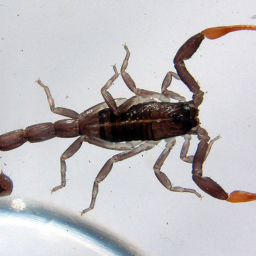}
   & \includegraphics[width=\linewidth]{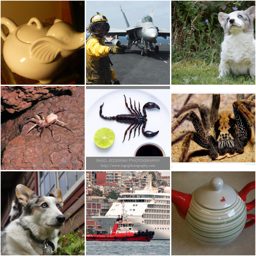}
   & \includegraphics[width=\linewidth]{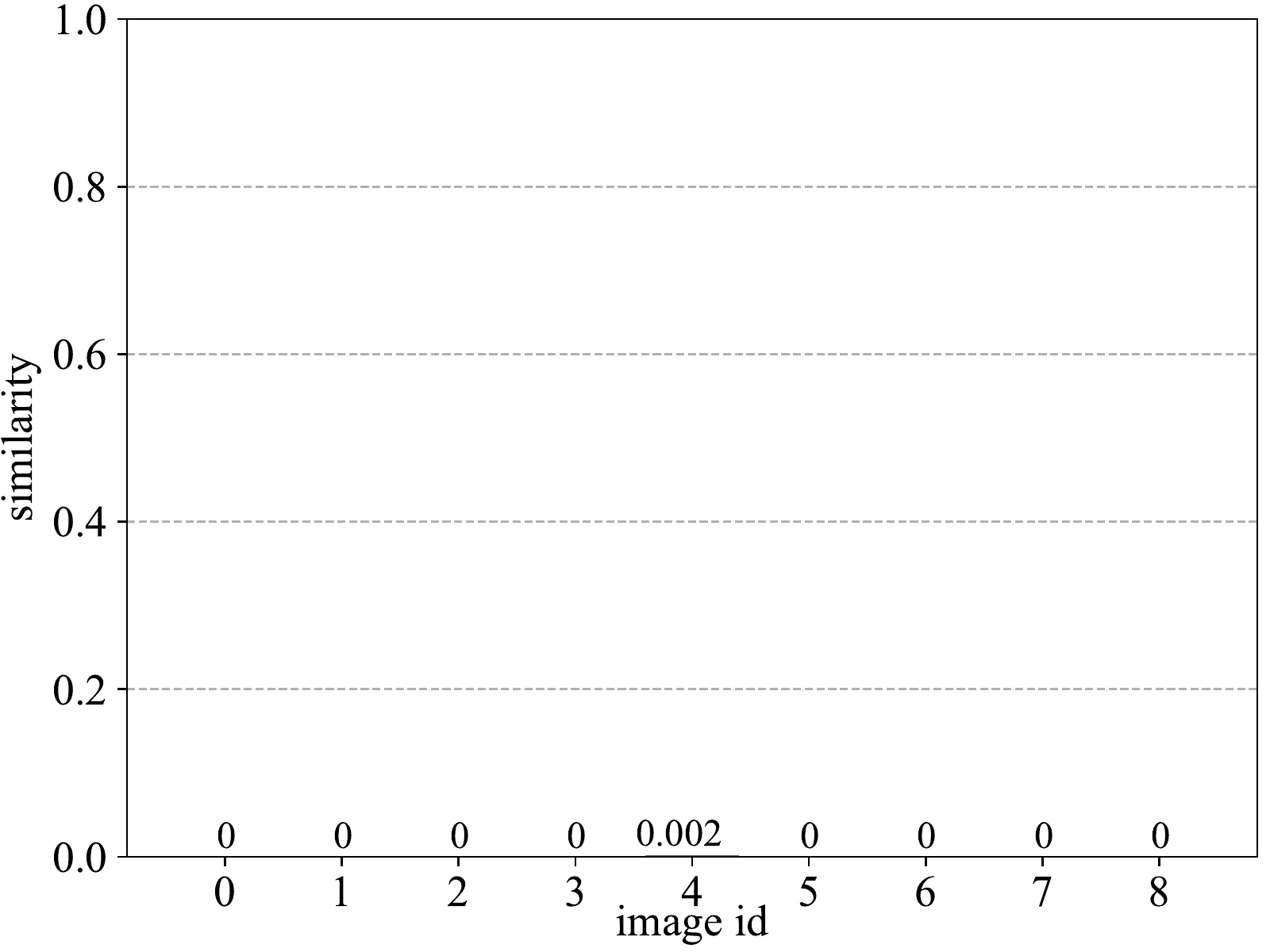}
   & \includegraphics[width=\linewidth]{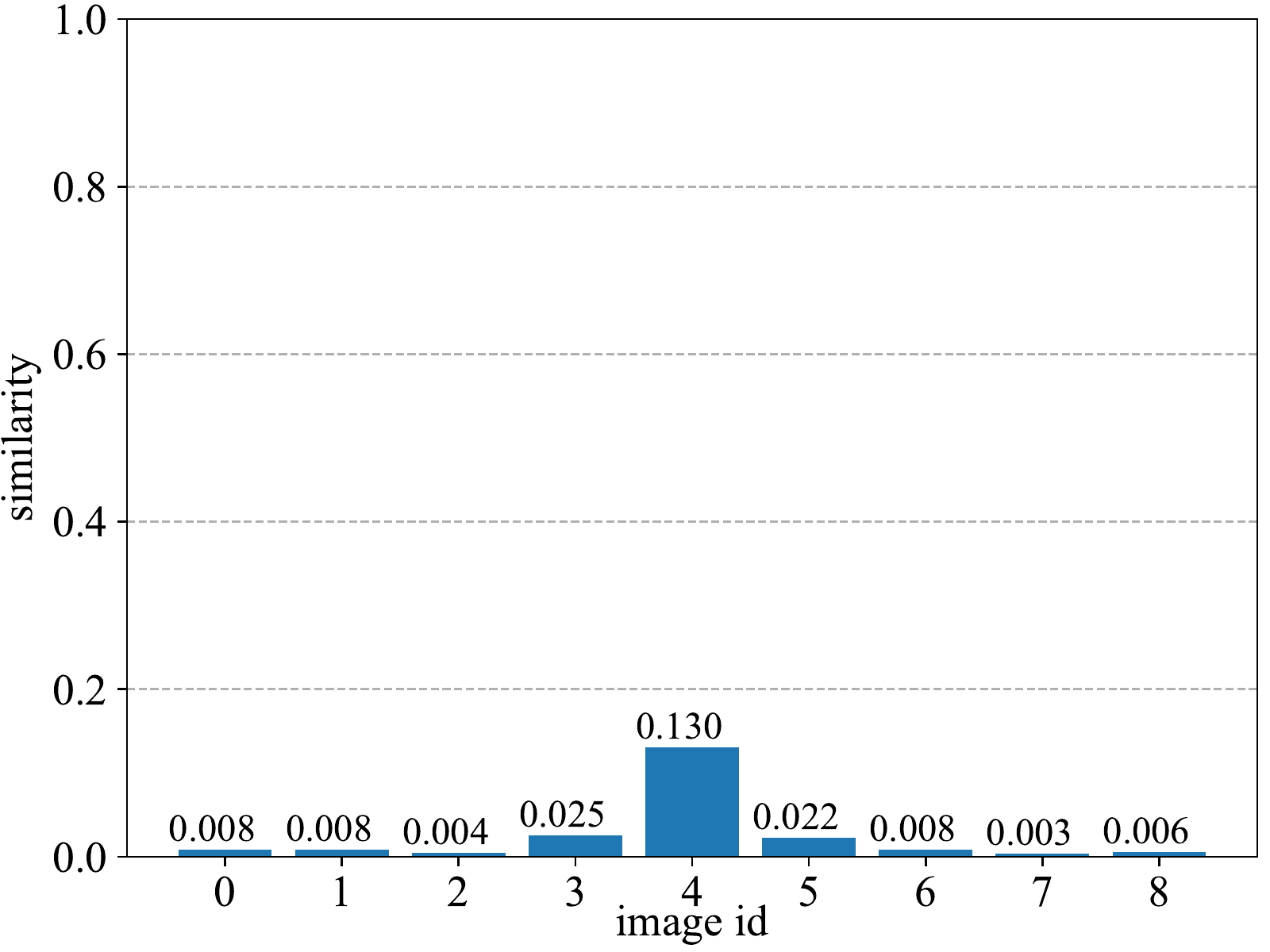}
   & \includegraphics[width=\linewidth]{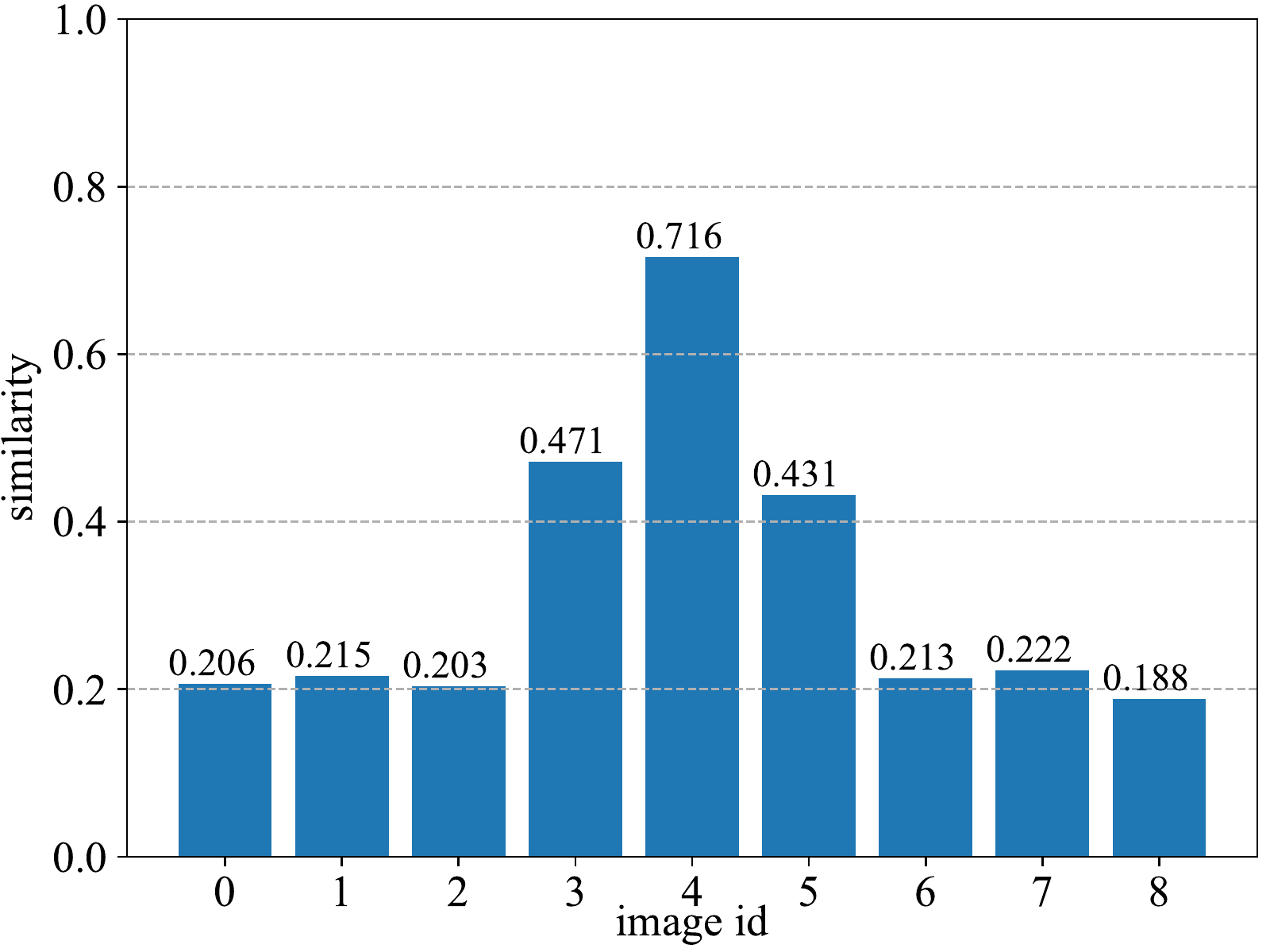} \\

   \includegraphics[width=\linewidth]{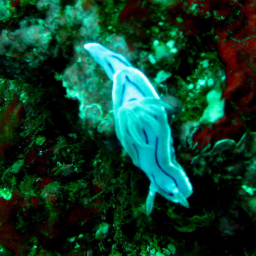}
   & \includegraphics[width=\linewidth]{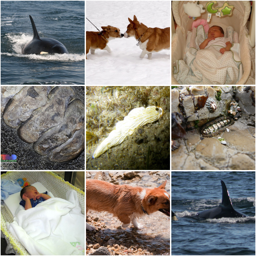}
   & \includegraphics[width=\linewidth]{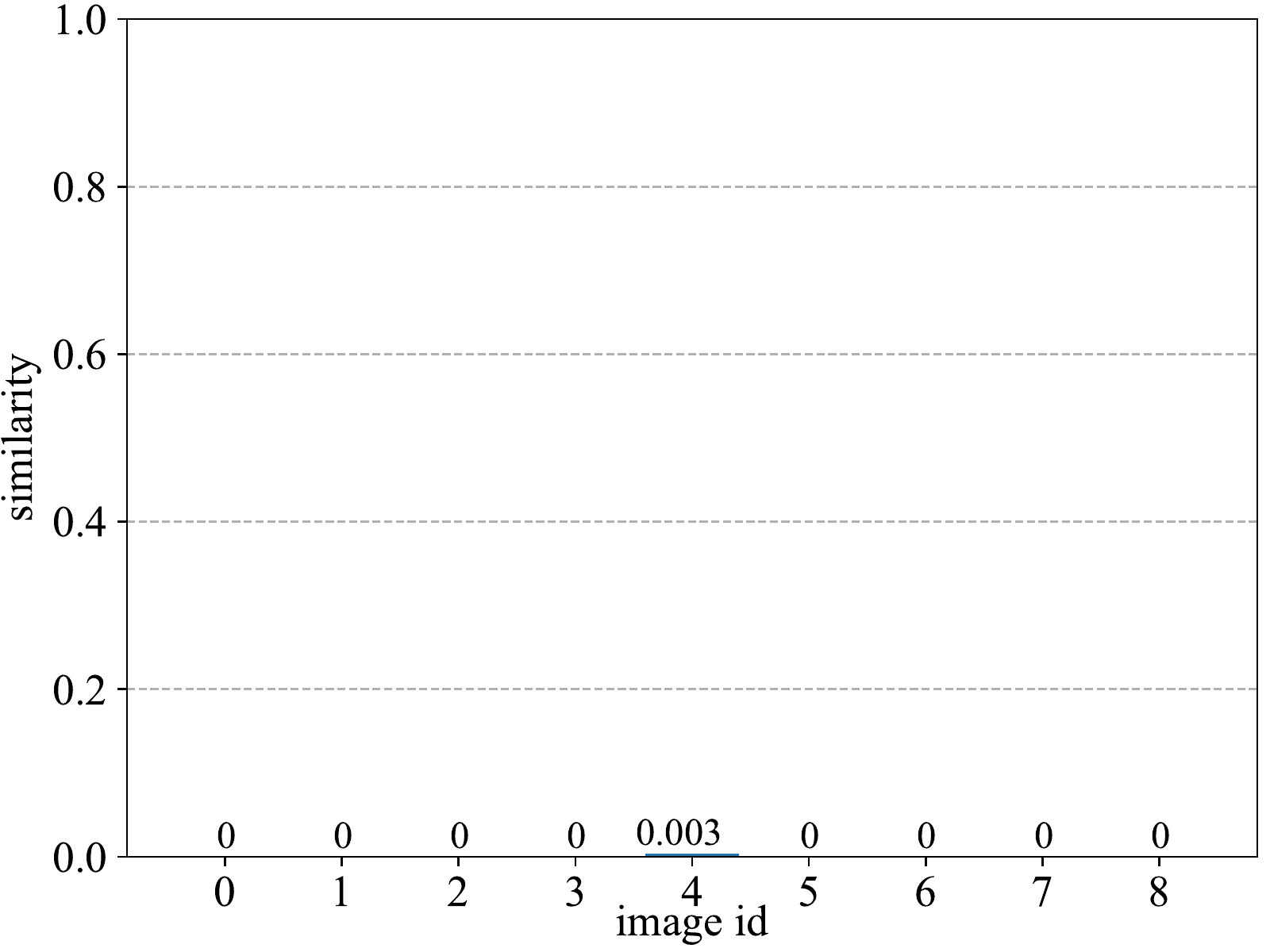}
   & \includegraphics[width=\linewidth]{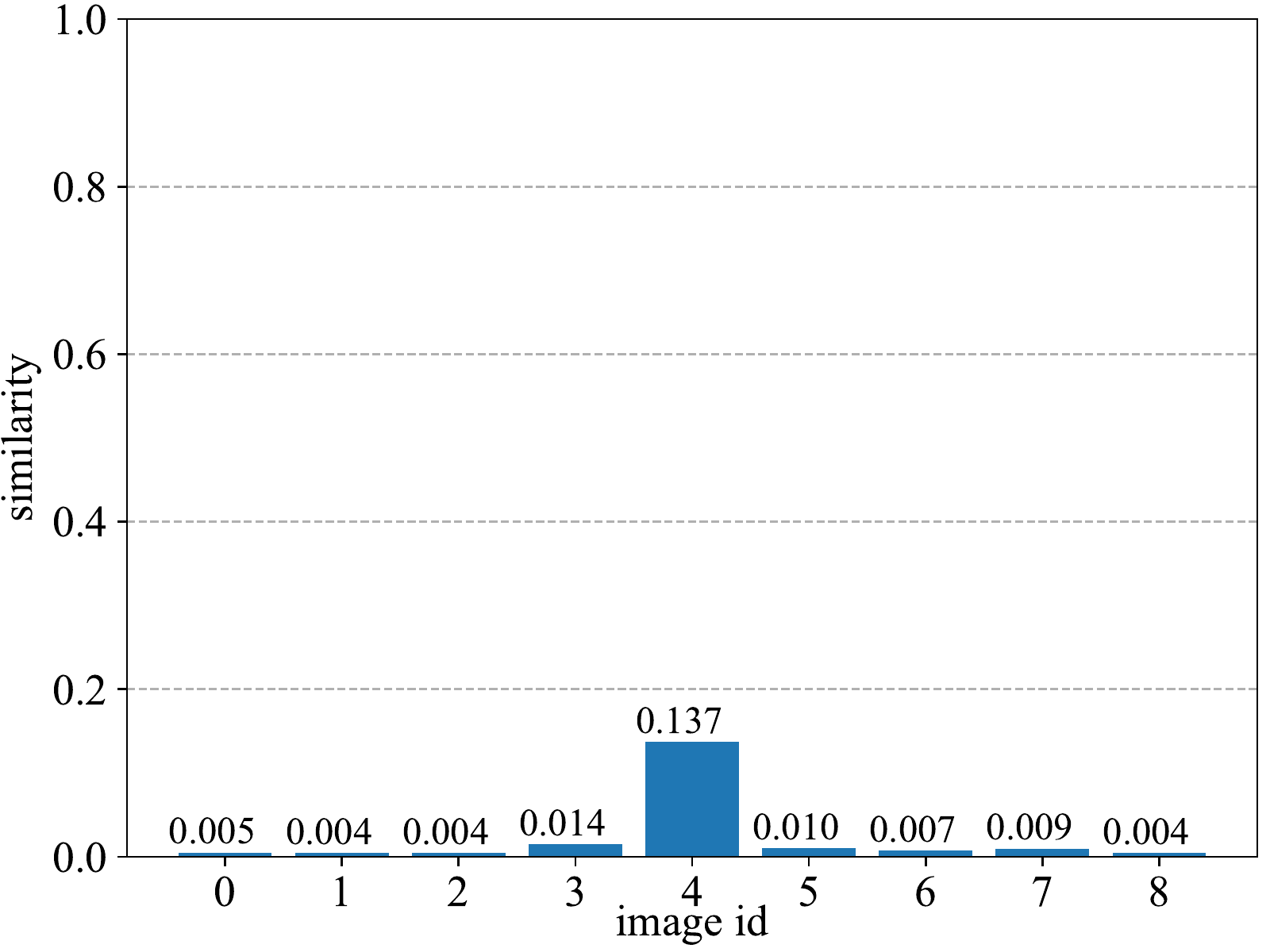}
   & \includegraphics[width=\linewidth]{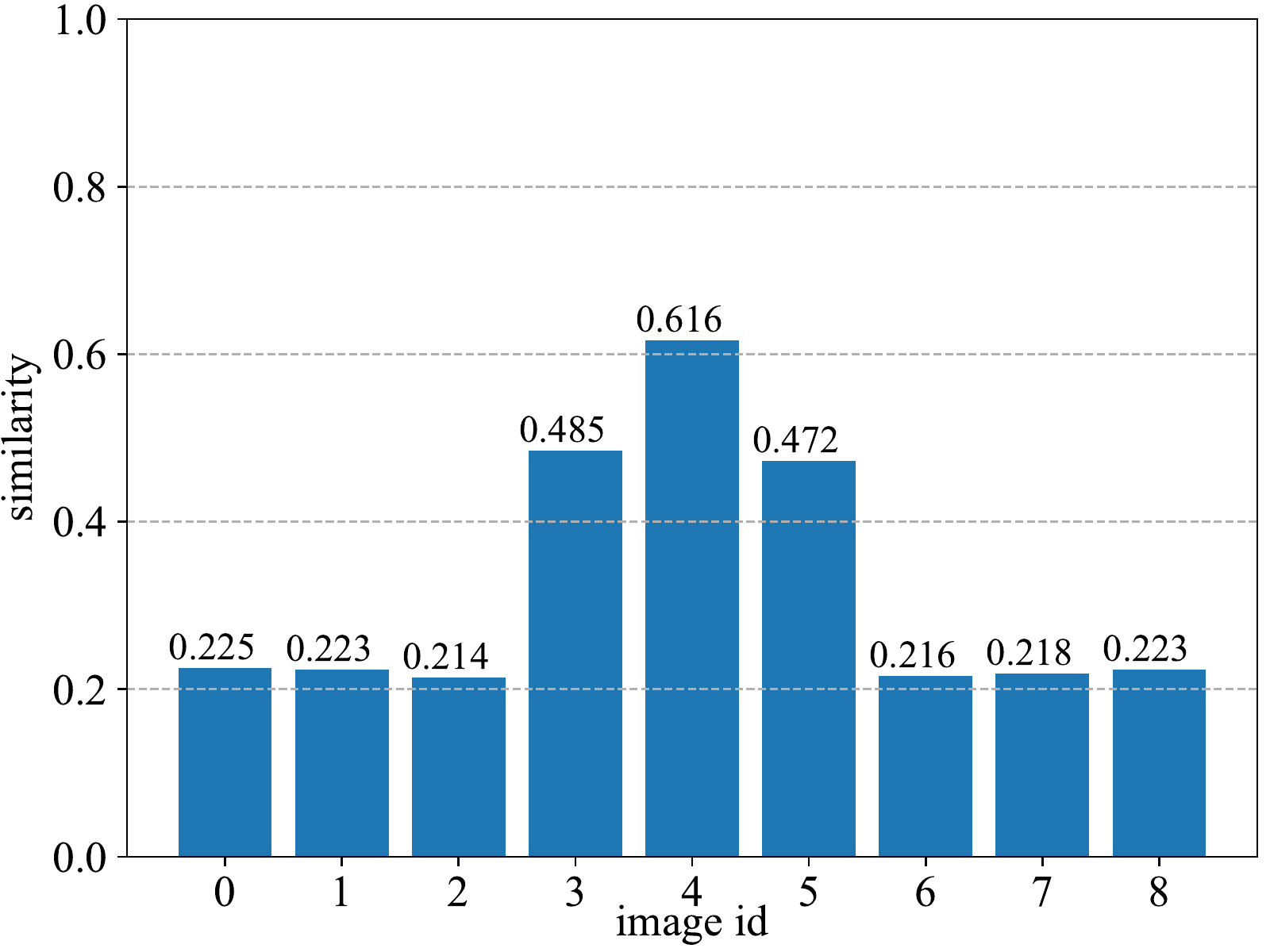} \\
  
   \includegraphics[width=\linewidth]{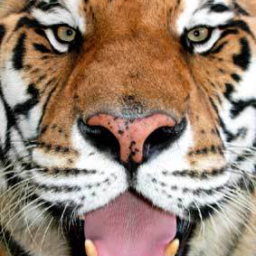}
   & \includegraphics[width=\linewidth]{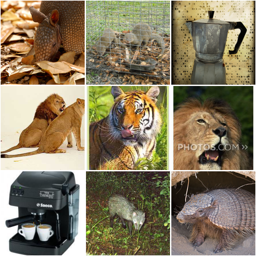}
   & \includegraphics[width=\linewidth]{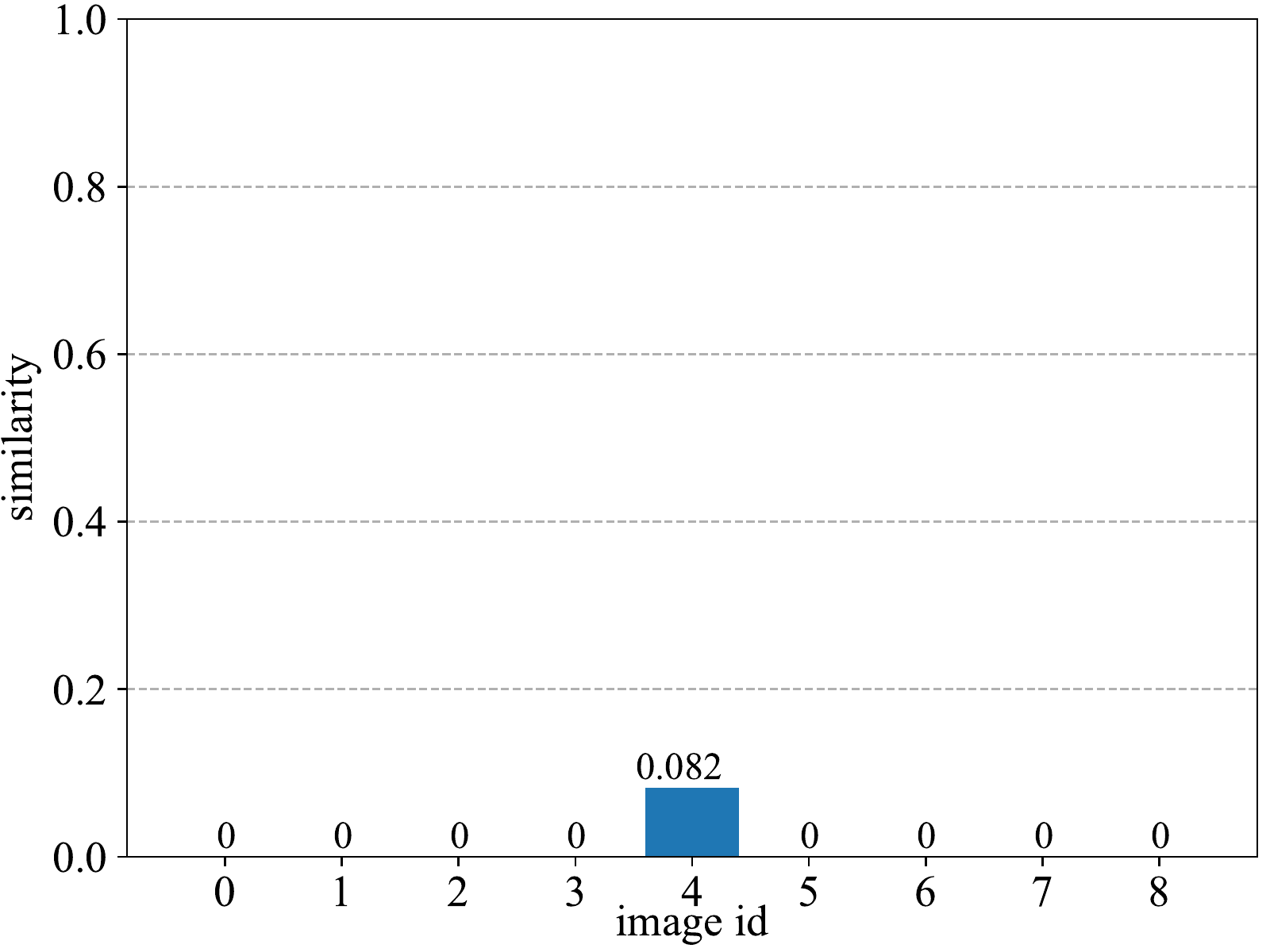}
   & \includegraphics[width=\linewidth]{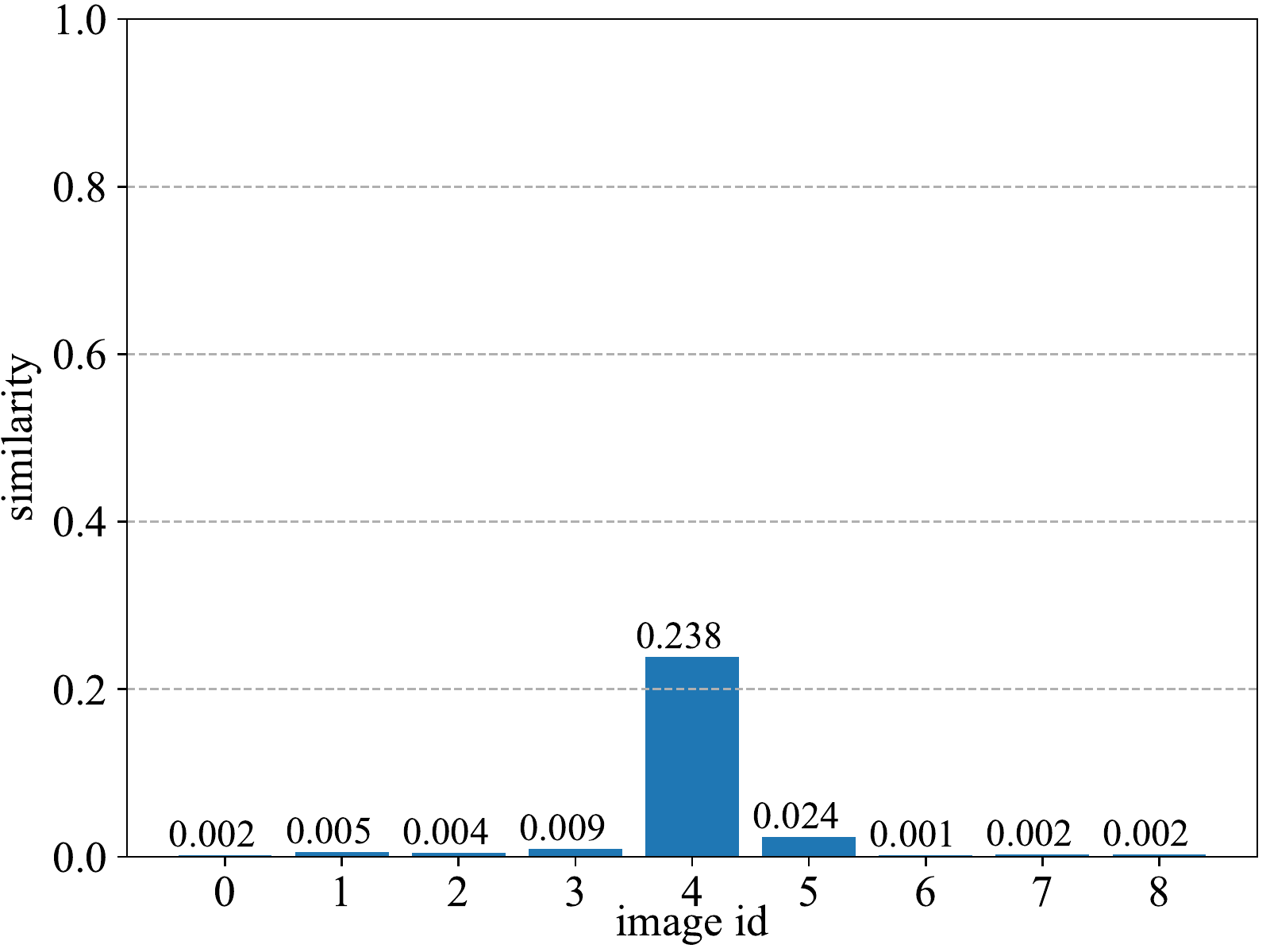}
   & \includegraphics[width=\linewidth]{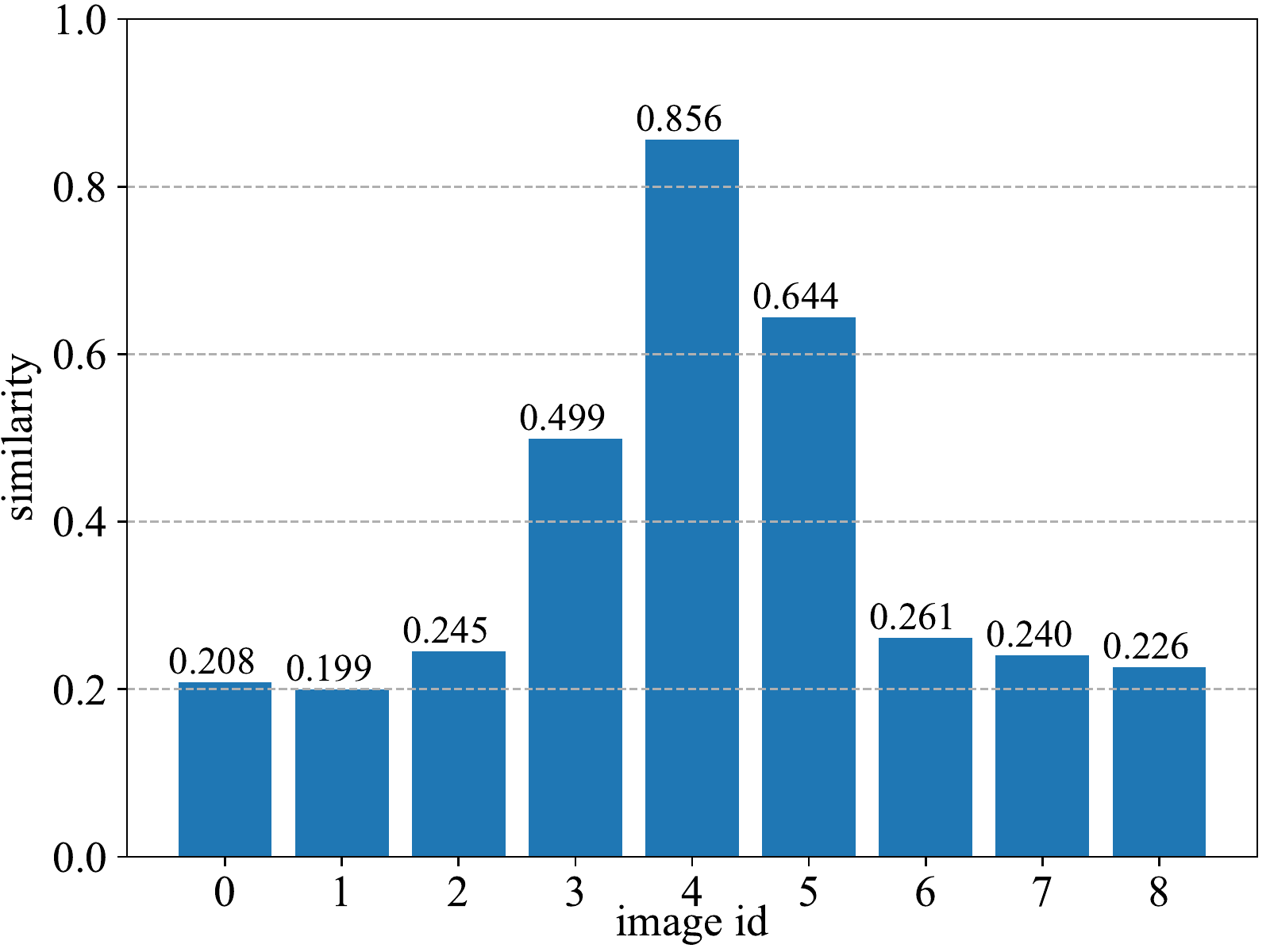} \\

\end{tabular}
}
   \vspace{-1em}
   \caption{The visualization of inter-instance similarities on ImageNet-1K.
   The query sample and the image with id 4 in key samples are from the same category. 
   The images with id 3 and 5 come from category similar to query sample.
   }
   \label{fig:visualization2}
   \vspace{-1.5em}
\end{figure*}

\begin{figure*}[ht]
   \renewcommand\arraystretch{0}
   \renewcommand\tabcolsep{1pt}
   \resizebox{\linewidth}{!}
   {
   \begin{tabular}{m{3.5cm}<{\centering} m{3.5cm}<{\centering} m{5cm}<{\centering} m{5cm}<{\centering} m{5cm}<{\centering}}
      \textbf{Query Sample} & \textbf{Key Samples} & \textbf{DINO} & \textbf{SDMP} & \textbf{PatchMix (ours)} \\

   \includegraphics[width=\linewidth]{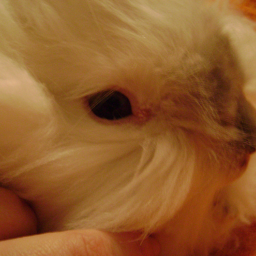}
   & \includegraphics[width=\linewidth]{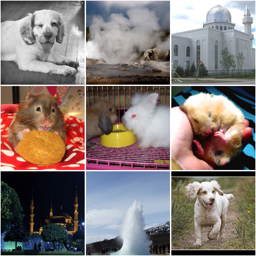}
   & \includegraphics[width=\linewidth]{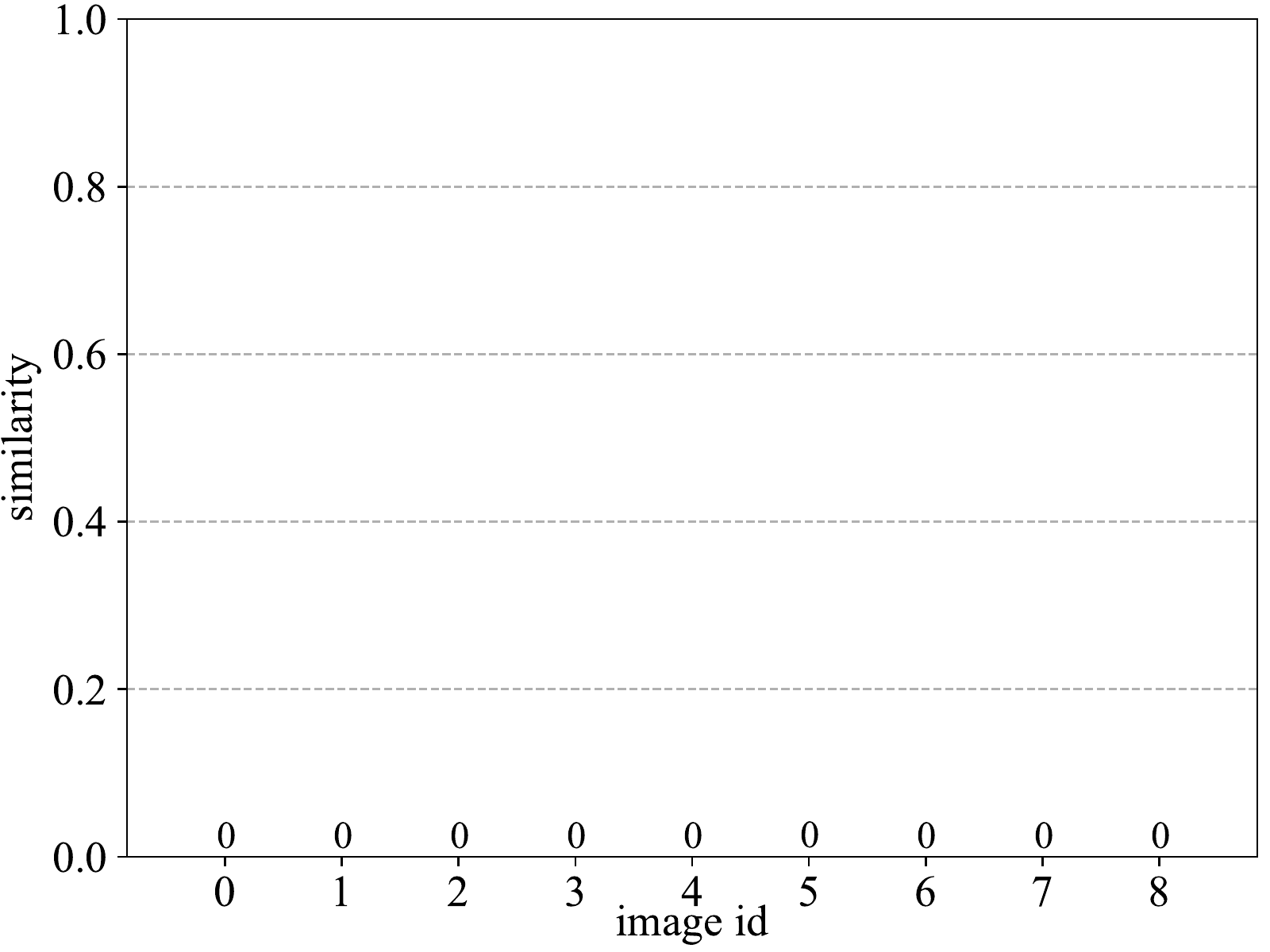}
   & \includegraphics[width=\linewidth]{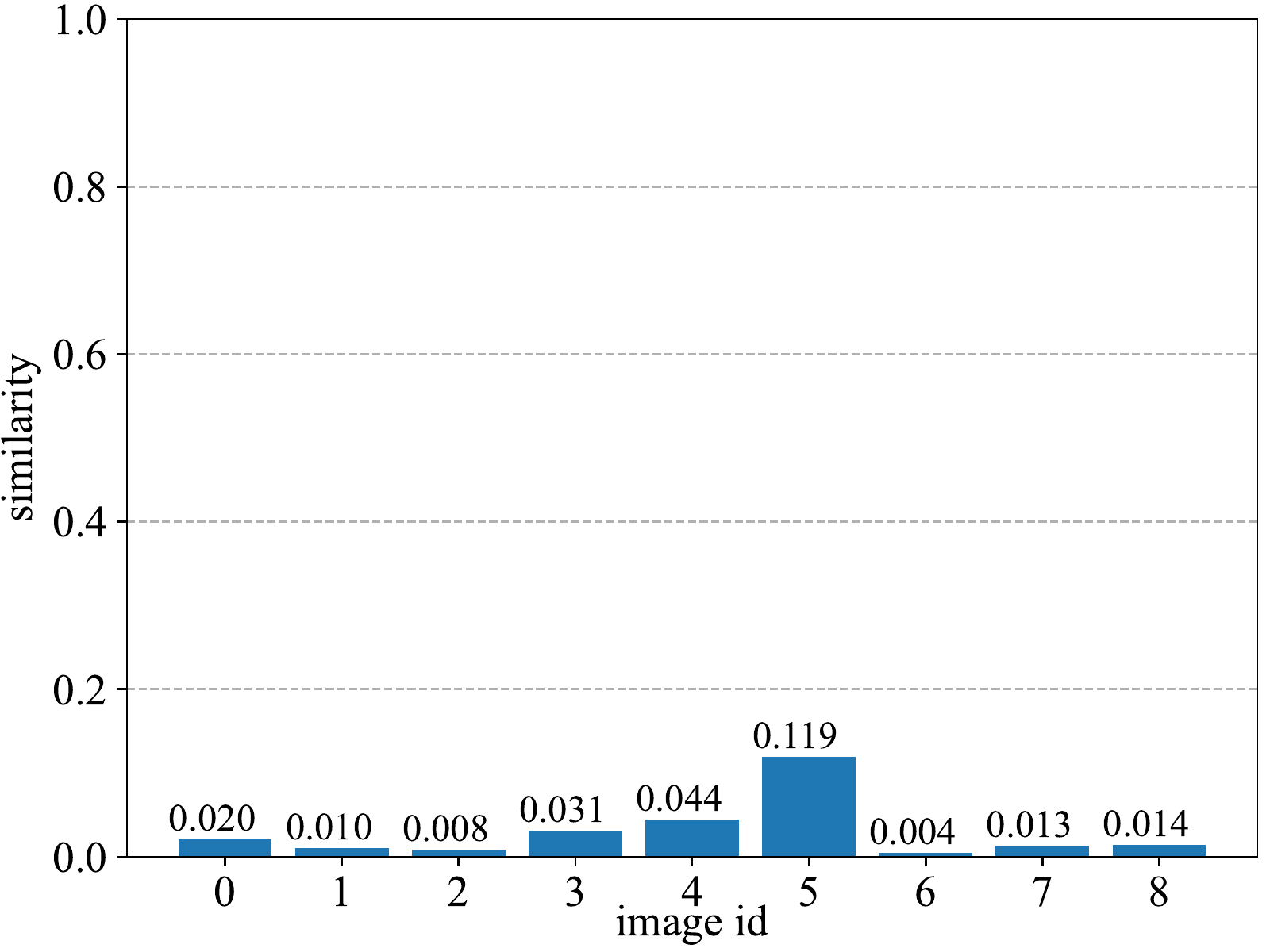}
   & \includegraphics[width=\linewidth]{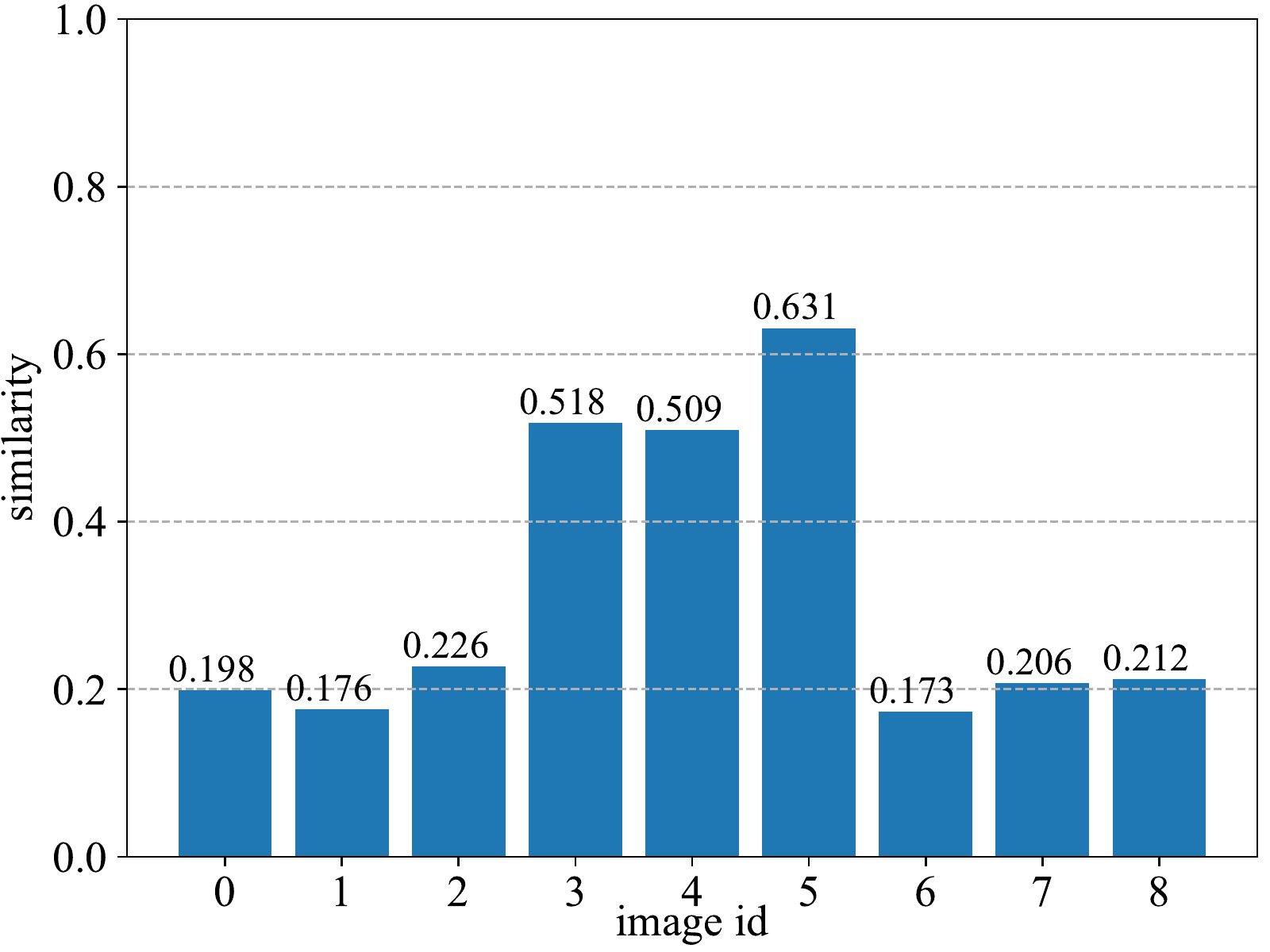} \\

   \includegraphics[width=\linewidth]{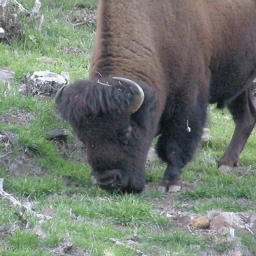}
   & \includegraphics[width=\linewidth]{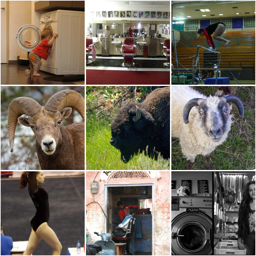}
   & \includegraphics[width=\linewidth]{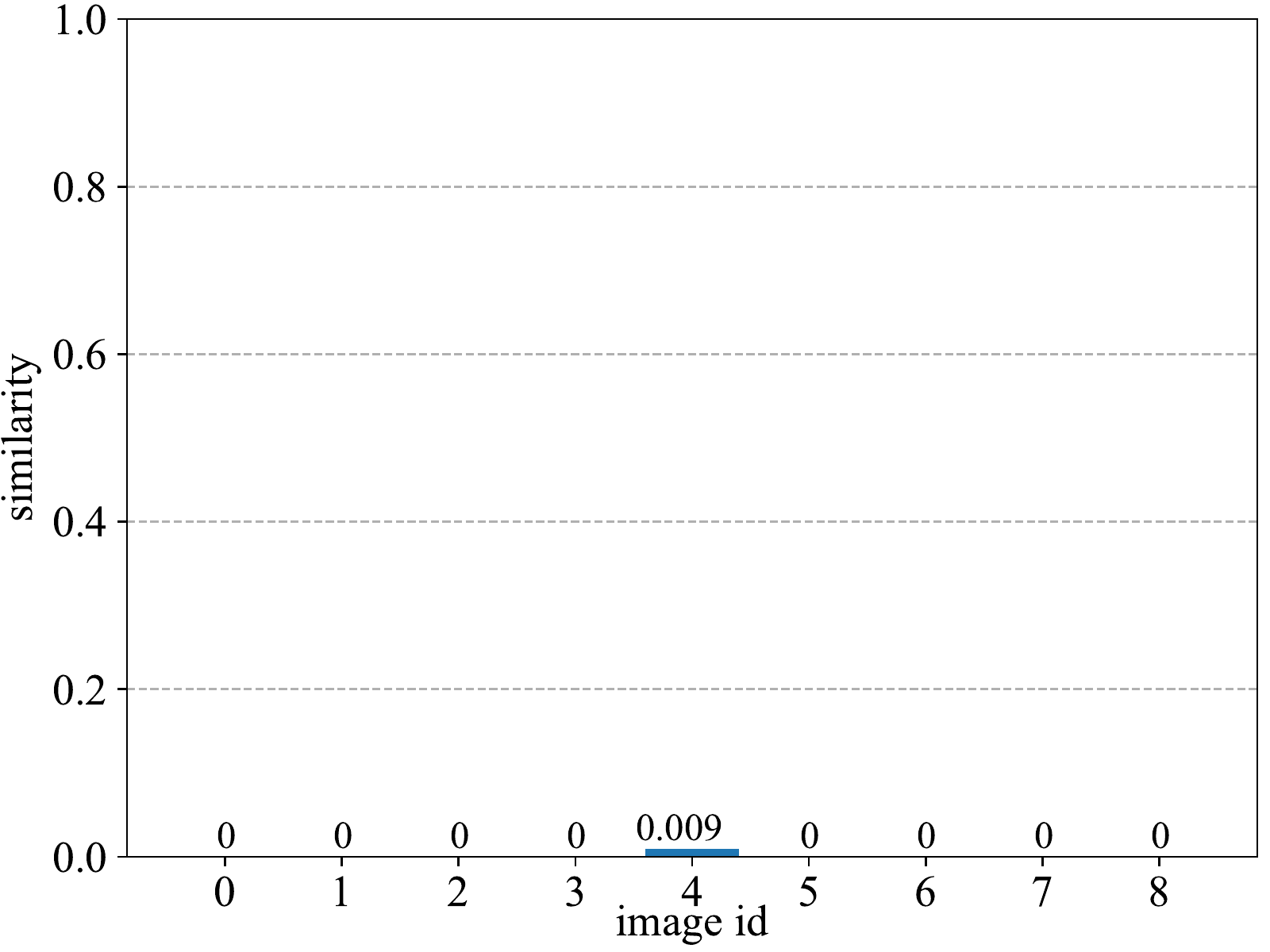}
   & \includegraphics[width=\linewidth]{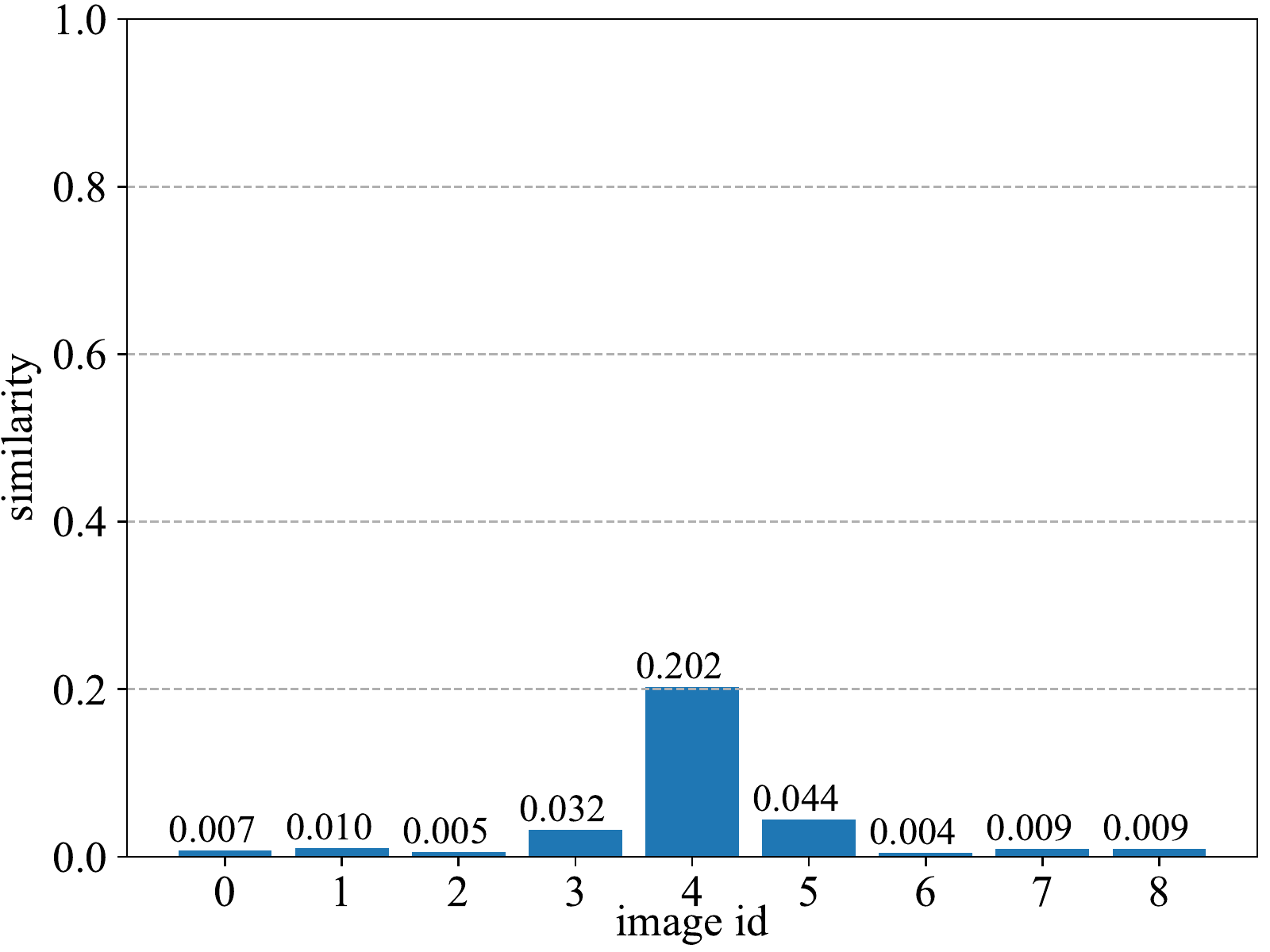}
   & \includegraphics[width=\linewidth]{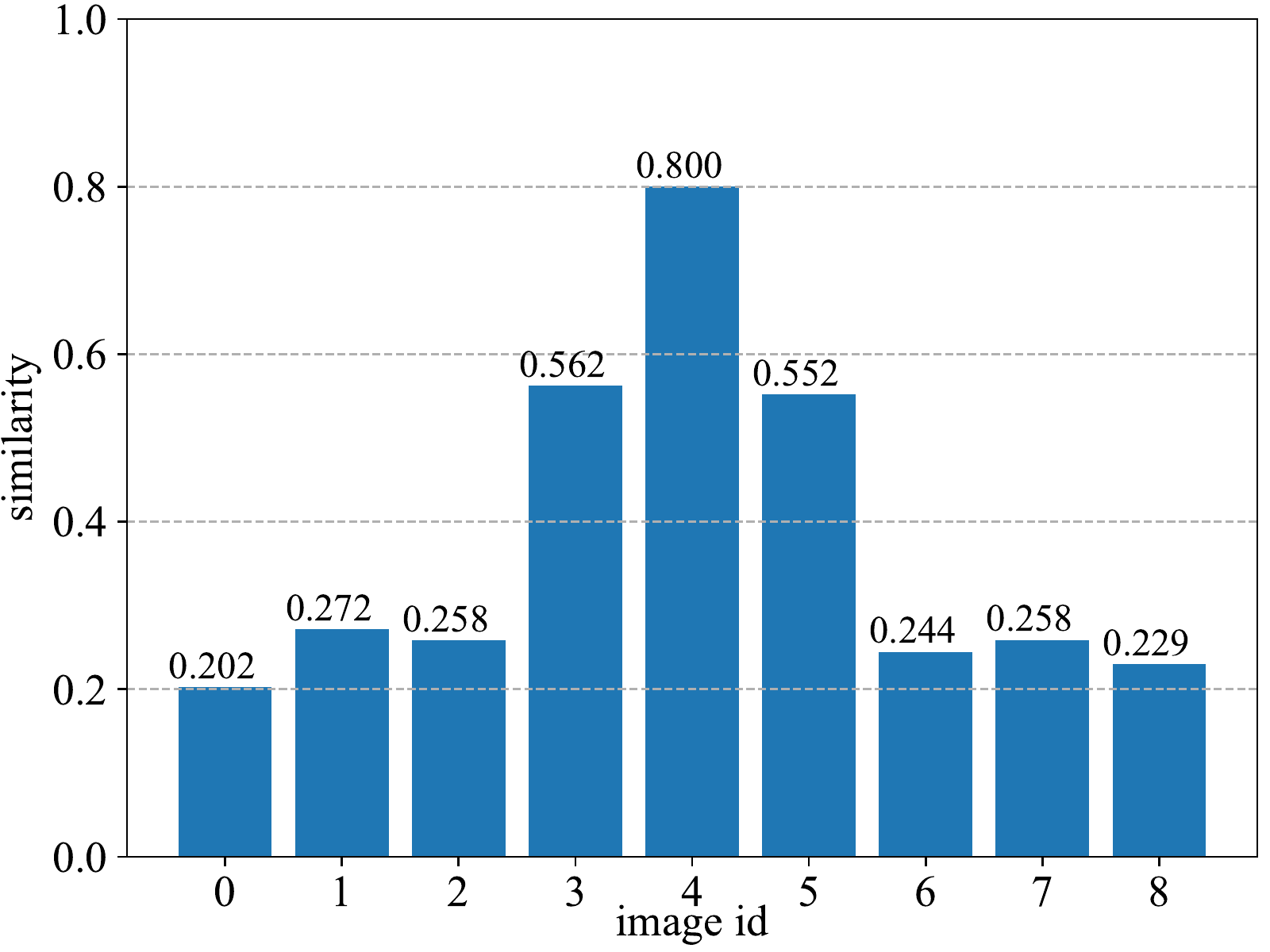} \\

   \includegraphics[width=\linewidth]{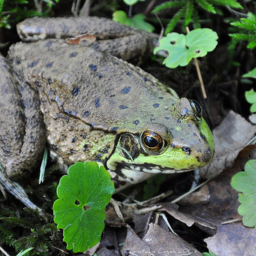}
   & \includegraphics[width=\linewidth]{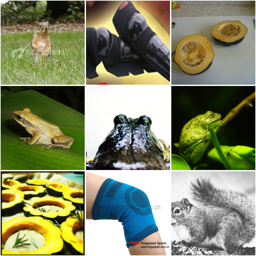}
   & \includegraphics[width=\linewidth]{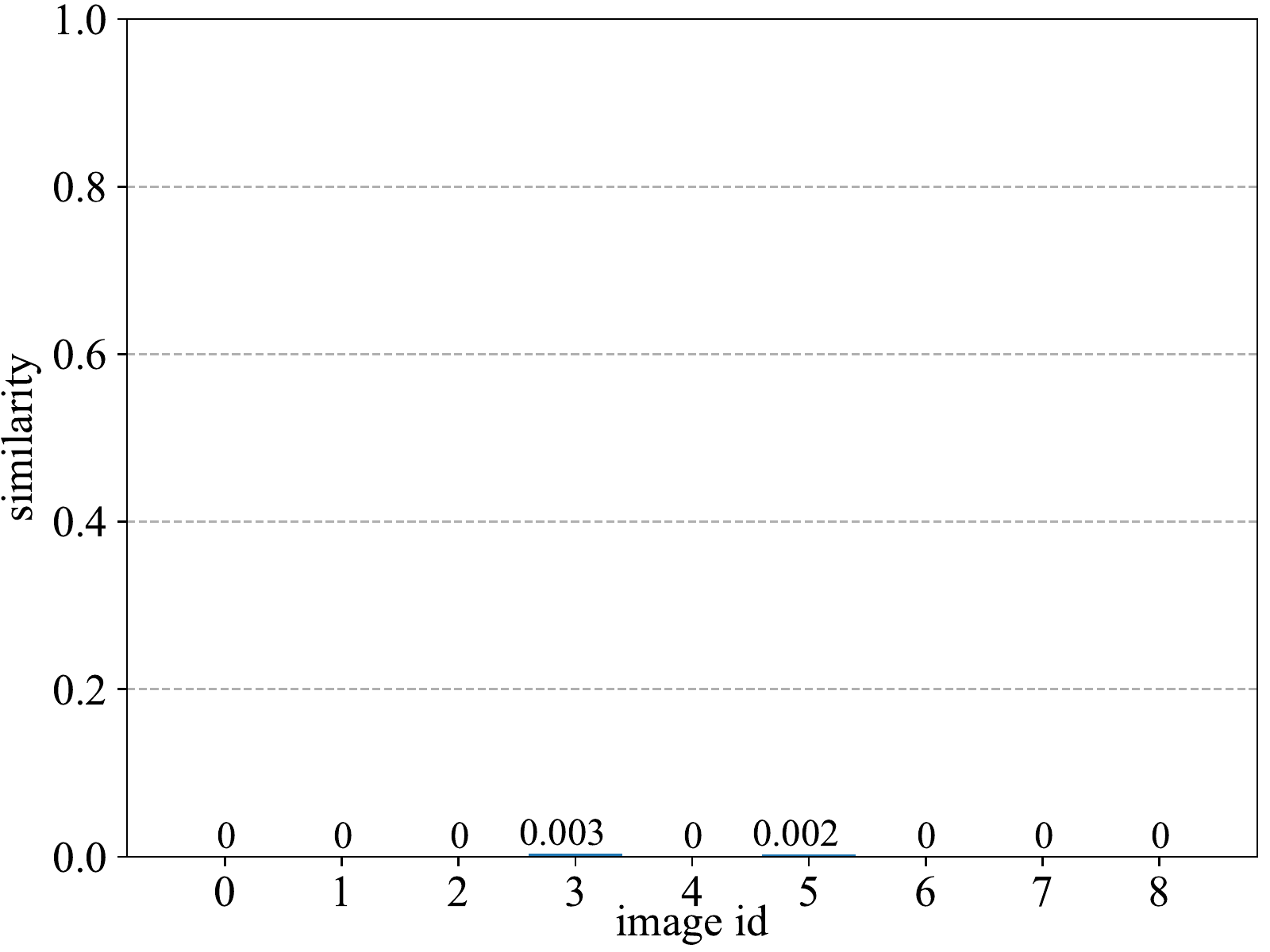}
   & \includegraphics[width=\linewidth]{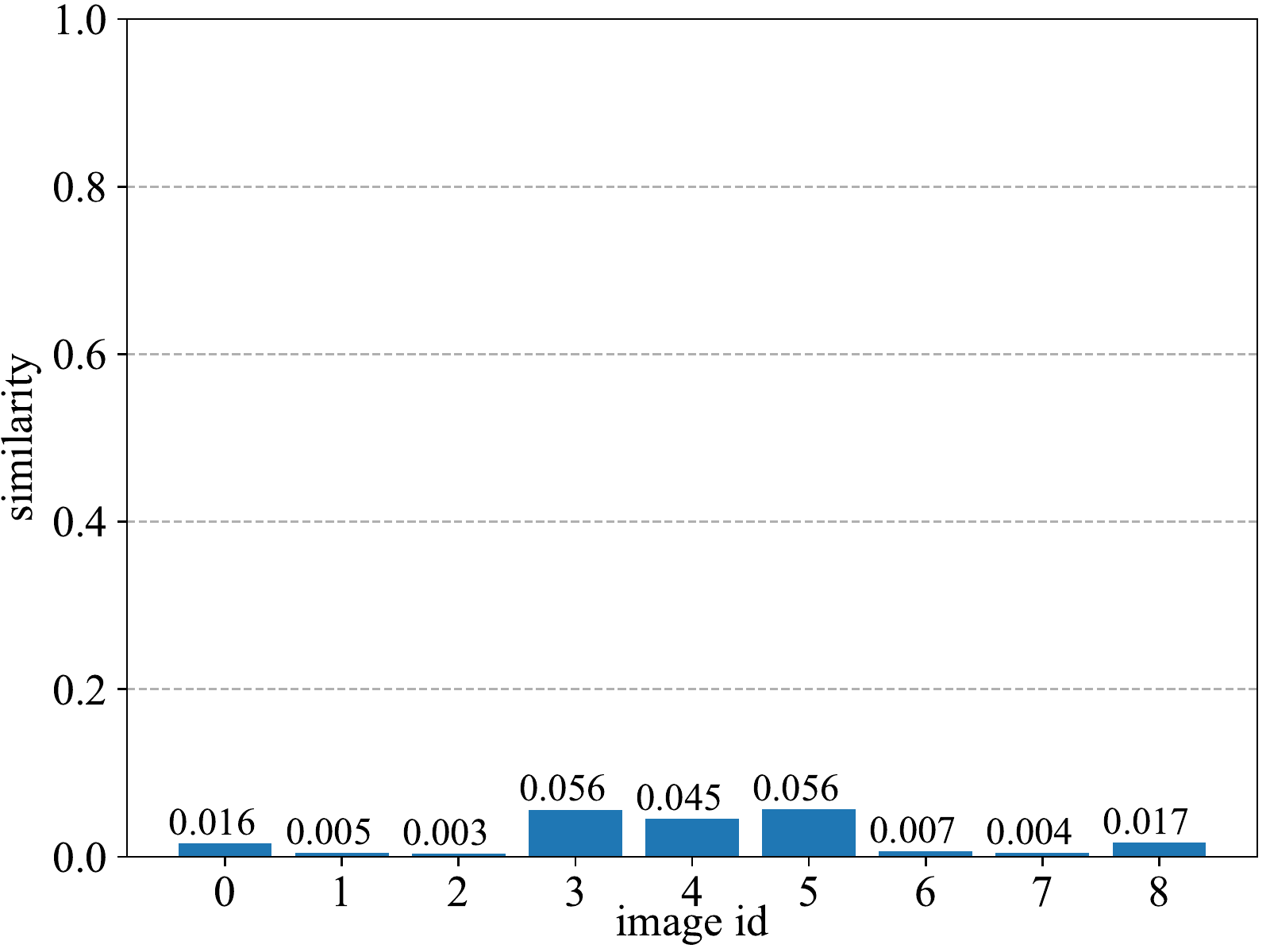}
   & \includegraphics[width=\linewidth]{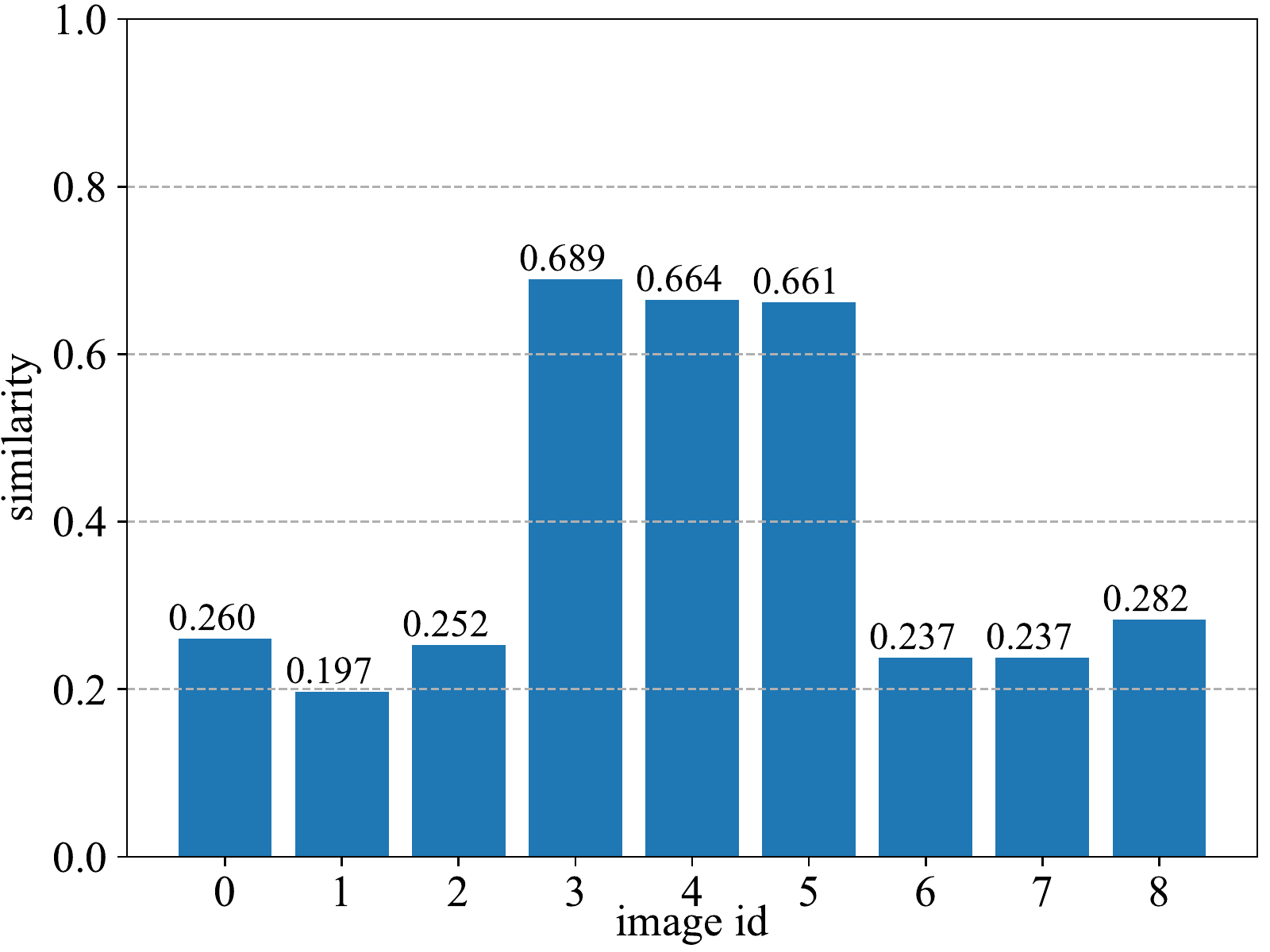} \\

   \includegraphics[width=\linewidth]{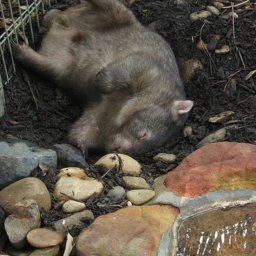}
   & \includegraphics[width=\linewidth]{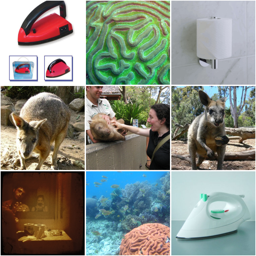}
   & \includegraphics[width=\linewidth]{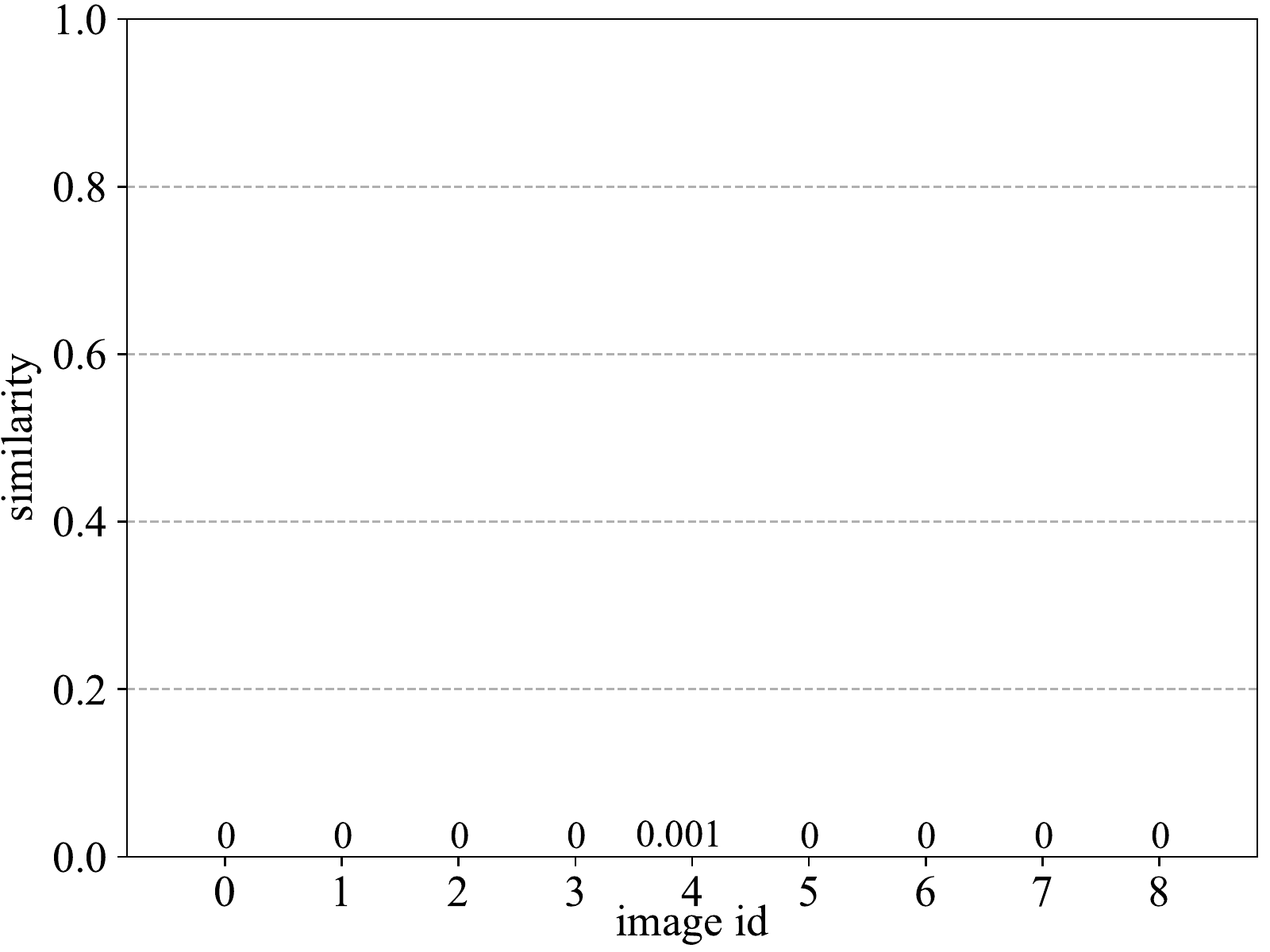}
   & \includegraphics[width=\linewidth]{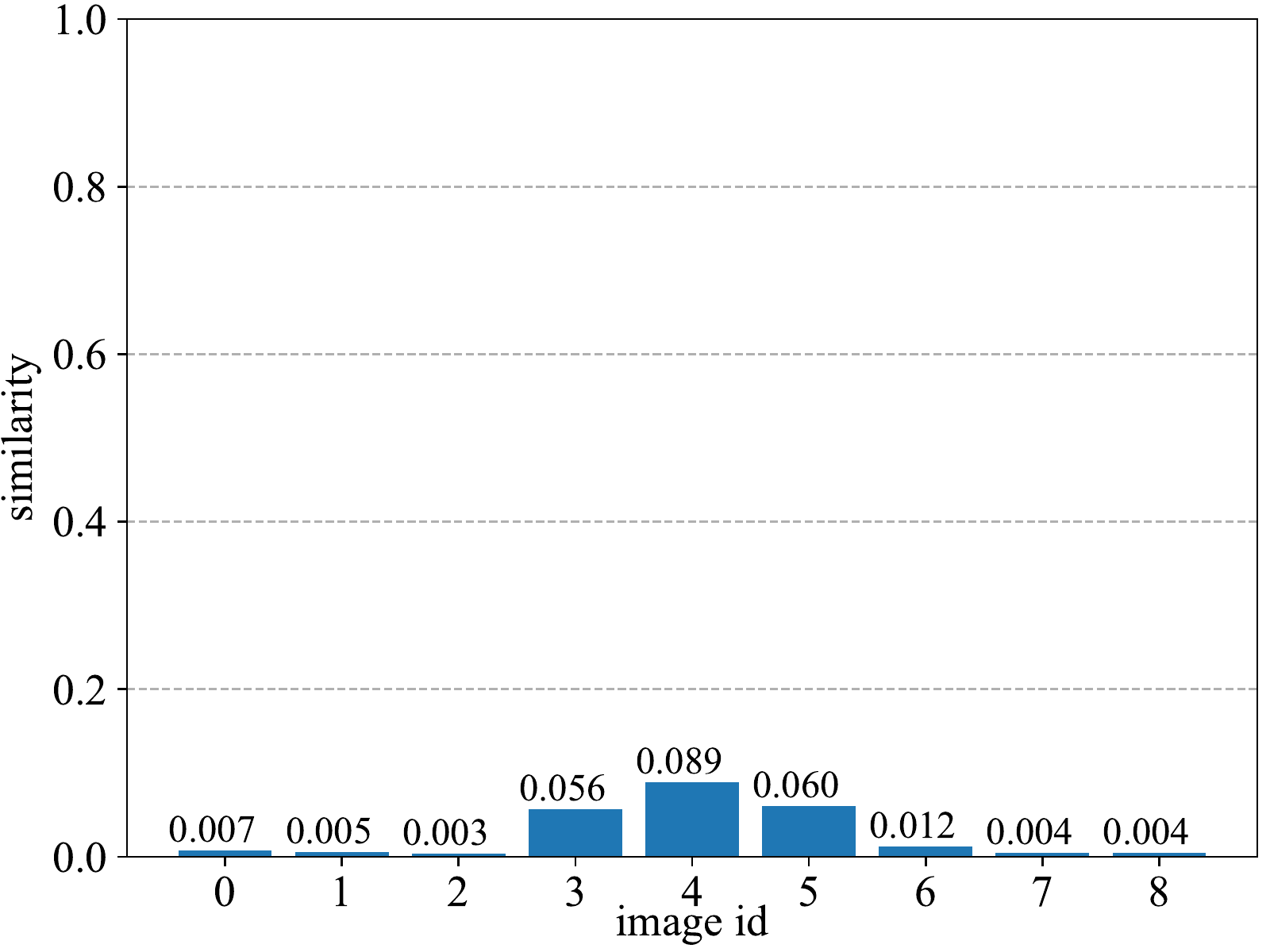}
   & \includegraphics[width=\linewidth]{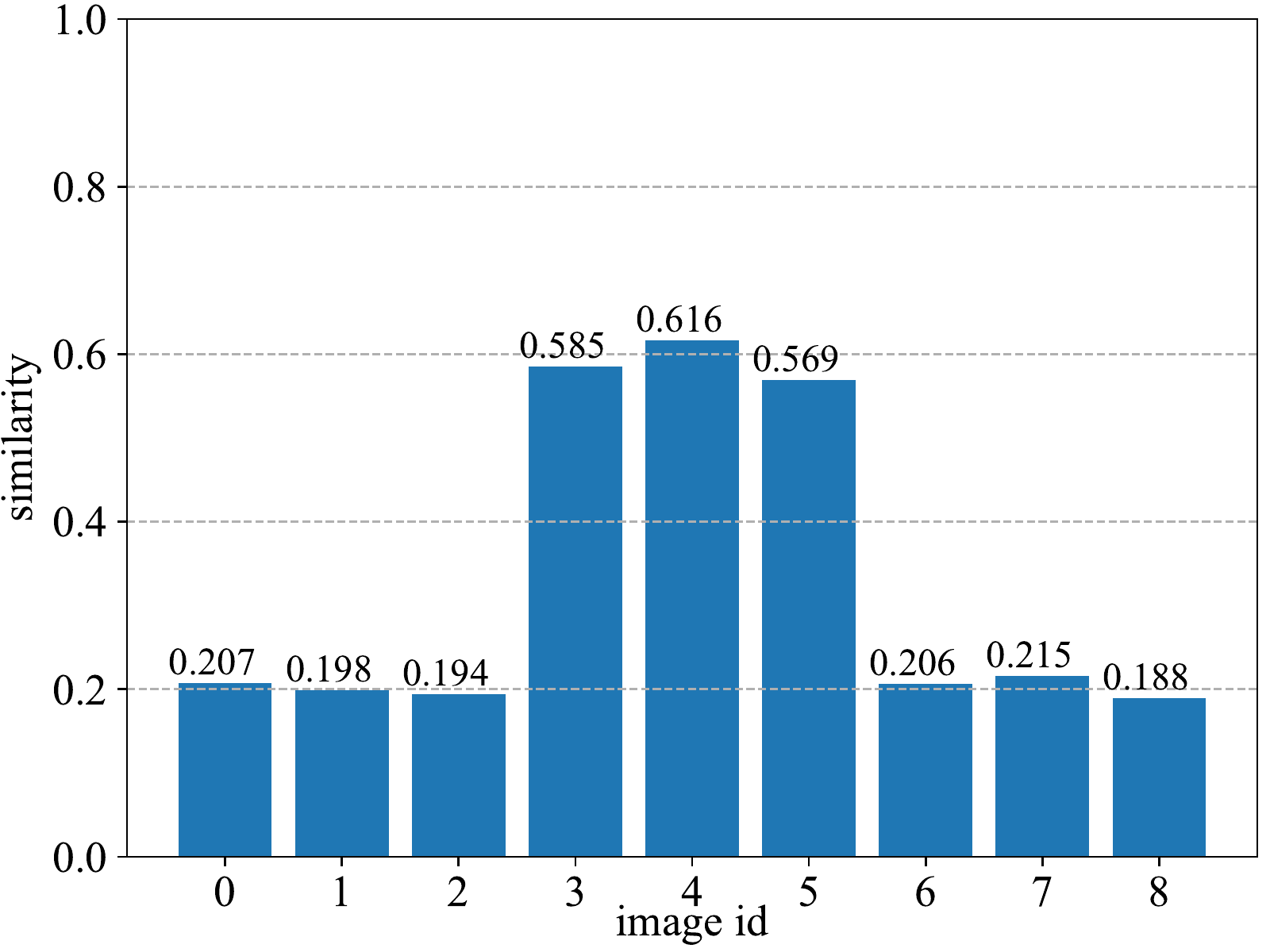} \\

   \includegraphics[width=\linewidth]{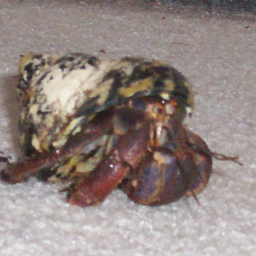}
   & \includegraphics[width=\linewidth]{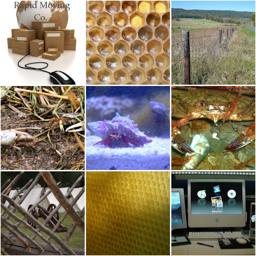}
   & \includegraphics[width=\linewidth]{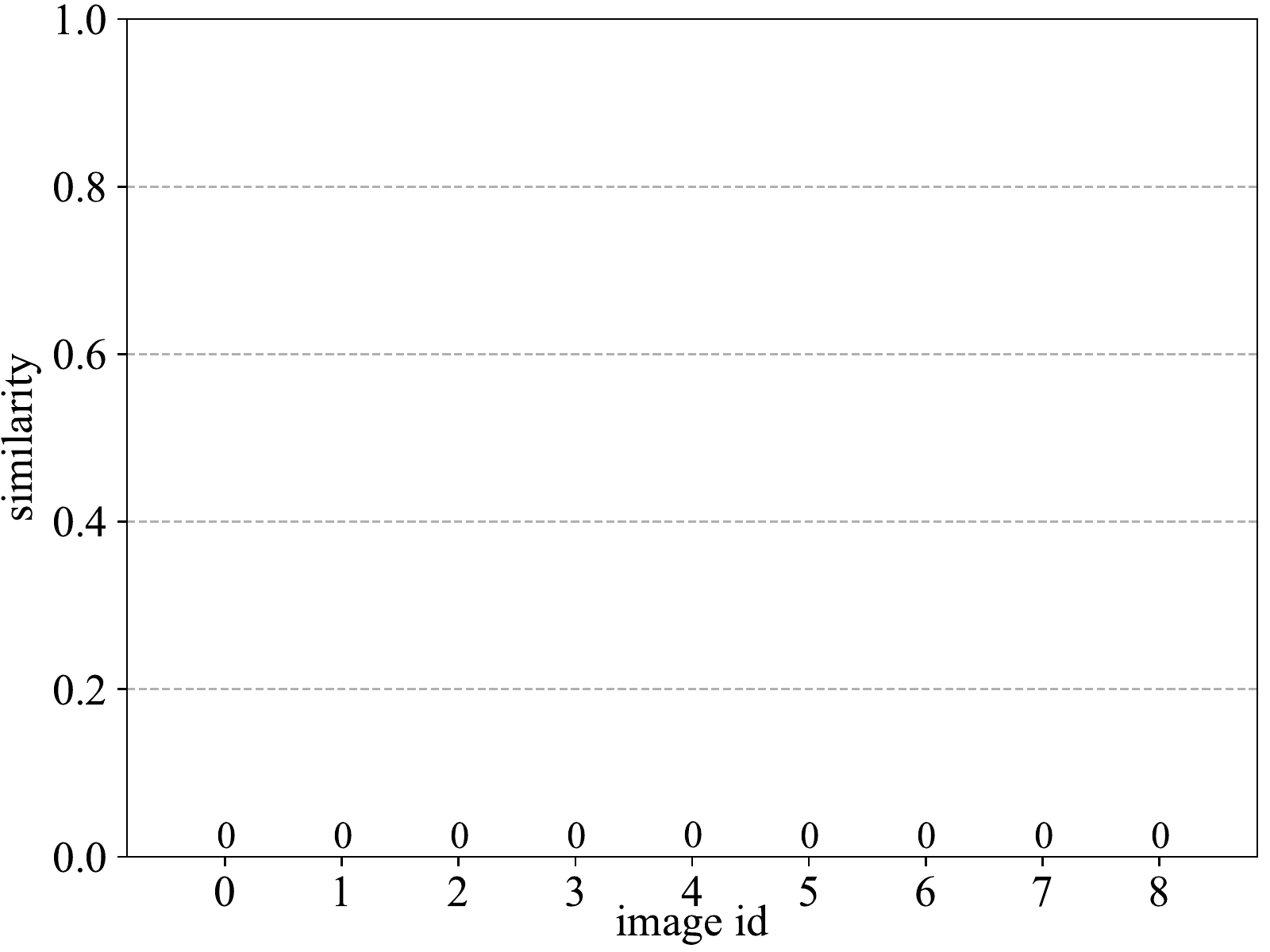}
   & \includegraphics[width=\linewidth]{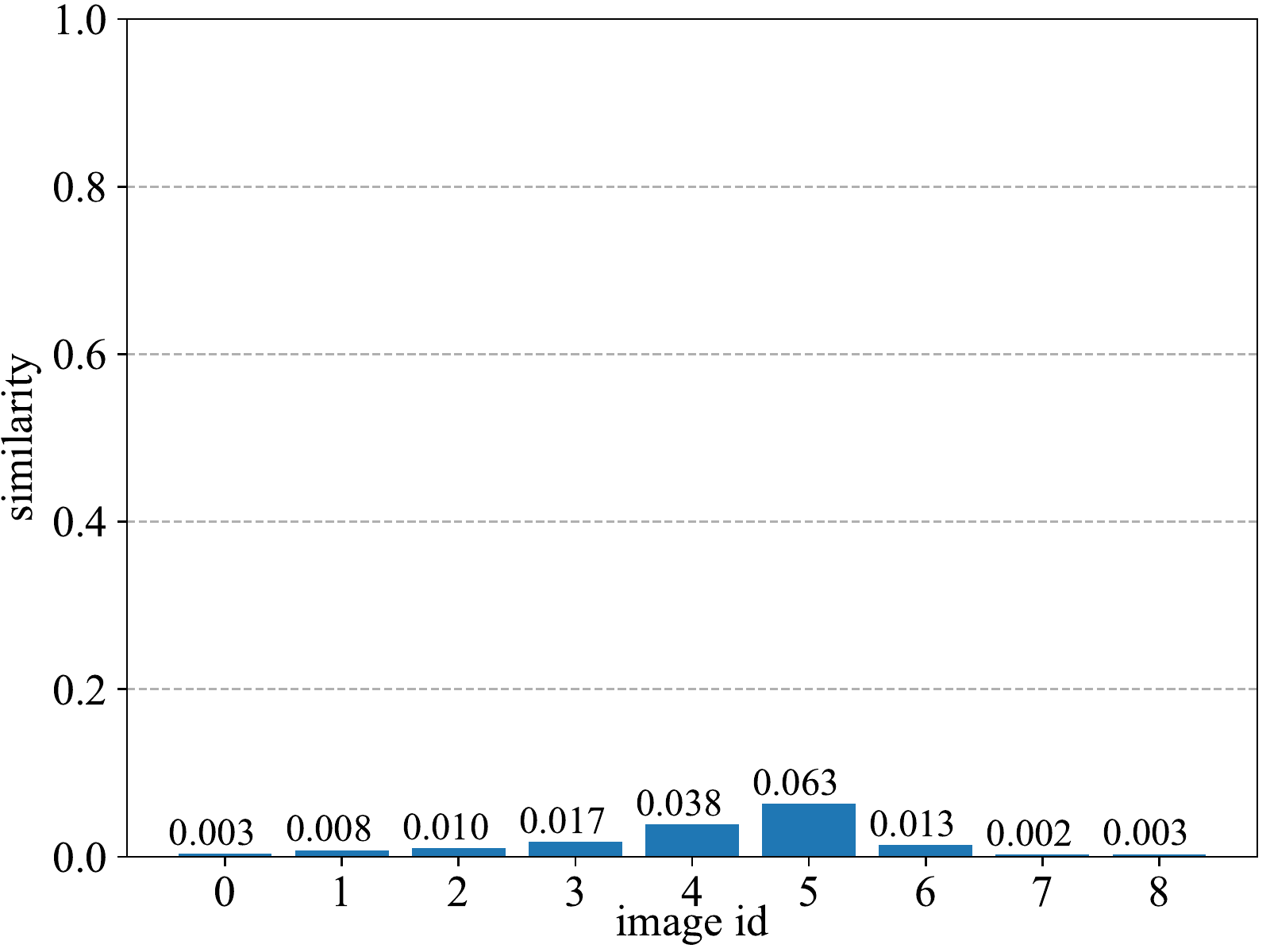}
   & \includegraphics[width=\linewidth]{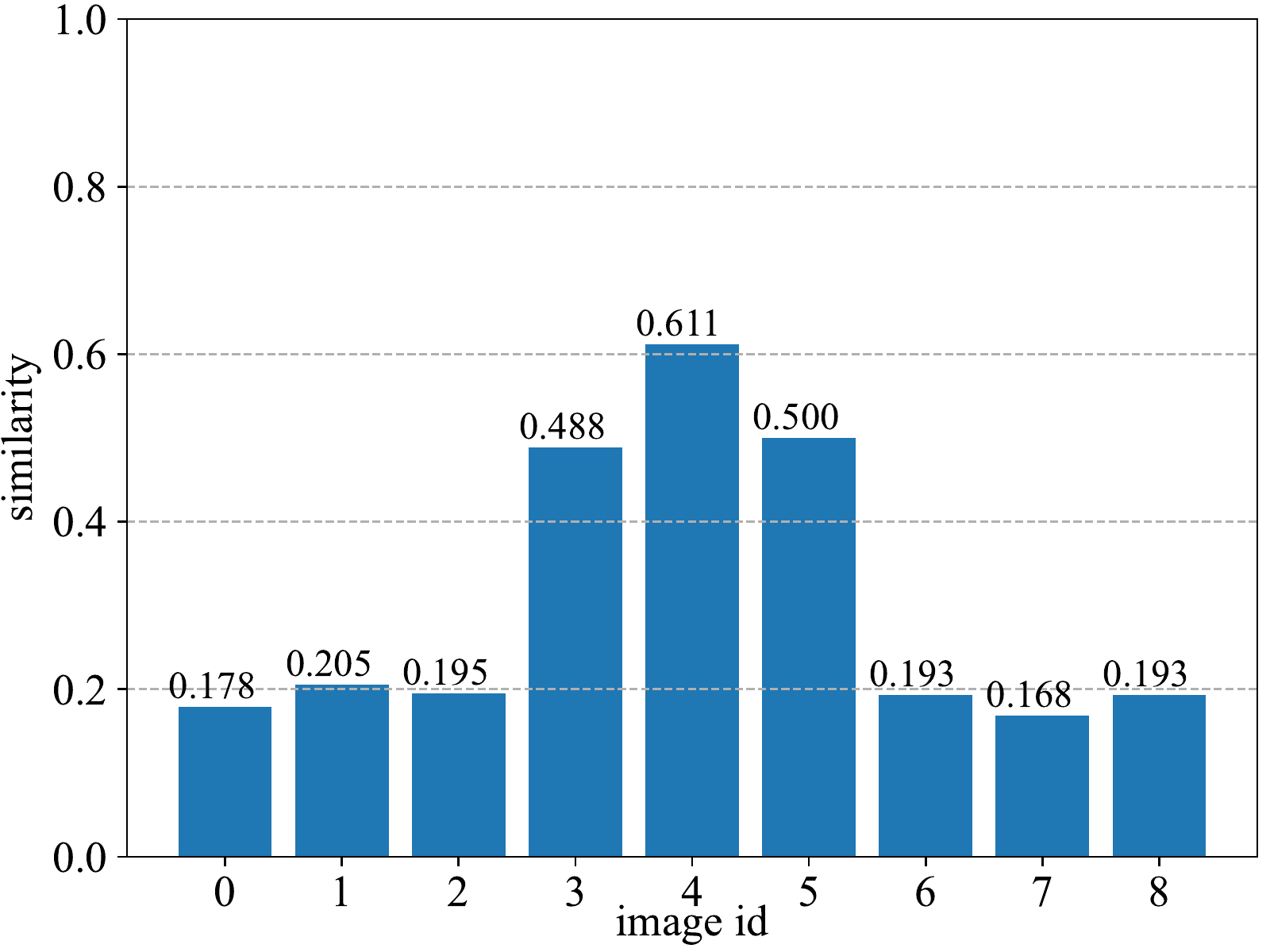} \\

   \includegraphics[width=\linewidth]{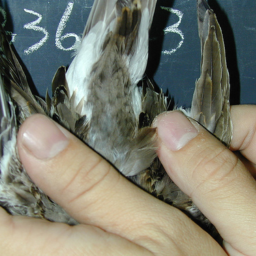}
   & \includegraphics[width=\linewidth]{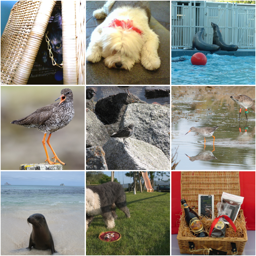}
   & \includegraphics[width=\linewidth]{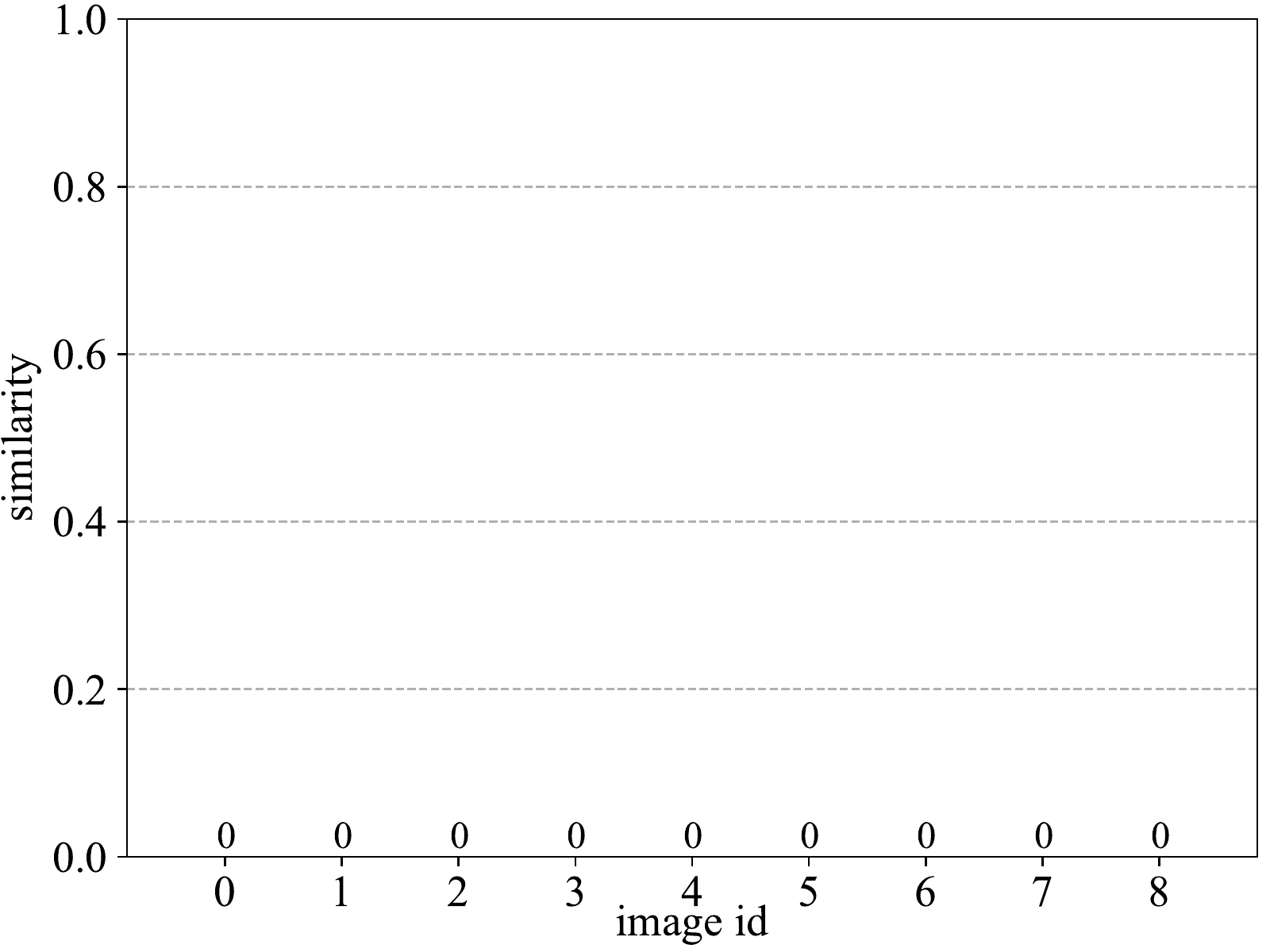}
   & \includegraphics[width=\linewidth]{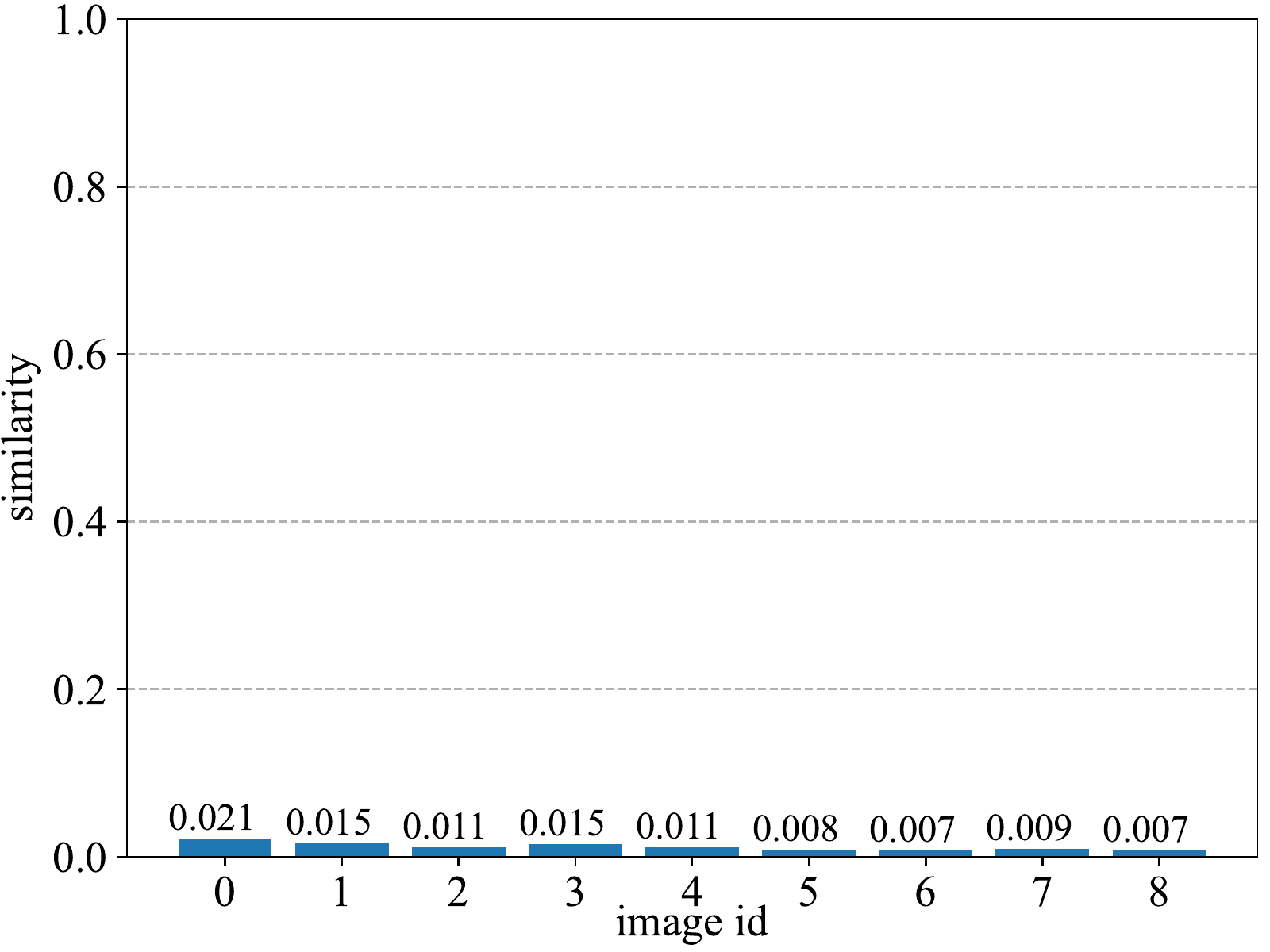}
   & \includegraphics[width=\linewidth]{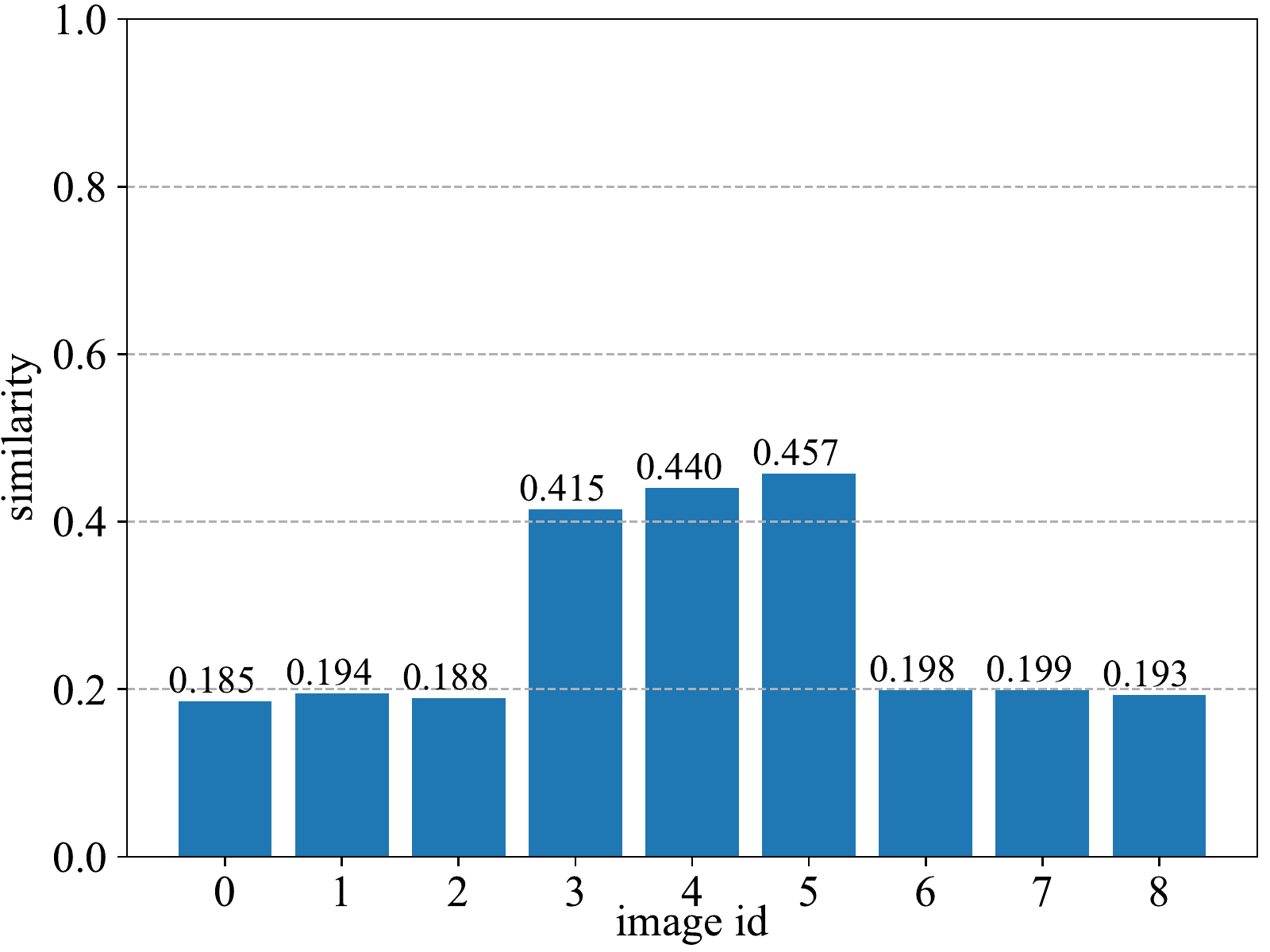} \\

   \includegraphics[width=\linewidth]{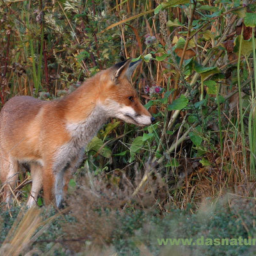}
   & \includegraphics[width=\linewidth]{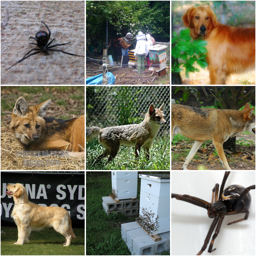}
   & \includegraphics[width=\linewidth]{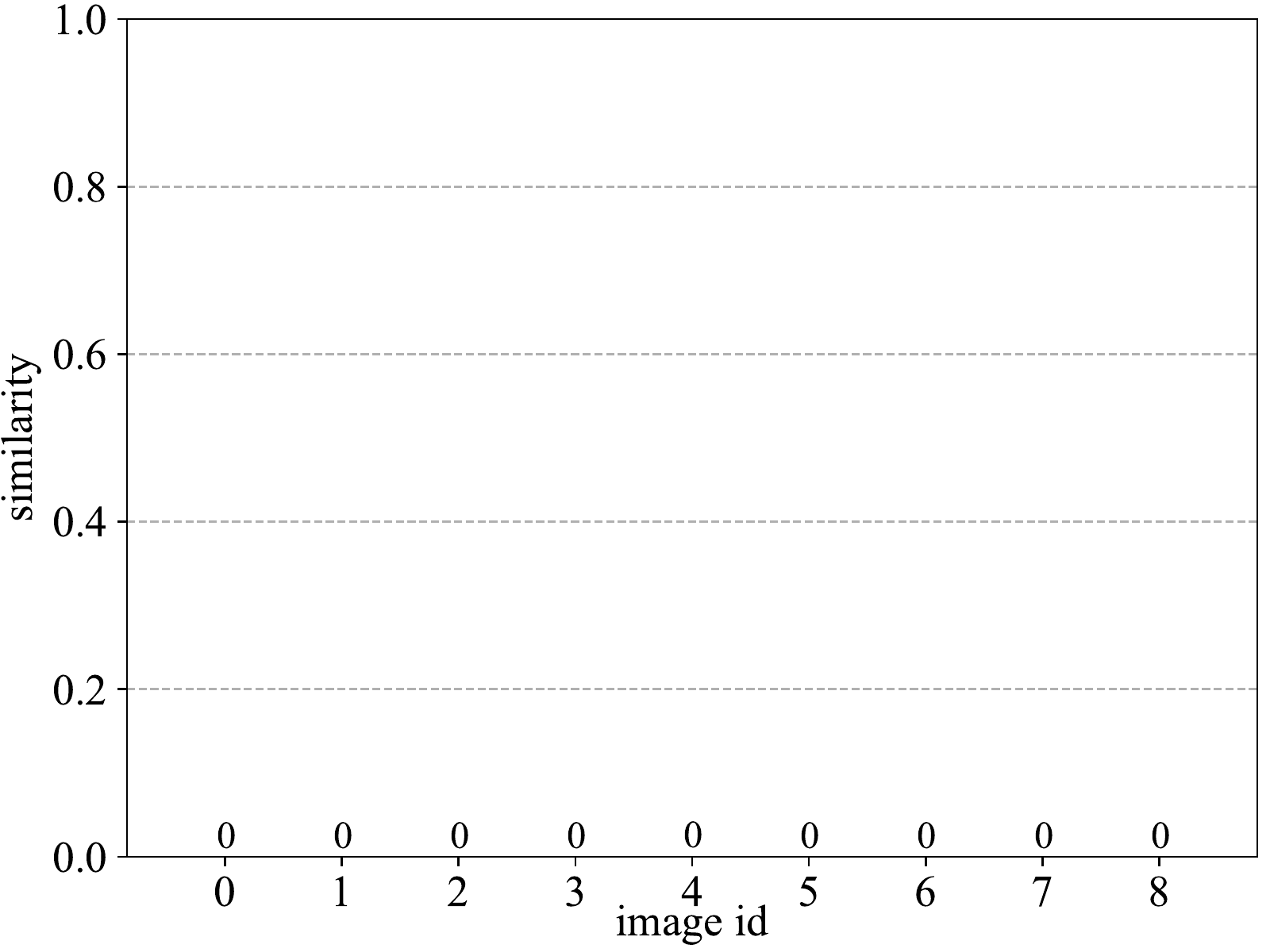}
   & \includegraphics[width=\linewidth]{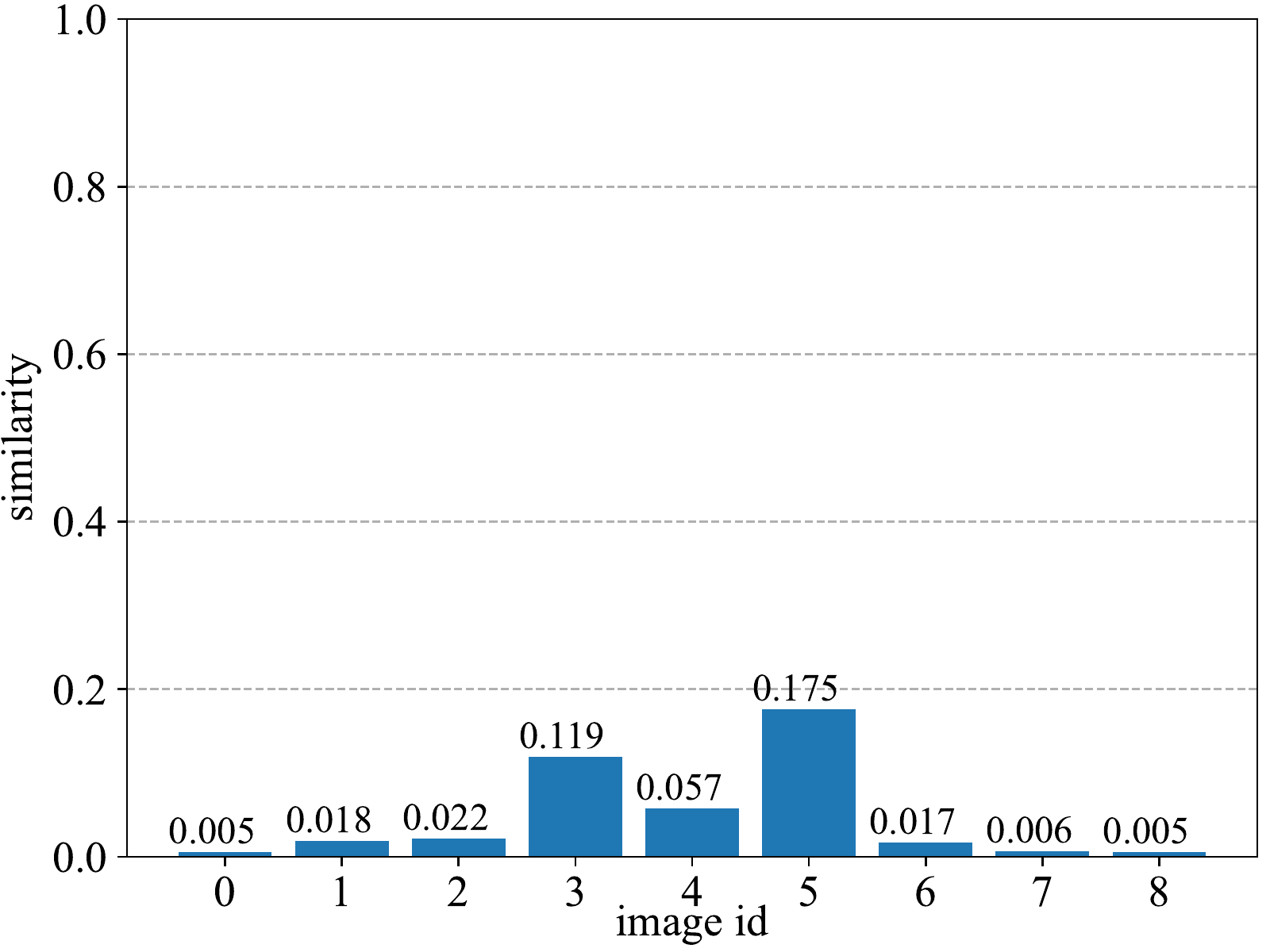}
   & \includegraphics[width=\linewidth]{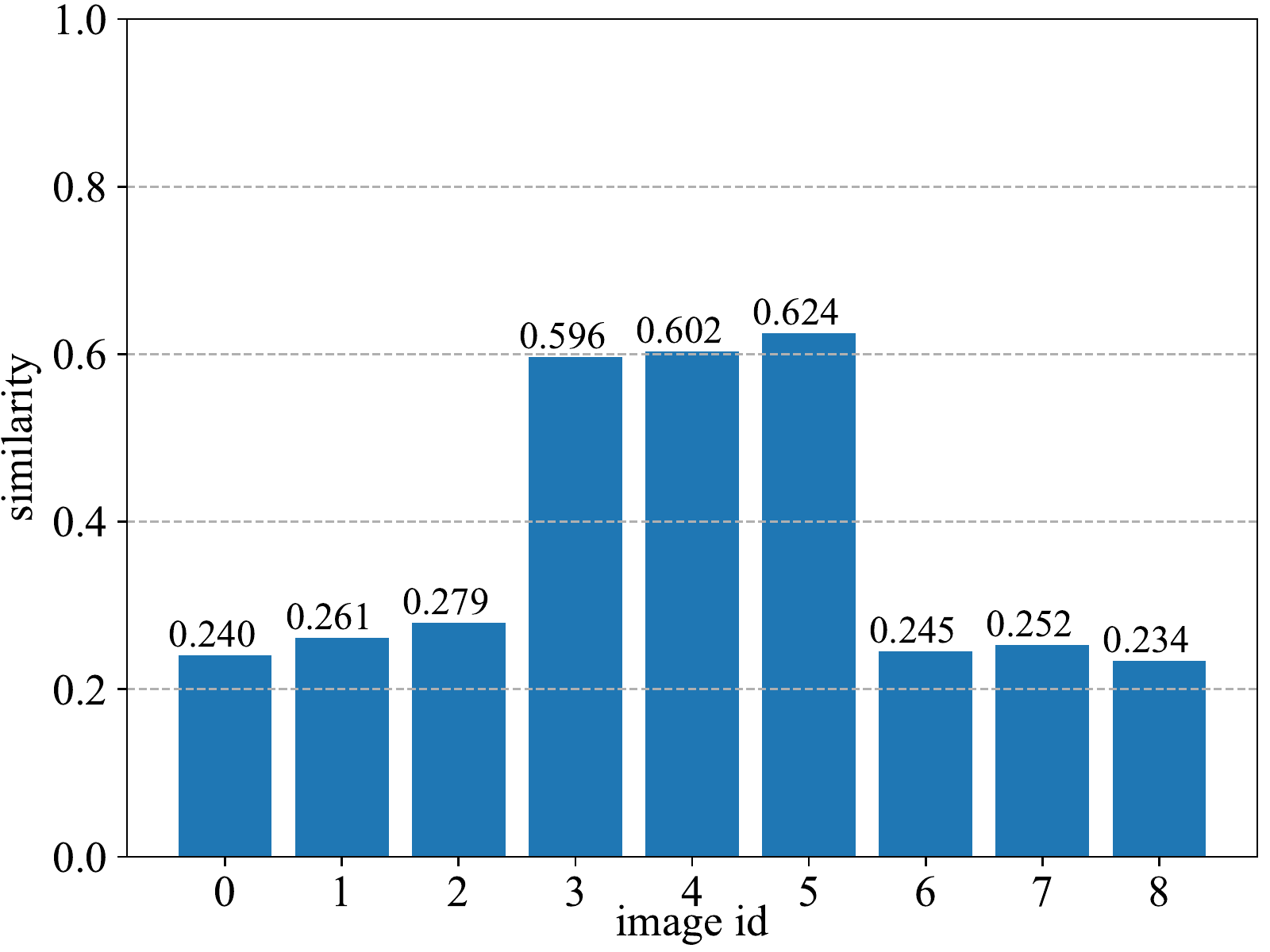} \\
  
\end{tabular}
}
   \vspace{-1em}
   \caption{The visualization of inter-instance similarities on ImageNet-1K.
   The query sample and the image with id 4 in key samples are from the same category. 
   The images with id 3 and 5 come from category similar to query sample.
   }
   \label{fig:visualization3}
\end{figure*}

To further validate the effectiveness of our proposed PatchMix on inter-instance similarity modeling, we provide more visualization results the similarity distribution among multiple images. 
The results are presented in Figure~\ref{fig:visualization1}, Figure~\ref{fig:visualization2} and Figure~\ref{fig:visualization3}. 
For DINO, only the samples with the same category present instance similarity with the corresponding query samples.
In some cases of Figure~\ref{fig:visualization3}, even the samples with the same category only have negligible similarities with the corresponding query samples.
For SDMP, it achieves richer inter-instance similarities among images, especially the samples from the similar categories. 
However, there are also some cases, such as the second and third row in Figure~\ref{fig:visualization}, where the samples with a similar category but significantly low similarity score. 
In contrast, our proposed PatchMix consistently achieves richer inter-instance similarities as expected in Figure~\ref{fig:visualization1}, Figure~\ref{fig:visualization2} and Figure~\ref{fig:visualization3}.

\section{Self-Attention Map Visualization}
\begin{figure*}[ht]
   \renewcommand\tabcolsep{1pt}
   \resizebox{\linewidth}{!}
   {
   \begin{tabular}{cc}
   \toprule
   \rotatebox{90}{{\scriptsize \quad iBOT}} 
   & \includegraphics[width=\linewidth]{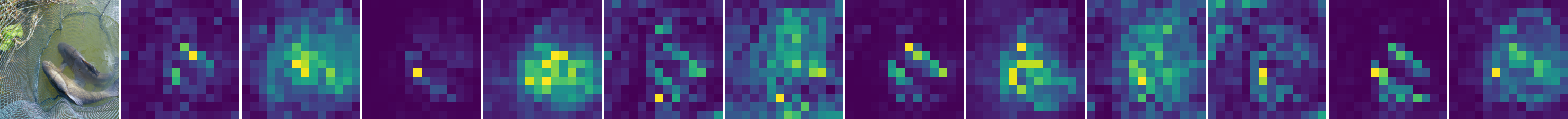} \\
   \rotatebox{90}{{\scriptsize \quad Ours}} 
   & \includegraphics[width=\linewidth]{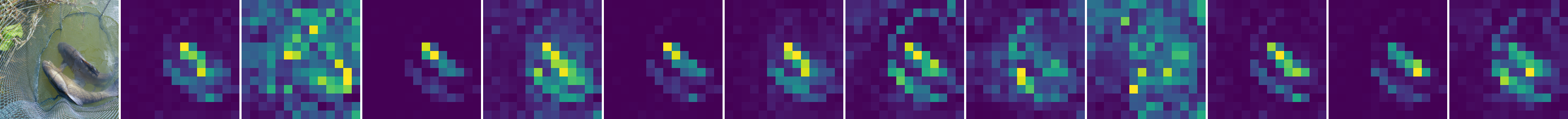} \\
   \toprule
   \rotatebox{90}{{\scriptsize \quad iBOT}} 
   & \includegraphics[width=\linewidth]{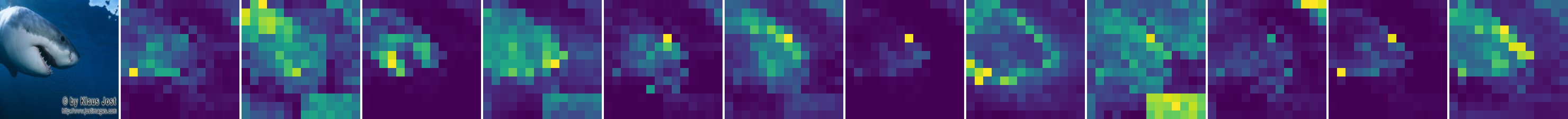} \\
   \rotatebox{90}{{\scriptsize \quad Ours}} 
   & \includegraphics[width=\linewidth]{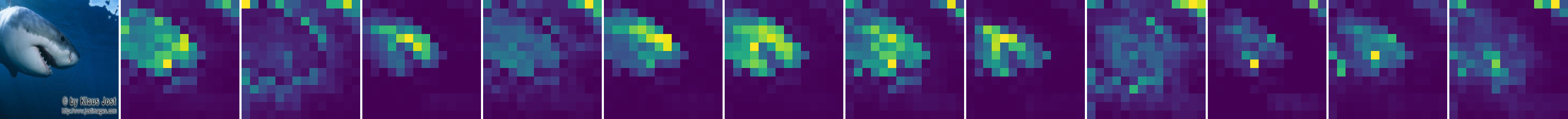} \\
   \toprule
   \rotatebox{90}{{\scriptsize \quad iBOT}} 
   & \includegraphics[width=\linewidth]{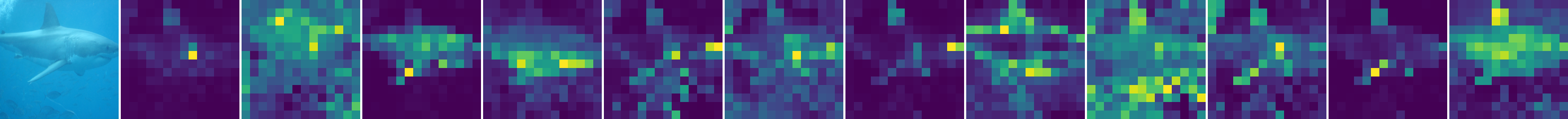} \\
   \rotatebox{90}{{\scriptsize \quad Ours}} 
   & \includegraphics[width=\linewidth]{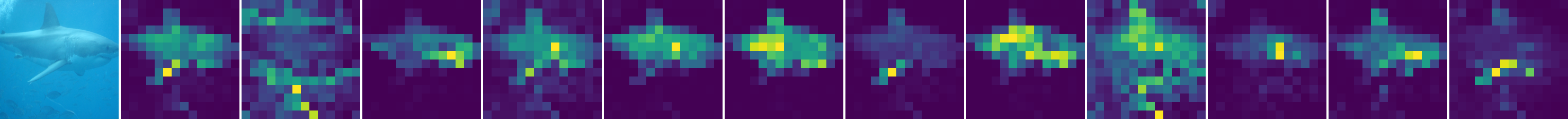} \\
   \toprule
   \rotatebox{90}{{\scriptsize \quad iBOT}} 
   & \includegraphics[width=\linewidth]{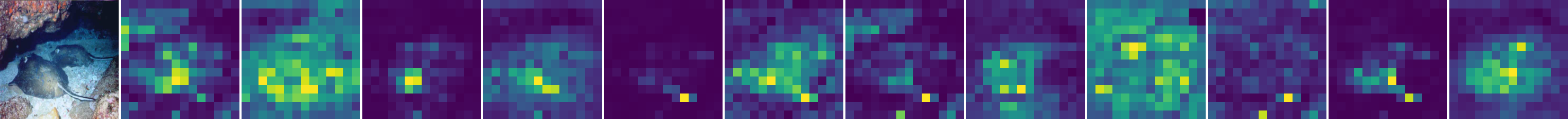} \\
   \rotatebox{90}{{\scriptsize \quad Ours}} 
   & \includegraphics[width=\linewidth]{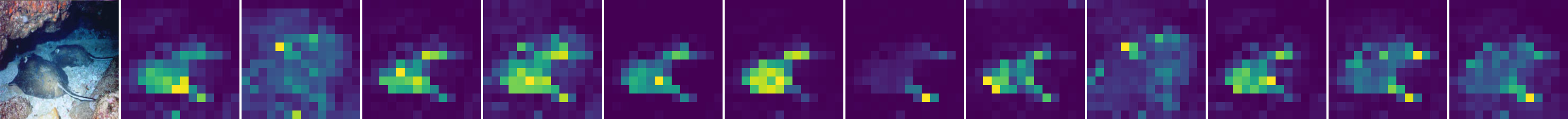} \\
   \toprule
   \rotatebox{90}{{\scriptsize \quad iBOT}} 
   & \includegraphics[width=\linewidth]{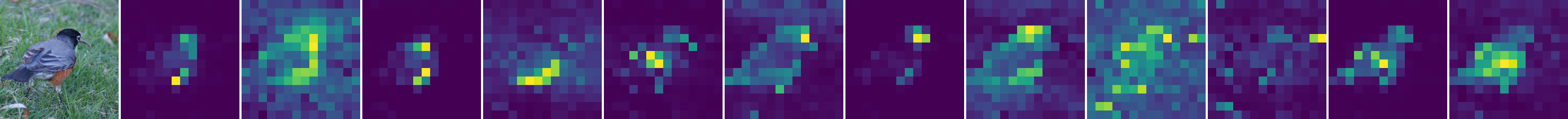} \\
   \rotatebox{90}{{\scriptsize \quad Ours}} 
   & \includegraphics[width=\linewidth]{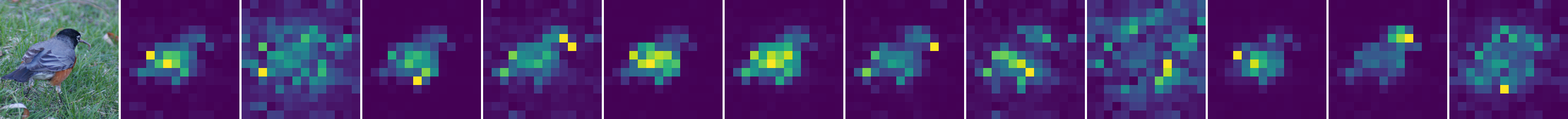} \\
   \toprule
   \rotatebox{90}{{\scriptsize \quad iBOT}} 
   & \includegraphics[width=\linewidth]{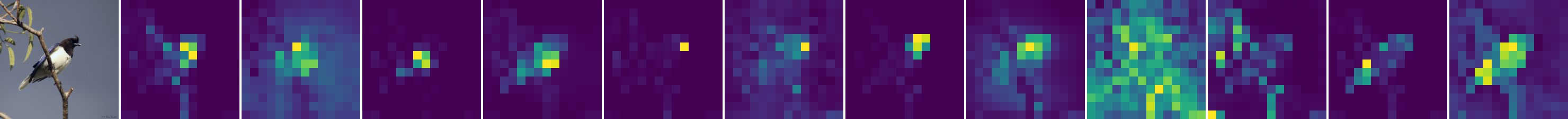} \\
   \rotatebox{90}{{\scriptsize \quad Ours}} 
   & \includegraphics[width=\linewidth]{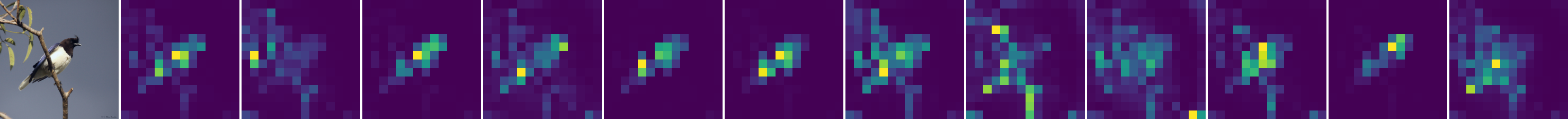} \\
   \toprule
   \rotatebox{90}{{\scriptsize \quad iBOT}} 
   & \includegraphics[width=\linewidth]{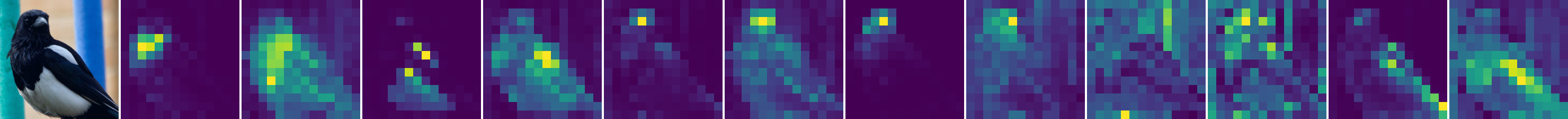} \\
   \rotatebox{90}{{\scriptsize \quad Ours}} 
   & \includegraphics[width=\linewidth]{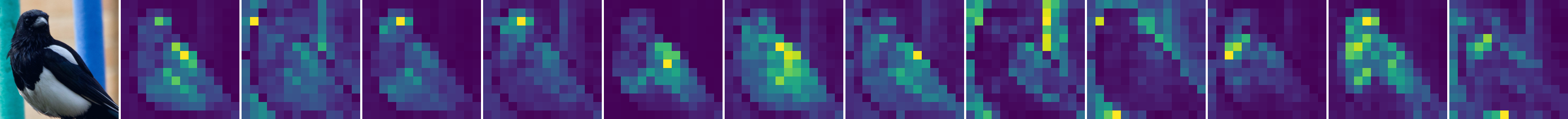} \\
   \toprule
   \rotatebox{90}{{\scriptsize \quad iBOT}} 
   & \includegraphics[width=\linewidth]{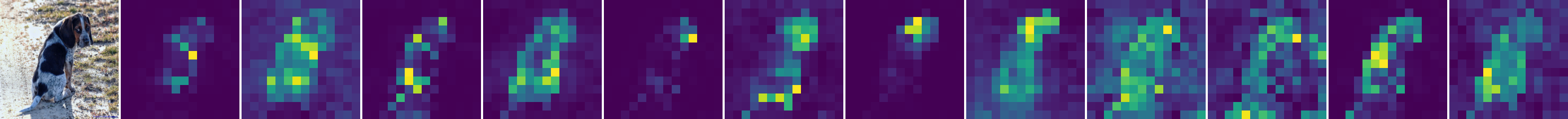} \\
   \rotatebox{90}{{\scriptsize Ours}} 
   & \includegraphics[width=\linewidth]{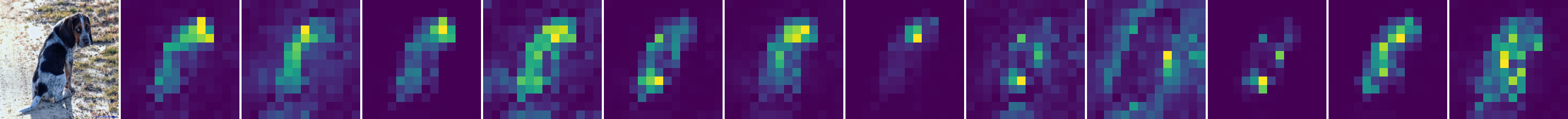} \\
   \toprule
  
   \end{tabular}
   }
   \caption{The visualization of self-attention maps from 12 heads of the last transformer block of ViT-S/16 models, which are pretrained on ImageNet-1K for 800 epochs using iBOT and our proposed PatchMix.
   }
   \label{fig:attention}
\end{figure*}

To analyze the representations learned by our proposed method, we visualize the self-attention scores of [CLS] token from multiple heads of the last transformer block, where the ViT-S/16 model is pretrained on ImageNet-1K for 800 
epochs. 
Specifically, for each head of the last transformer block, the attention scores between [CLS] token and all patch token of the input image is reshaped into the size of 2D map for better visualization, illustrating the importance of patches in the final decision layer. 
For example, the input image patch sequence with 196 tokens (not including [CLS] token) is reshaped into the 2D map with size of $14 \times 14$.
The visualization results are shown in Figure~\ref{fig:attention}, where each sample contains attention maps from 12 heads of [CLS] token.

\end{document}